\documentclass{bmvc2k}

\usepackage{import}
\usepackage{amsmath,amssymb,amsfonts}
\usepackage{algorithmic}
\usepackage{graphicx}
\usepackage{textcomp}
\usepackage{xcolor}
\usepackage{multirow}
\usepackage{lipsum}
\usepackage{amsmath}
\usepackage{amssymb}
\usepackage[ruled,vlined]{algorithm2e}
\usepackage{graphicx}
\usepackage{anyfontsize}
\usepackage{graphics} 
\usepackage{epsfig} 
\usepackage{amsmath,amssymb,amsfonts}
\usepackage{algorithmic,tikz,pgfplots}
\usepackage{bm}
\usepackage{graphicx,color}
 \usetikzlibrary{positioning,shapes,arrows,arrows.meta,fit,backgrounds,calc}
\usepackage{makecell}
\usepackage{multirow}
\usepackage{adjustbox}
 \usepackage{textcomp}
 \usepackage{booktabs,relsize}
 \usepackage{anyfontsize}
 \usepgfplotslibrary{fillbetween}
 \usepackage{wrapfig}
 \usepackage{bm}
\usepackage{subfigure}
\usepackage[subpreambles=true]{standalone}
\graphicspath{{./Figures/}}

\title{Distributed Semantic Segmentation \\ With Improved Rate-Distortion Trade-Off}

\addauthor{Danish Nazir}{danish.nazir@volkswagen.de}{1,2}
\addauthor{Timo Bartels}{timo.bartels1@tu-bs.de}{2}
\addauthor{Thorsten Bagdonat}{thorsten.bagdonat@volkswagen.de}{1}
\addauthor{Tim Fingscheidt}{t.fingscheidt@tu-bs.de}{2}

\addinstitution{
 Group Innovation\\
 Volkswagen AG\\
 Wolfsburg, Germany
}

\addinstitution{
 Institute for Communications Technology\\
 TU Braunschweig\\
 Braunschweig, Germany
}

\runninghead{Nazir et al.}{Improved Distributed Semantic Segmentation}

\begin{document}

\maketitle

\begin{abstract}

 Distributed deep neural networks (DNNs) for dense perception tasks such as semantic segmentation execute an encoder DNN on edge devices, and a decoder DNN typically on a large-scale cloud platform with a particular constraint on transmission bitrate. Recent works employ source codecs to enable bitrate-efficient transmission between the edge device and the cloud. However, as these approaches are typically bound to a particular type of source codec and alternative network architectures are often not explored, this results in a suboptimal rate-distortion (RD) trade-off in the low-bitrate regime. In this work, we propose two novel source codecs that \textit{enable extremely low bitrates, while improving RD performance}. We demonstrate the effectiveness of our proposed source codecs by achieving state-of-the-art performance in distributed semantic segmentation at below 0.2 (0.03) bits per pixel, measured using the mean intersection-over-union metric on ADE20K (Cityscapes). 
 
\end{abstract}

\section{Introduction}
\label{sec:introduction}

Deep neural networks (DNNs) have demonstrated strong performance in many dense machine perception tasks. Particularly in semantic segmentation \cite{long2015fully,houben2022inspect,FCN2,contextualinfo1,contextualinfo2}, which is widely adopted in various industrial applications for environment perception \cite{tang2021dffnet,anand2021agrisegnet,chakravarthy2022dronesegnet}, DNNs are typically deployed on resource-constrained camera-enabled edge devices. However, modern semantic segmentation DNNs \cite{deeplabv3,segdeformer,Segformer} have grown in size and computational complexity, making it difficult to execute them on camera-enabled edge devices due to their limited power and computational capabilities \cite{nazirjd}. Accordingly, current approaches \cite{lohdefink2019low,liu2022improving,torfason2018towards,wang2022learning,liu2022semantic} propose a distributed semantic segmentation paradigm, where the semantic segmentation DNN is distributed such that the image encoder DNN is executed on the edge device, and the decoder DNN on a large-scale cloud platform. Although the distributed application mitigates the computational and power constraints on the edge device, it introduces an additional constraint on transmission bitrate, which is addressed     \begin{wrapfigure}{l}{0.5\textwidth}
        \hspace{-0.8em} 
        \resizebox{1.06\linewidth}{!}{\begin{tikzpicture} [node distance = 1.5cm,font=\fontsize{25}{27}\selectfont]

\node (true_bg) [draw, rectangle, minimum width=16cm, minimum height=4.4cm,  line width=1pt,draw=none, fill=none,yshift=0.cm] at (0,0) {};

\node (background) [draw, rectangle, minimum width=8.5cm, minimum height=4.4cm,  line width=1pt,draw=none, fill=none,yshift=0.0cm,xshift=3.77cm]  at (true_bg.west) {};

\fill[color=bg, opacity=0.75, line width=1.5pt] ([yshift=0cm,xshift=0cm]background.north west) rectangle ([xshift=0cm,yshift=0cm]background.south east);

\node (distributed) [draw, rectangle, minimum width=7.5cm, minimum height=4.4cm,
  line width=1pt, draw=none, fill=none, anchor=north west] 
  at ([xshift=0.7cm]background.north east) {};

\fill[color=bg, opacity=0.75, line width=1.5pt] ([yshift=0cm,xshift=0cm]distributed.north west) rectangle ([xshift=0cm,yshift=0cm]distributed.south east);

\path (background.west); \pgfgetlastxy{\xLeft}{\yBase};
\path (distributed.east); \pgfgetlastxy{\xRight}{\yBase};

\node (cloud_txt) [font=\fontsize{25}{27}\selectfont,  label, text width=4cm, above of=distributed, align=center,yshift=0.1cm, xshift=2.6cm, text=blue ] {{Cloud}};
 
\node (car_txt_1) [font=\fontsize{25}{27}\selectfont,  label, text width=4cm, above of=background, align=center , xshift=-3.05cm,yshift=0.15cm, text=blue] {{Edge} };

\node (car_txt_2) [font=\fontsize{25}{27}\selectfont,  label, text width=4cm, above of=background, align=center , xshift=-2.74cm,yshift=-0.55cm, text=blue] {{Device} };

\node (encoder_1) [encoder, left of=background, xshift=-0.75cm,yshift=-0.75cm, fill=encoder_decoder] { $\mathbf{E}$} ;

\node (source_codec) [draw, rectangle, minimum width=11.1cm, minimum height=3.8cm,  
  line width=1pt, draw=none, fill=none, anchor=north, xshift=-0.01cm, yshift=-0.45cm] 
  at ($(background.north east)!0.51!(distributed.north west)$) {};

\fill[color=rectangle, opacity=0.6, line width=1.5pt] ([yshift=0cm,xshift=0cm]source_codec.north west) rectangle ([xshift=0cm,yshift=0cm]source_codec.south east);

\node[font=\fontsize{24}{26}\selectfont, label, text width=6cm, above of=source_codec, align=center ,yshift=0.05cm ] {Source Codec};
 
\node (source_encoder)[draw, rectangle, minimum width=5.18cm, minimum height=3cm, line width=1pt,draw=none, fill=none, right=0.cm of source_codec.west,xshift=0cm,yshift=-0.42cm]  {};

\fill[color=se, opacity=0.85, line width=1.5pt] ([yshift=0cm,xshift=0cm]source_encoder.north west) rectangle ([xshift=0cm,yshift=0cm]source_encoder.south east);

\node (source_decoder)[draw, rectangle, minimum width=5.18cm, minimum height=3cm,  line width=1pt,draw=none, fill=none,yshift=-0.42cm, xshift=0.cm,right=-5.215cm of source_codec.east] {};

\fill[color=se, opacity=0.85, line width=1.5pt] ([yshift=0cm,xshift=0cm]source_decoder.north west) rectangle ([xshift=0cm,yshift=0cm]source_decoder.south east);

\node[font=\fontsize{23}{25}\selectfont, label, text width=6cm, above of=source_decoder, align=center ,yshift=-0.32cm ] {Source Decoder};

\node (be_encoder) [bottleneck_encoder , right of=encoder_1,xshift=1.18cm, fill=bottleneck_color] { $\mathbf{FE}$} ;
\node (ce_encoder) [compression_encoder,fill=ce_color, right of=be_encoder,xshift=0.9cm] { $\mathbf{CE}$} ;
\node (ce_decoder) [compression_decoder , right of=ce_encoder,xshift=2.8cm, fill=ce_color] { $\mathbf{CD}$} ;

\node[font=\fontsize{23}{25}\selectfont,  label, text width=6cm, above of=source_encoder, align=center ,yshift=-0.32cm ] {Source Encoder};
 
\node (decoder_1) [decoder,right of=ce_decoder, xshift=2.4cm, fill=encoder_decoder,minimum width=2.1cm, minimum height=1.2cm,fill=_decoder] { $\mathbf{JD}$} ;

\draw[arrow] 
   ([xshift=-1.75cm, yshift=0cm]encoder_1.west) 
   to node[midway, xshift=0.2cm, yshift=0.4cm, font=\fontsize{25}{27}\selectfont] 
   {$\mathbf{x}$} 
   (encoder_1);
   
\draw [arrow] (encoder_1) to node[midway, xshift=-0.12cm,yshift=0.4cm,font=\fontsize{25}{27}\selectfont] {$\mathbf{z}$} (be_encoder);
\draw[arrow] (be_encoder) to node[midway, yshift=0.4cm,font=\fontsize{25}{27}\selectfont] {$\mathbf{r}$} (ce_encoder);
\draw[arrow] (ce_encoder) to node[midway, xshift=-0.35cm,yshift=0.47cm,font=\fontsize{25}{27}\selectfont] {$\mathbf{b}$} (ce_decoder);
 \draw[arrow] (ce_decoder) to node[midway, yshift=0.47cm,font=\fontsize{25}{27}\selectfont] {$\hat{\mathbf{r}}$} (decoder_1);
 \draw[arrow] 
   (decoder_1) 
   to node[midway, xshift=-0.24cm, yshift=0.4cm, font=\fontsize{25}{27}\selectfont] 
   {$\mathbf{m}$} 
   ([xshift=1.3cm]decoder_1.east);

\end{tikzpicture}}
        \caption{\textbf{High-level overview of state-of-the-art distributed semantic segmentation} \cite{nazir2025efficient,nazirjd}, applicable also to our proposed methods. Here, $\mathbf{E}$ and $\mathbf{JD}$ are the image encoder and the joint feature and task decoder, respectively. Block $\mathbf{FE}$ represents the feature encoder. We propose improved compression encoders and decoders ($\mathbf{CE}$ and $\mathbf{CD}$), respectively, whose network architectures are detailed in Figs.\ \ref{fig:mean_scale_codec} and \ref{fig:litcm_codec}.}
      \label{fig:high_level_overview}
    \end{wrapfigure}
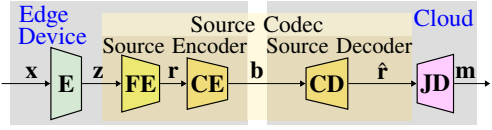by employing a low-complexity source codec to compress the image encoder DNN output (bottleneck) features \cite{ahuja2023neural,nazirjd,nazir2025efficient,matsubara2022sc2,matsubara2019distilled,shao2020bottlenet,matsubara2021neural}.~However, a careful network 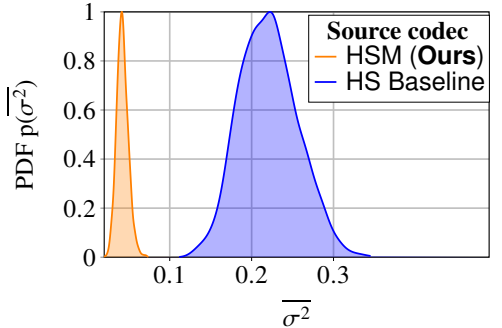
\begin{wrapfigure}{r}{0.5\textwidth}
        \vspace{-1.4em} 
        \hspace{-0.7em} 
            \resizebox{1.0\linewidth}{!}{\begin{tikzpicture}

\begin{axis}[
    width=0.95\columnwidth,
    height=0.65\columnwidth,
    name=meanvar_vs_pdf,
    xmin=0.02, xmax=0.49,
    xtick distance=0.1,
    ytick distance=0.2,
    xtick={0.1,0.2,0.3},
    major tick length=3pt,
    minor tick length=2pt,
    xtick style={color=black},
    x tick label style={/pgf/number format/precision=10},
    ymin=0, ymax=1,
    x grid style={gray!50, line width=1.3pt},
    y grid style={gray!50, line width=1.3pt},
    xmajorgrids,
    ymajorgrids,
    xlabel={$\overline{{\sigma}^{2}}$},
    ylabel={PDF p($\overline{{\sigma}^{2}}$)},
    ylabel style={yshift=13pt},
    xlabel style={yshift=-6pt},
    scaled ticks=false,
    tick align=outside,
    tick pos=left,
    tick label style={color=black,font=\fontsize{20}{22}\selectfont},
    label style={font=\fontsize{20}{22}\selectfont},
    axis line style={line width=0.9pt},
    samples=200,
    smooth,
]

\path[name path=axis] (axis cs:\pgfkeysvalueof{/pgfplots/xmin},0) -- (axis cs:\pgfkeysvalueof{/pgfplots/xmax},0);

\addplot[name path=ours, orange, thick, smooth, line width=1.4pt] table [x index=0, y index=1] {Figures/results/ours_kde_masked.txt};
\label{hsm_ours}

\addplot[orange, fill=orange, opacity=0.3, forget plot] fill between[of=ours and axis];

\addplot[name path=baseline, blue, thick, smooth, line width=1.4pt] table [x index=0, y index=1] {Figures/results/baseline_kde_masked.txt};
\label{baseline}

\addplot[blue, fill=blue, opacity=0.3, forget plot]
fill between[of=baseline and axis];

\end{axis}

\node[draw, fill=white,  font=\fontsize{20}{22}\selectfont]
at ([xshift=-2.5cm,yshift=-1.3cm]meanvar_vs_pdf.north east) {%
\begin{tabular}{@{}l@{\hspace{0.3em}}l@{}}
\multicolumn{2}{@{}l@{}}{\hspace*{4mm}\textbf{Source codec}}\\[-2pt]
\makebox[1em][c]{\ref*{hsm_ours}} & \textsf{HSM (\textbf{Ours})}\\[-2pt]
\makebox[1em][c]{\ref*{baseline}}          & \textsf{HS Baseline}  
\end{tabular}%
};

\end{tikzpicture}}
            \caption{Probability density function (PDF) p($\overline{{\sigma}^{2}}$) of the mean variance $\overline{{\sigma}^{2}}= \overline{{\sigma}^{2}_{i,f}}$, with pixel index $i$ and feature index $f$, computed over all images in $\mathcal{D}^{\mathrm{val}}_{\mathrm{ADE20K}}$, from the latent representation $\hat{\mathbf{r}}$ of the proposed \textsf{HSM} source codec vs.\ the state-of-the-art \textsf{HS} baseline source codec \cite{nazirjd,nazir2025efficient}, whose network architectures are detailed in Figs.\ \ref{fig:mean_scale_codec} and \ref{fig:hyperprior_block}.}
        \label{fig:mean_distribution}
    \end{wrapfigure}design of the source codec is required to achieve optimal rate-distortion (RD) trade-off. Here, the RD trade-off describes the relationship between the transmission bitrate and the semantic segmentation performance, where a better trade-off indicates higher semantic segmentation performance at a given bitrate. A high-level overview of the state-of-the-art (SOTA) distributed semantic segmentation methods \cite{nazirjd,nazir2025efficient} is shown in Fig.\ \ref{fig:high_level_overview}.  

    To achieve a favorable RD trade-off, inspired by the success of the hyperprior architecture \cite{balle2018variational} for source codecs in the learned image compression domain, current SOTA distributed semantic segmentation methods \cite{nazir2025efficient,nazirjd} also employ a hyperprior architecture for their source codecs. In contrast to learned image compression \cite{balle2018variational}, the source codecs used in distributed semantic segmentation operate on latent representation (bottleneck) features rather than on raw images and are designed with lower computational complexity to enable deployment on edge devices. However, the hyperprior architecture \cite{balle2018variational} suffers from a key limitation: It employs a zero mean Gaussian distribution assumption for entropy modeling of the latent representation, which does not fully capture any local context-dependent distribution shifts and therefore produces an inflated  predicted variance \cite{wang2026theoretical,minnen2018joint}, see also Fig.\ \ref{fig:mean_distribution}.
    
    This results in a suboptimal RD trade-off in the low-bitrate regime. To address this limitation, current SOTA learned image compression methods \cite{minnen2018joint,li2025learned,liu2023learned} predict the mean of the latent representation and employ autoregressive context models, which improve the RD performance, but at the cost of significantly increased computational complexity. Since distributed semantic segmentation methods are deployed on resource-constrained edge devices, directly adopting such approaches \cite{minnen2018joint,li2025learned,liu2023learned} remains challenging.
    
     Our contributions with this work are threefold. First, we propose a novel source codec called hyperprior standard deviation and mean (\textsf{HSM}), enabling extremely low bitrates and improving over existing SOTA distributed semantic segmentation methods \cite{nazirjd,nazir2025efficient} by modeling each latent element with \textit{both} its predicted mean and standard deviation to better capture any local context‑dependent distribution shifts in the latent representation, thereby reducing predicted variance of latent representation. Second, to further push the RD performance in the low-bitrate regime, we incorporate channel-wise autoregressive entropy modeling into \textsf{HSM} and thereby propose another novel source codec, namely autoregressive hyperprior standard deviation and mean (\textsf{AR-HSM}). Third, we show that our proposed \textsf{HSM} and \textsf{AR-HSM} source codecs achieve SOTA distributed semantic segmentation performance at below 0.2 (0.03) bpp, measured using the mean intersection-over-union metric on the ADE20K (Cityscapes) datasets.  


	\section{Related Works}
	\label{sec:related_works}
    In this section, we begin with an overview of semantic segmentation methods, followed by their extension to distributed semantic segmentation.
    
    \subsection{Semantic Segmentation}
    
    Semantic segmentation is a dense prediction task that assigns each pixel of an input image to a predefined set of semantic categories. The field of semantic segmentation has made significant advancements with the emergence of fully convolutional deep neural networks (FCNs) \cite{long2015fully,FCN2,FCN3,badrinarayanan2017segnet,ronneberger2015u}, which enabled end-to-end dense prediction, while achieving state-of-the-art (SOTA) performance on several public benchmarks \cite{cityscapes,pascalvoc,cocodataset,ADE20K}. In the following, several works \cite{dialted1,dialted2,dai2017deformable,wang2020deep} proposed various architectural refinements to FCNs \cite{dialted1,dialted2,dai2017deformable,deeplabv3,chen2018atrous,wang2020deep} to further improve their performance. Nowadays, transformer-based methods \cite{Segformer,segdeformer,strudel2021segmenter} have gained prominence in semantic segmentation and emerged as the new SOTA methods, owing to their ability to efficiently model long-range dependencies. Subsequent works proposed various enhancements in transformer-based methods \cite{liu2021swin,katharopoulos2020transformers}, including efficient self-attention for transformer-based image encoders employed in $\texttt{SegFormer}$ \cite{Segformer} and internal-external context mining for a transformer-based decoder architecture utilized in $\texttt{SegDefomer}$ \cite{segdeformer}, upon which we build in this work.
    
    \subsection{Distributed Semantic Segmentation}
    Camera-enabled edge devices are subject to strict computational and power constraints, which may limit their ability to fully execute semantic segmentation DNNs. To address this limitation, earlier works \cite{lohdefink2019low,liu2022improving,torfason2018towards,wang2022learning,liu2022semantic} follow a distributed semantic segmentation paradigm, where the execution of a semantic segmentation DNN is distributed over a client$/$server architecture. While the distributed semantic segmentation application alleviates the computational and power limitations of the edge device, it introduces a new constraint on transmission bitrate between the edge device and the cloud, which directly impacts the performance of the semantic segmentation task.
    
    To address the bitrate efficiency constraint, inspired by learned image compression methods \cite{balle2018variational,factorizedprior}, recent works \cite{ahuja2023neural,nazirjd,matsubara2022bottlefit,matsubara2022sc2,matsubara2019distilled,shao2020bottlenet,matsubara2021neural} propose to employ low-complexity source codecs to compress the latent (bottleneck) features. Early source codec designs \cite{matsubara2019distilled,matsubara2021neural,matsubara2022sc2,shao2020bottlenet,matsubara2022bottlefit} employed a factorized-prior architecture \cite{factorizedprior}, which exhibits inferior rate-distortion (RD) performance compared to the hyperprior architectures utilized in more recent methods \cite{ahuja2023neural,nazirjd,nazir2025efficient}. However, the hyperprior architecture \cite{balle2018variational} assumes a zero-mean Gaussian distribution for the entropy modeling of the latent representation. As the latent representation exhibits both spatial and contextual distribution dependencies, enforcing a zero mean Gaussian entropy model leads to an inflated predicted variance, resulting in a suboptimal RD trade-off, especially in the low-bitrate regime. To address this limitation, inspired by \cite{minnen2018joint}, \textit{we propose a novel hyperprior standard deviation and mean (\textsf{HSM}) source codec} for bottleneck feature compression and take no such zero-mean assumption on the latent representation. So far, our proposed \textsf{HSM} does not exploit inter-channel dependencies. Therefore, inspired by learned image compression methods \cite{minnen2020channel,liu2023learned,li2025learned}, we propose to incorporate channel-wise autoregressive entropy modeling into the \textsf{HSM} source codec, leading to the proposed \textit{autoregressive} hyperprior standard deviation and mean (\textsf{AR-HSM}) source codec that explicitly models the inter-channel dependencies.

	\section{Methods}
	\label{sec:method}

    In this section, we first provide an overview of current state-of-the-art distributed semantic segmentation methods, to then allow an in-depth description of our two proposed source codecs for bottleneck feature compression.
    
    \subsection{Overview of Distributed Semantic Segmentation}
    \label{subsec:overview_distributed_semantic_segmentation}
    
 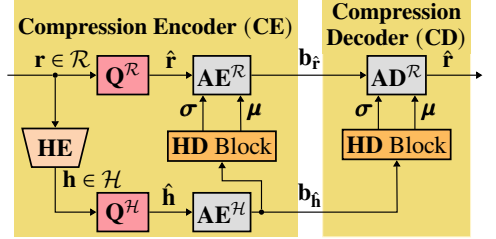
\begin{wrapfigure}{r}{0.5\textwidth}
        \vspace{-1.1em} 
        \hspace{-2em} 
            \resizebox{1.1\linewidth}{!}{




\definecolor{encoder_decoder}{RGB}{213, 232, 212}
\definecolor{rectangle}{RGB}{255, 245, 204}
\definecolor{bottleneck_color}{RGB}{230, 221, 184}
\definecolor{_decoder}{RGB}{255, 204, 255} 
\definecolor{conv}{RGB}{255, 255, 255}
\definecolor{quant_color}{RGB}{252, 153, 162}
\definecolor{ae_color}{RGB}{221, 221, 221}
\definecolor{_rectangle}{RGB}{221, 221, 221}
\definecolor{source_color}{RGB}{200, 201, 164}
\definecolor{hyperprior_encoder_color}{RGB}{253, 211, 177}
\definecolor{hyper_color}{RGB}{189, 230, 241}
\definecolor{ce_color}{RGB}{240, 220, 120}
\definecolor{hyperprior_block_color}{RGB}{253, 187, 93}

\tikzstyle{process} = [rectangle,minimum width=5cm, minimum height=1cm, text
centered, draw=black, text=black,line width = 1pt, fill=conv]
 \tikzstyle{arrow} = [-Triangle, line width=1pt]
 \tikzstyle{label} = [text width= 0.2cm, align=center]
 \tikzstyle{waypoint}=[fill,circle,minimum size=4.5pt,inner sep=0pt]
 \tikzstyle{encoder} = [trapezium,
    trapezium angle=55,
    trapezium stretches=true,
    minimum width=7.95cm,
    minimum height=5.2cm,
    trapezium right angle=90,
    trapezium left angle=90,
    shape border rotate=180,
    text centered,
    ]

  \tikzstyle{decoder} = [trapezium,
    trapezium angle=55,
    trapezium stretches=true,
    minimum width=8cm,
    minimum height=4.6cm,
    trapezium right angle=85,
    trapezium left angle=85,
    line width=1.5pt,
    shape border rotate=180,
    text centered,
    draw=black]    

\tikzstyle{hyper_encoder} = [trapezium,
    trapezium angle=55,
    trapezium stretches=true,
    minimum width=2cm,
    minimum height=1.2cm,
    trapezium right angle=70,
    trapezium left angle=70,
    line width=1pt,
    shape border rotate=180,
    text centered,
    draw=black]


\begin{tikzpicture}[node distance = 1.5cm,font=\LARGE]

\coordinate (left_origin) at (-6.2,0);
\coordinate (right_origin) at (1.1,0);

\node (source_encoder) [draw, rectangle, minimum width=8.59cm, minimum height=5.65cm,  line width=0pt,draw=none, fill=ce_color,yshift=0cm]   at (left_origin) {};

\node (source_encoder_label) [draw, rectangle, minimum width=8.55cm, minimum height=1.7cm,  line width=0pt,draw=none, fill=ce_color,yshift=2.01cm, above of=source_encoder, text=black, font=\fontsize{20}{22}\selectfont]  { $\mathbf{Compression \ Encoder \ (CE)}$ };

 \node (quant_latent) [draw, rectangle, text=black, minimum width=1.55cm, minimum height=1.17cm,  line width=1.5pt, draw=black,yshift=-0.75cm,xshift=3.3cm,fill=quant_color,font=\fontsize{21}{23}\selectfont] at (source_encoder.north west)  {$\mathbf{Q^{\mathcal{R}}}$};
 
\node (way_quant)[waypoint,yshift=0cm, xshift=-0.55cm, left of= quant_latent]  {};

 \node (hyperprior_encoder) [draw, hyper_encoder, text=black, minimum width=2cm, minimum height=1.17cm,   line width=1.5pt,draw=black, fill=hyperprior_encoder_color,xshift=-2.05cm,yshift=-0.7cm,below of=quant_latent,font=\fontsize{21}{23}\selectfont]   {$\mathbf{HE}$};

 \node (quant_hyperprior) [draw, rectangle, text=black, minimum width=1.55cm, minimum height=1.17cm,  line width=1.5pt, draw=black,yshift=-0.45cm,xshift=2.05cm,below of=hyperprior_encoder,fill=quant_color,font=\fontsize{21}{23}\selectfont]  {$\mathbf{Q^{\mathcal{H}}}$};

 \node (ae_hyperprior) [draw, rectangle, text=black, minimum width=1.55cm, minimum height=1.17cm,  line width=1.5pt,draw=black, fill=ae_color,yshift=0cm,xshift=1.5cm,right of=quant_hyperprior,font=\fontsize{21}{23}\selectfont]   {$\mathbf{AE^{\mathcal{H}}}$};
 
\node (way_ae)[waypoint,yshift=0cm, xshift=-0.3cm, right of= ae_hyperprior]  {};
 
\node (hyperprior_block) [process, xshift=0cm, yshift=0.5cm, minimum width=2.3cm, minimum height=0.9cm, line width=1.5pt,fill=hyperprior_block_color, above of=ae_hyperprior,font=\fontsize{21}{23}\selectfont]     {$\mathbf{HD}$ {Block} };

\node (ae_latent) [draw, rectangle, text=black, minimum width=1.55cm, minimum height=1.17cm,  line width=1.5pt,draw=black, fill=ae_color,yshift=0cm,xshift=1.5cm,right of=quant_latent,font=\fontsize{21}{23}\selectfont]   {$\mathbf{AE^{\mathcal{R}}}$};

\draw[line width=1pt] ($([xshift=0cm]way_quant)+(-1.45,0cm)$)  to node[midway, right,xshift=0.0cm, yshift=0.5cm,font=\fontsize{21}{23}\selectfont] {$\mathbf{r} \in \mathcal{R}$}  (way_quant);
\draw[arrow] (way_quant) to   (quant_latent);
\draw[arrow] (way_quant) to   (hyperprior_encoder);
\draw[arrow] (hyperprior_encoder) to node[midway, right,xshift=0.1cm, yshift=0.33cm,font=\fontsize{21}{23}\selectfont]{$\mathbf{h} \in \mathcal{H}$}  ++(0,-1.95)   --  (quant_hyperprior.west);
\draw[arrow] (quant_hyperprior) to node[midway, right,xshift=-0.4cm, yshift=0.55cm,font=\fontsize{21}{23}\selectfont] {$\hat{\mathbf{h}} $}    (ae_hyperprior);
\draw[arrow] (way_ae) to node[midway, right,xshift=-1.1cm, yshift=0.01cm] {}  ++(0,0.95) --++(-1.19,0)  --     (hyperprior_block);
 \draw[arrow] (quant_latent) to node[midway, right,xshift=-0.3cm, yshift=0.48cm,font=\fontsize{21}{23}\selectfont] {$\hat{\mathbf{r}}$}    (ae_latent);
 \draw[arrow] ([xshift=1.09cm,yshift=0.45cm]hyperprior_block.west) to node[midway, right,xshift=-0.85cm, yshift=0.09cm,font=\fontsize{21}{23}\selectfont] {$\boldsymbol{\sigma}$}     ([xshift=0.32cm,yshift=0cm]ae_latent.south west);
\draw[arrow] ([xshift=-1.09cm,yshift=0.45cm]hyperprior_block.east) to node[midway, right,xshift=0.05cm, yshift=0.05cm,font=\fontsize{21}{23}\selectfont] {$\boldsymbol{\mu}$}  ([xshift=-0.33cm,yshift=0cm]ae_latent.south east);

\node (source_decoder) [draw, rectangle, minimum width=4.48cm, minimum height=5.65cm,  line width=0pt, fill=ce_color,yshift=0cm]   at (right_origin) {};
\node (source_decoder_label) [draw, rectangle, minimum width=4.3cm, minimum height=1.51cm,  line width=0pt,draw=none, fill=ce_color, xshift=0.cm,yshift=2.05cm, above of=source_decoder, text=black, font=\fontsize{20}{22}\selectfont]  {\shortstack{$\mathbf{Compression}$ \\ [0.05em] $\mathbf{Decoder \ (CD)} $}};
%
\node (sd_hyperprior_block) [process, xshift=0cm, yshift=2.8cm, minimum width=2.3cm, minimum height=0.9cm, line width=1.5pt,fill=hyperprior_block_color,font=\fontsize{21}{23}\selectfont]   at (source_decoder.south)  { $\mathbf{HD}$ {Block} };
 \node (sd_ae_latent) [draw, rectangle, text=black, minimum width=1.55cm, minimum height=1.17cm,  line width=1.5pt,draw=black, fill=ae_color,yshift=0.6cm,xshift=0cm,above of=sd_hyperprior_block,font=\fontsize{21}{23}\selectfont]   {$\mathbf{AD^{\mathcal{R}}}$};
 \draw[arrow] ([xshift=1.11cm,yshift=0.45cm]sd_hyperprior_block.west) to node[midway, right,xshift=-0.85cm, yshift=0.059cm,font=\fontsize{21}{23}\selectfont] {$\boldsymbol{\sigma}$}  ([xshift=0.36cm,yshift=0cm]sd_ae_latent.south west);
\draw[arrow] ([xshift=-1.11cm,yshift=0.45cm]sd_hyperprior_block.east) to node[midway, right,xshift=0.05cm, yshift=0.029cm,font=\fontsize{21}{23}\selectfont] {$\boldsymbol{\mu}$}   ([xshift=-0.36cm,yshift=0cm]sd_ae_latent.south east);

\draw[arrow] (ae_latent) to node[midway, right,xshift=-0.45cm, yshift=0.5cm,font=\fontsize{21}{23}\selectfont]{$\mathbf{b}_{\hat{\mathbf{r}}}$} (sd_ae_latent);
\draw[arrow] (ae_hyperprior) to node[midway, right,xshift=-0.94cm, yshift=0.5cm,font=\fontsize{21}{23}\selectfont] {$\mathbf{b}_{\hat{\mathbf{h}}}$} ++(5.31,0)   --   (sd_hyperprior_block);
%
\draw[arrow] (sd_ae_latent)  to node[midway, right,xshift=-0.5cm, yshift=0.5cm,font=\fontsize{21}{23}\selectfont] {$\hat{\mathbf{r}}$}  ($(sd_ae_latent)+(2.6,0cm)$);

\end{tikzpicture}

}
            \caption{\textbf{Proposed {\textsf{HSM} source codec} (with $\boldsymbol{\sigma}$ and $\boldsymbol{\mu}$) and baseline (with $\mathbf{\boldsymbol{\mu}}$)} network architecture with compression encoder $\mathbf{CE}$ and compression decoder $\mathbf{CD}$, utilized in Fig.\ \ref{fig:high_level_overview}. Details of the hyperprior encoder ($\mathbf{HE}$) are shown in Fig.\ \ref{fig:he_hsd_hmd}(a), details of the baseline and proposed hyperprior decoder ($\mathbf{HD}$) blocks are shown in Figs.\ \ref{fig:hyperprior_block}(a) and Figs.\ \ref{fig:hyperprior_block}(b), respectively.}
         \label{fig:mean_scale_codec}
    \end{wrapfigure}
 
    In Figure \ref{fig:high_level_overview}, we present a high-level overview of the state-of-the-art (SOTA) distributed semantic segmentation \cite{nazirjd,nazir2025efficient}, which also applies to our methods. The framework employs low-complex source codecs to enable bitrate-efficient communication between the edge device and the cloud. The image encoder $\mathbf{E}$ with parameters $\bm{\theta}^{\mathrm{E}}$, together with a source encoder is deployed on the edge device. The source encoder consists of a feature encoder $\mathbf{FE}$ with parameters $\bm{\theta}^{\mathrm{FE}}$, and a compression encoder $\mathbf{CE}$ with parameters $\bm{\theta}^{\mathrm{CE}}$. The feature encoder $\mathbf{FE}$ employs bottleneck features $\bf{z=E(x;\bm{\theta}^{\mathrm{E}})}$$\ \in \mathbb{R}^{F \times \frac{H}{d} \times \frac{W}{d}}$ with the number of kernels $F$, and a downsampling factor $d \in \mathbb{N}$ to compute the latent representation $\bf{r=FE(z;\bm{\theta}^{\mathrm{FE}})}$$\ \in \mathbb{R}^{F \times \frac{H}{d^{\prime}} \times \frac{W}{d^{\prime}}}$ with a downsampling factor $d^{\prime}=2d \in \mathbb{N}$. Here, $\bf{x} = (\bf{x}_{\textit{i}}) \in \mathbb{I}$$^{C \times H \times W}$ is a normalized image of $C\!=\!3$ color channels, height $H$ and width $W$, with pixel $\bf{x}_{\textit{i}} \in \mathbb{I}$$^C$, pixel index $i \in \mathcal{I}= \{ 1,...,H \cdot W \}$ and $\mathbb{I}=[0,1]$. The compression encoder $\mathbf{CE}$ receives $\mathbf{r}$ and generates a latent core bitstream $\mathbf{b}_{\mathbf{\hat{r}}}$ and enhancement bitstream $\mathbf{b}_{\mathbf{\hat{h}}}$, both jointly denoted as $\mathbf{b}\!=\!(\mathbf{b}_{\mathbf{\hat{r}}}, \mathbf{b}_{\mathbf{\hat{h}}})\!=\!\mathbf{CE}(\mathbf{r};\bm{\theta}^{\mathrm{CE}})$. In the cloud, a respective source decoder, consisting of a compression decoder $\mathbf{CD}$, is employed to reconstruct the latent representation $\hat{\mathbf{r}}=\mathbf{CD}(\mathbf{b};\bm{\theta}^{\mathrm{CD}})  \in \mathbb{R}^{F \times \frac{H}{d^{\prime}} \times \frac{W}{d^{\prime}}}$, which is utilized by the joint feature and task decoder $\mathbf{JD}$ to generate a semantic segmentation map $\mathbf{m} \in \mathcal{S}^{H \times W}$ with height $H$, width $W$, classes $\mathcal{S}=\{ 1,2, .., S \}$  and number of classes $S$. Note that the $\mathbf{FE}$ network architecture is adopted from \cite{ahuja2023neural} for compressing the bottleneck features $\mathbf{z}$ in the proposed source codecs described in the following subsection.

    \subsection{Proposed Source Codecs for Bottleneck Feature Compression}
    \label{subsec:proposed_source_codecs}

    \subsubsection{Hyperprior Standard Deviation and Mean (\textsf{HSM}) Source Codec}
    \label{subsubsec:hyp_ms_src_codec}

    Figure \ref{fig:mean_scale_codec} illustrates the detailed network architectures of the compression encoder $\mathbf{CE}$ and compression decoder $\mathbf{CD}$ utilized in our proposed hyperprior standard deviation and mean (\textsf{HSM}) source codec. Both $\mathbf{CE}$ and $\mathbf{CD}$ follow a hyperprior architecture, where the hyperprior is used to predict standard deviation and mean parameters for efficient entropy modeling of the latent representation $\mathbf{r}$ \cite{minnen2018joint,balle2018variational}. The compression encoder $\bf{CE}$ with parameters $\bm{\theta}^{\mathrm{CE}}$ receives the latent representation $\mathbf{r}\in\mathcal{R}$ with latent feature space $\mathcal{R} \in \mathbb{R}^{F \times \frac{H}{d^{\prime}} \times \frac{W}{d^{\prime}}}$ and passes it through a quantizer $\bf{Q}^{\mathcal{R}}$ to compute its quantized representation $\hat{\mathbf{r}} = \bf{Q}^{\mathcal{R}}(\mathbf{r})$, followed by an arithmetic encoder $\bf{AE}^{\mathcal{R}}$ \cite{AE} to generate a latent bitstream $\bf{b_{\hat{r}}=CE^{(r)}(r;\bm{\theta}^{\mathrm{CE}})}$. The symbols of $\bf{b_{\hat{r}}}$ typically exhibit significant spatial correlations, resulting in a suboptimal transmission bitrate \cite{balle2018variational}. Accordingly, an enhancement bitstream $\bf{b_{\hat{h}}}$ is introduced in $\mathbf{CE}$ to model the spatial dependencies and correlations in $\mathbf{b}_{\hat{\mathbf{r}}}$ through a learned standard deviation $\bf{\boldsymbol{\sigma}}$$\ \in \mathbb{R}^{F \times \frac{H}{d^{\prime}} \times \frac{W}{d^{\prime}}}$ and mean $\bf{\boldsymbol{\mu}}$$\ \in \mathbb{R}^{F \times \frac{H}{d^{\prime}} \times \frac{W}{d^{\prime}}}$. The hyperprior representation $\mathbf{h}=\mathbf{HE}(\mathbf{r};\bm{\theta}^{\mathrm{HE}}) \in \mathcal{H}$ with hyperprior feature space $\mathcal{H} \in \mathbb{R}^{F \times \frac{H}{e} \times \frac{W}{e}}$ is quantized through $\bf{Q}^{\mathcal{H}}$ and an arithmetic encoder $\mathbf{AE}^{\mathcal{H}}$ to compute the enhancement bitstream as $\bf{b_{\hat{h}}= CE^{(h)}(r;\bm{\theta}^{\mathrm{HE}})}$. Here, $e=2d^\prime \in \mathbb{N}$ denotes the downsampling factor, and $\mathbf{HE}$ represents a hyperprior encoder \cite{ahuja2023neural} with parameters $\bm{\theta}^{\mathrm{HE}}$. Fig.\ \ref{fig:he_hsd_hmd}(a) shows the detailed network architecture of $\mathbf{HE}$. It contains convolutional layers, which are represented by $\mathrm{Conv(\textit{h} \times \textit{h},\textit{F}, \rho)}$, where $\mathrm{\textit{h} \times \textit{h}}$ is the kernel size, $F$ is the number of kernels, and stride $\rho$.

\begin{wrapfigure}{r}{0.5\textwidth}
        \vspace{-1.2em} 
        \hspace{-2em} 
            \resizebox{1.1\linewidth}{!}{



\definecolor{encoder_decoder}{RGB}{213, 232, 212}
\definecolor{rectangle}{RGB}{255, 245, 204}
\definecolor{bottleneck_color}{RGB}{230, 221, 184}
\definecolor{_decoder}{RGB}{255, 204, 255} 
\definecolor{conv}{RGB}{255, 255, 255}
\definecolor{quant_color}{RGB}{232, 123, 142}
\definecolor{ae_color}{RGB}{221, 221, 221}
\definecolor{_rectangle}{RGB}{221, 221, 221}
\definecolor{source_color}{RGB}{200, 201, 164}
\definecolor{hyper_color}{RGB}{189, 230, 241}
\definecolor{ce_color}{RGB}{230, 221, 184}
\definecolor{hyperprior_block_color}{RGB}{253, 187, 93}

\tikzstyle{encoder} = [trapezium,
    trapezium angle=55,
    trapezium stretches=true,
    minimum width=7.95cm,
    minimum height=5.2cm,
    trapezium right angle=90,
    trapezium left angle=90,
    shape border rotate=180,
    text centered,
    ]

\tikzstyle{hyper_encoder} = [trapezium,
    trapezium angle=55,
    trapezium stretches=true,
    minimum width=2cm,
    minimum height=1.2cm,
    trapezium right angle=70,
    trapezium left angle=70,
    line width=1pt,
    shape border rotate=180,
    text centered,
    draw=black]
\tikzstyle{process} = [rectangle,minimum width=5cm, minimum height=1cm, text
centered, draw=black, text=black,line width = 1pt, fill=conv]
 \tikzstyle{arrow} = [-Triangle, line width=1pt]
 \tikzstyle{label} = [text width= 0.2cm, align=center]
 \tikzstyle{waypoint}=[fill,circle,minimum size=5.0pt,inner sep=0pt]


\begin{tikzpicture} [node distance = 1.5cm,font=\fontsize{27}{29}\selectfont]

\coordinate (left_origin) at (0,0);
\coordinate (right_origin) at (9.5,0);

\node (baseline_hd_block) [encoder ,minimum width=4.9cm, minimum height=4.1cm, xshift=-4cm, yshift=-0.35cm,  fill=hyperprior_block_color] at (left_origin) { } ;

\node (baseline_hb_ad) [draw, rectangle, text=black, minimum width=1.55cm, minimum height=1.17cm,  line width=1.5pt,draw=black, fill=ae_color,yshift=0.85cm,xshift=0.95cm] at (baseline_hd_block.south)    {$\mathbf{AD^{\mathcal{H}}}$};
 
\node (heading) [xshift=0.9cm,yshift=-1.3cm] at (baseline_hd_block.north west)  {\shortstack{ \!\!\!\!\!\!$\mathbf{HD}$ \\ [0.05em] $\mathbf{Block}$ }  };
  
\node (hb_hyperprior_decoder) [draw, hyper_encoder, text=black, minimum width=2.5cm, minimum height=1.17cm,   line width=1.5pt,draw=black, fill=hyper_color,xshift=0cm,yshift=0.95cm,above of=baseline_hb_ad]   {$\mathbf{HSD}$};

\node[align=center, anchor=north,xshift=0.4cm,yshift=-0.9cm, draw=none] 
  at ($(baseline_hd_block.south) + (0,-0.45)$)
  {\parbox{7.2cm}{
    (a) \textbf{Baseline} $\mathbf{HD}$ block \cite{nazirjd,nazir2025efficient}.
  }};

\draw[arrow] ($([xshift=0cm]baseline_hb_ad.south)+(0,-1.2cm)$)  to node[midway, right,xshift=0.1cm, yshift=-0.17cm,font=\fontsize{28}{30}\selectfont] {$\mathbf{b}_{\hat{\mathbf{h}}}$}  (baseline_hb_ad.south);

\draw[arrow] (baseline_hb_ad)  to node[midway, right,xshift=0.1cm, yshift=0.06cm] {$\hat{\mathbf{h}}$}  (hb_hyperprior_decoder);

 \draw[arrow] (hb_hyperprior_decoder)  to node[midway, right, xshift=0.1cm, yshift=0.12cm]  {$\boldsymbol{\sigma}$} ($([xshift=0cm]hb_hyperprior_decoder)+(0,1.6cm)$);

\node (proposed_hd_block) [encoder ,minimum width=8.5cm, minimum height=4.1cm, xshift=-4cm, yshift=-0.35cm,  fill=hyperprior_block_color] at (right_origin) { } ;

\node (hb_ad) [draw, rectangle, text=black, minimum width=1.55cm, minimum height=1.17cm,  line width=1.5pt,draw=black, fill=ae_color,yshift=0.85cm,xshift=-0.65cm] at (proposed_hd_block.south)    {$\mathbf{AD^{\mathcal{H}}}$};
 
\node (heading) [xshift=-0.9cm,yshift=-1.4cm] at (proposed_hd_block.north west)  {\shortstack{ \!\!\!\!\!\!$\mathbf{HD}$ \\ [0.05em] $\mathbf{Block}$ }  };
 
\node (way_ad)[waypoint,yshift=-0.75cm, xshift=0cm, above of= hb_ad]  {};
 
\node (hb_hyperprior_decoder) [draw, hyper_encoder, text=black, minimum width=2.5cm, minimum height=1.17cm,   line width=1.5pt,draw=black, fill=hyper_color,xshift=0cm,yshift=0.85cm,above of=hb_ad]   {$\mathbf{HSD}$};
 
\node (hb_hyperprior_mean_decoder) [draw, hyper_encoder, text=black, minimum width=2cm, minimum height=1.17cm,   line width=1.5pt,draw=black, fill=hyper_color,xshift=1.8cm,yshift=0cm,right of=hb_hyperprior_decoder]   {$\mathbf{HMD}$};

\node[align=center, anchor=north,xshift=0cm,yshift=-0.9cm, draw=none] 
  at ($(proposed_hd_block.south) + (0,-0.45)$)
  {\parbox{9.6cm}{
    (b) \textbf{Proposed} $\mathbf{HD}$ block architecture.
  }};

\draw[arrow] ($([xshift=0cm]hb_ad.south)+(0,-1.2cm)$)  to node[midway, right,xshift=0.1cm, yshift=-0.17cm,font=\fontsize{28}{30}\selectfont] {$\mathbf{b}_{\hat{\mathbf{h}}}$}  (hb_ad.south);

\draw[arrow] (hb_ad)  to node[midway, right,xshift=0.1cm, yshift=0.11cm] {$\hat{\mathbf{h}}$}  (hb_hyperprior_decoder);

\draw[arrow] (way_ad)  to node[midway, right,xshift=0.1cm, yshift=0.15cm] {} ++(3.3,0)   --    (hb_hyperprior_mean_decoder);

 \draw[arrow] (hb_hyperprior_decoder)  to node[midway, right, xshift=0.1cm, yshift=0.12cm]  {$\boldsymbol{\sigma}$} ($([xshift=0cm]hb_hyperprior_decoder)+(0,1.6cm)$); 

 \draw[arrow] (hb_hyperprior_mean_decoder)  to node[midway, right, xshift=0.1cm, yshift=0.16cm]  {$\boldsymbol{\mu}$} ($([xshift=0cm]hb_hyperprior_mean_decoder)+(0,1.6cm)$); 

\end{tikzpicture}

}
            \caption{\textbf{Details of hyperprior decoder ($\mathbf{HD}$) architectures} for distributed semantic segmentation. Fig.\ \ref{fig:hyperprior_block}(a) shows the baseline hyperprior decoder ($\mathbf{HD}$), taken from \cite{nazirjd,nazir2025efficient}. Fig.\ \ref{fig:hyperprior_block}(b) illustrates the proposed $\mathbf{HD}$ architecture block, utilized in Fig.\ \ref{fig:mean_scale_codec} (\textsf{HSM}) and Fig.\ \ref{fig:litcm_codec} (\textsf{AR-HSM}). Details of the hyperprior standard deviation decoder ($\mathbf{HSD}$) and the hyperprior mean decoder ($\mathbf{HMD}$) are shown in Fig.\ \ref{fig:he_hsd_hmd}(b).}
         \label{fig:hyperprior_block}
    \end{wrapfigure}
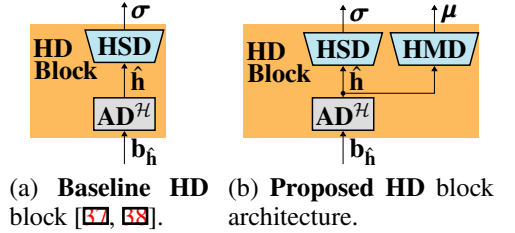
 
In Figure \ref{fig:hyperprior_block}, we show the detailed network architecture of the proposed hyperprior decoder $\mathbf{HD}$ against the baseline $\mathbf{HD}$ block \cite{nazirjd,nazir2025efficient}. Fig.\ \ref{fig:hyperprior_block}(a) illustrates that the baseline $\mathbf{HD}$ block receives the enhancement bitstream $\bf{b}_{\hat{h}}$ and passes it through an arithmetic decoder $\mathbf{AD}^{\mathcal{H}}$ to reconstruct the quantized hyperprior representation $\hat{\mathbf{h}}$, which is sent to the hyperprior standard deviation decoder $\mathbf{HSD}$ with parameters $\bm{\theta}^{\mathrm{HSD}}$ shown in Fig.\ \ref{fig:he_hsd_hmd}(b) to predict $\boldsymbol{\sigma}$. It employs only $\mathbf{HSD}$ and assumes that $\hat{\mathbf{r}}$ is zero mean and follows a Gaussian distribution, resulting in a suboptimal $\mathbf{b}_{\hat{\mathbf{h}}}$ \cite{minnen2018joint}. \textit{In contrast to the baseline $\mathbf{HD}$ block (cf. Fig.\ \ref{fig:hyperprior_block}(a)), our proposed $\mathbf{HD}$ block (cf. Fig.\ \ref{fig:hyperprior_block}(b)) does not assume that $\hat{\mathbf{r}}$ follows a zero-mean Gaussian distribution, since we employ a separate hyperprior mean decoder $\mathbf{HMD}$ to model the mean $\boldsymbol{\mu}$ of $\hat{\mathbf{r}}$ from $\hat{\mathbf{h}}$ besides the $\mathbf{HSD}$ (for $\boldsymbol{\sigma}$ prediction)}. Although, the introduction of $\mathbf{HMD}$ in our proposed $\mathbf{HD}$ block slightly increases the computational complexity and the number of learnable parameters, it leads to a significant improvement in the entropy modeling of $\hat{\mathbf{r}}$ and achieves an accordingly better RD trade-off. Through $\mathbf{HSD}$ and $\mathbf{HMD}$, each element in $\bf{\hat{r}}$ can now be modeled with its individual learned standard deviation $\sigma_{j}$ and mean $\mu_{j}$ as $\boldsymbol{\sigma}=(\sigma_{\textit{j}})=\mathbf{HSD}(\hat{\mathbf{h}};\bm{\theta}^{\mathrm{HSD}})$  
    and $\boldsymbol{\mu}=(\mu_{\textit{j}})=\mathbf{HMD}(\hat{\mathbf{h}};\bm{\theta}^{\mathrm{HMD}})$ with quantized latent representation element index $j \in \mathcal{J}=\{ 1, \cdots, \frac{H}{d^\prime} \cdot \frac{W}{d^\prime} \}$ and $\mathbf{HMD}$ parameters $\bm{\theta}^{\mathrm{HMD}}$. 
 
    \begin{wrapfigure}{l}{0.5\textwidth}
            \hspace{-1.8em} 
                \resizebox{1.1\linewidth}{!}{
 



\definecolor{encoder_decoder}{RGB}{213, 232, 212}
\definecolor{rectangle}{RGB}{255, 245, 204}
\definecolor{bottleneck_color}{RGB}{230, 221, 184}
\definecolor{_decoder}{RGB}{255, 204, 255} 
\definecolor{conv}{RGB}{255, 255, 255}
\definecolor{quant_color}{RGB}{232, 123, 142}
\definecolor{ae_color}{RGB}{221, 221, 221}
\definecolor{_rectangle}{RGB}{221, 221, 221}
\definecolor{source_color}{RGB}{200, 201, 164}

\definecolor{hyperprior_encoder_color}{RGB}{253, 211, 177}
\definecolor{hyper_color}{RGB}{189, 230, 241}

\definecolor{ce_color}{RGB}{230, 221, 184}

\tikzstyle{encoder} = [trapezium,
    trapezium angle=55,
    trapezium stretches=true,
    minimum width=7.95cm,
    minimum height=5.2cm,
    trapezium right angle=90,
    trapezium left angle=90,
    shape border rotate=180,
    text centered,
    ]

\tikzstyle{process} = [rectangle,minimum width=5cm, minimum height=1cm, text
centered, draw=black, text=black,line width = 1pt, fill=conv]
 \tikzstyle{arrow} = [-Triangle, line width=1pt]
 \tikzstyle{label} = [text width= 0.2cm, align=center]
 \tikzstyle{waypoint}=[fill,circle,minimum size=3.5pt,inner sep=0pt]


\begin{tikzpicture} [node distance = 1.5cm,font=\fontsize{24}{26}\selectfont]

\coordinate (hyperprior_encoder) at (-9.05,0);
\coordinate (mean_encoder) at (1.3,0);

\node (hyperprior_encoder) [encoder ,minimum width=7.75cm, minimum height=5.55cm, draw=none,xshift=0cm, yshift=-0.35cm,  fill=hyperprior_encoder_color] at (hyperprior_encoder) { } ;
\node (heading) [above of=hyperprior_encoder,xshift=-2.99cm,yshift=0.8cm,font=\fontsize{28}{30}\selectfont]  {$\mathbf{HE}$};
\node (he_conv1) [process, yshift=-1.45cm,minimum width=6cm,minimum height=0.9cm] at (hyperprior_encoder.north) {$\mathrm{Conv(3 \times 3, }F, \rho=2)$};
 \node (he_batchnorm) [process,minimum width=7.3cm,yshift=0.33cm, minimum height=0.9cm,below of=he_conv1] {$\mathrm{BatchNorm + ReLU}$};
\node (he_conv2) [process, yshift=0.365cm,minimum width=7.3cm,minimum height=0.85cm,below of=he_batchnorm]  {$\mathrm{Conv(1 \times 1, }F)$};
\node (he_batchnorm2) [process,minimum width=7.3cm,yshift=0.365cm, minimum height=0.9cm,below of=he_conv2]  { $\mathrm{BatchNorm + ReLU}$};
\node (input_label) [below of = he_batchnorm2,xshift = -1.55cm,yshift=0.2cm,font=\fontsize{22}{24}\selectfont]  {$F \! \times \! \frac{H}{e} \! \times \frac{W}{e} $};
\node (output_label) [above of = he_conv1,xshift = -1.57cm,yshift=0.5cm,font=\fontsize{22}{24}\selectfont ]  {$F \! \times \! \frac{H}{d^{\prime}} \! \times \frac{W}{d^{\prime}} $};
\draw[arrow] (he_batchnorm2.south)  to node[midway, right,xshift=0.1cm, yshift=-0.05cm, font=\fontsize{29}{31}\selectfont] {$\mathbf{h}$}  ($([xshift=0cm]he_batchnorm2.south)+(0,-1.4cm)$);
\draw[arrow] ($([xshift=0cm]he_conv1)+(0,2.6cm)$)  to node[midway, right, xshift=0.1cm, yshift=0.35cm,font=\fontsize{29}{31}\selectfont]  {$\mathbf{r}$} (he_conv1); 
  \node[align=center, anchor=north,xshift=0cm,yshift=-0.8cm, font=\fontsize{28}{30}\selectfont] 
   at ($(hyperprior_encoder.south) + (0,-0.45)$)
   {\parbox{8.0cm}{
     (a) \textbf{Hyperprior encoder} ($\mathbf{HE}$)
  }};

 \node (mean_encoder) [encoder ,minimum width=8.7cm, minimum height=6.35cm, xshift=0cm, yshift=-0.05cm,  fill=hyper_color] at (mean_encoder) { } ;
  
\node (heading) [above of=mean_encoder,xshift=-3.1cm,yshift=0.82cm,font=\fontsize{28}{30}\selectfont]  {\shortstack{\!\!\!$\mathbf{HSD \ and}$\!\!\!\!\!\!\!\!\!\!\!\!\!\\[0.1em]~$\mathbf{HMD}$}};
\node (me_batchnorm2) [process,minimum width=8.3cm,yshift=0.73cm, minimum height=0.91cm] at (mean_encoder.south)   {$\mathrm{UpConv(1 \times 1, }F)$ };
\node (me_conv2) [process, yshift=-0.365cm,minimum width=8.3cm,minimum height=0.85cm,above of=me_batchnorm2]  {$\mathrm{BatchNorm + ReLU}$};
 \node (me_batchnorm) [process,minimum width=7.3cm,yshift=0.77cm, minimum height=0.9cm,above of=me_batchnorm2] { $\mathrm{UpConv(3 \times 3, }F, \rho=2)$ };
\node (me_conv1) [process, yshift=-0.335cm,minimum width=8.3cm,minimum height=0.9cm, above of=me_batchnorm]  {$\mathrm{BatchNorm + ReLU}$};

\node (input_label) [below of = me_batchnorm2,xshift = -1.55cm,yshift=0.2cm,font=\fontsize{22}{24}\selectfont]  {$F \! \times \! \frac{H}{e} \! \times \frac{W}{e} $};
\node (output_label) [above of = me_conv1,xshift = -1.65cm,yshift=1.25cm, font=\fontsize{22}{24}\selectfont ]  {$F \! \times \! \frac{H}{d^{\prime}} \! \times \frac{W}{d^{\prime}} $};
\draw[arrow] ($([xshift=0cm]me_batchnorm2.south)+(0,-1.4cm)$)  to node[midway, right,xshift=0.1cm, yshift=-0.1cm,font=\fontsize{29}{31}\selectfont] {$\hat{\mathbf{h}}$}  (me_batchnorm2.south);
 \draw[arrow] (me_conv1)  to node[midway, right, xshift=0.1cm, yshift=0.8cm, font=\fontsize{29}{31}\selectfont]  {$\mathbf{g}$} ($([xshift=0cm]me_conv1)+(0,3.3cm)$); 
\node[align=center,draw=none, anchor=north,xshift=-0.2cm,yshift=-0.8cm,font=\fontsize{28}{30}\selectfont] 
  at ($(mean_encoder.south) + (0,-0.45)$)
  {\parbox{10.5cm}{
    (b) \textbf{Hyperprior standard deviation and mean decoders} ($\mathbf{HSD}$, $\mathbf{HMD}$)
  }};

\end{tikzpicture}

}
                \caption{Network architecture of the hyperprior encoder $\mathbf{HE}$, hyperprior standard deviation decoder $\mathbf{HSD}$ (both adopted from \cite{ahuja2023neural}), and the \textbf{proposed} hyperprior mean decoder $\mathbf{HMD}$, used in Figs.\ \ref{fig:mean_scale_codec} and \ref{fig:litcm_codec}.}
              \label{fig:he_hsd_hmd}
    \end{wrapfigure}
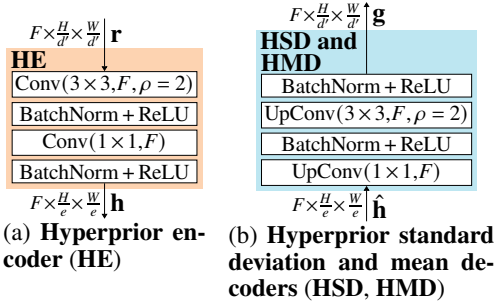  
    Fig.\ \ref{fig:he_hsd_hmd}(b) presents the detailed network architectures of $\mathbf{HSD}$ \cite{ahuja2023neural} and the proposed $\mathbf{HMD}$. For simplicity, the network architectures of $\mathbf{HSD}$ and $\mathbf{HMD}$ are kept identical. Both contain transposed convolutional layers, which are represented by $\mathrm{UpConv(\textit{h} \times \textit{h},\textit{F}, \rho)}$, where $\mathrm{\textit{h} \times \textit{h}}$ is the kernel size, $F$ is the number of kernels, and stride $\rho$. Further, the proposed $\mathbf{HMD}$ outputs $\mathbf{g}\!=\!\bm{\mu}$, while the $\mathbf{HSD}$ outputs $\mathbf{g}=\boldsymbol{\sigma}$. The compression decoder $\mathbf{CD}$ with parameters $\bm{\theta}^{\mathrm{CD}}$ reconstructs the latent representation through $\bf{AD}^{\mathcal{R}}$ and $\bf{AD}^{\mathcal{H}}$ \cite{AE}, along with $\bf{HMD}$ and $\bf{HSD}$ as $\bf{\hat{r}}=\bf{CD}(\bf{b_{\hat{r}}},\bf{b_{\hat{h}}};\bm{\theta}^{\mathrm{CD}})$. Although the proposed \textsf{HSM} source codec improves the entropy modeling of the latent representation $\mathbf{r}$ by modeling its spatial dependencies through  $\boldsymbol{\mu}$, and $\boldsymbol{\sigma}$, it does not explicitly capture the inter-channel dependencies within $\mathbf{r}$. To address this limitation, we propose the \textsf{AR-HSM} source codec described in the following.
\subsubsection{Autoregressive Hyperprior Standard Deviation and Mean (\textsf{AR-HSM}) Source Codec}
\label{subsubsec:ar_hyp_ms_src_codec}

\begin{figure*}[t!]
      \hspace{-1.em} 
      \resizebox{0.92\linewidth}{!}{\input{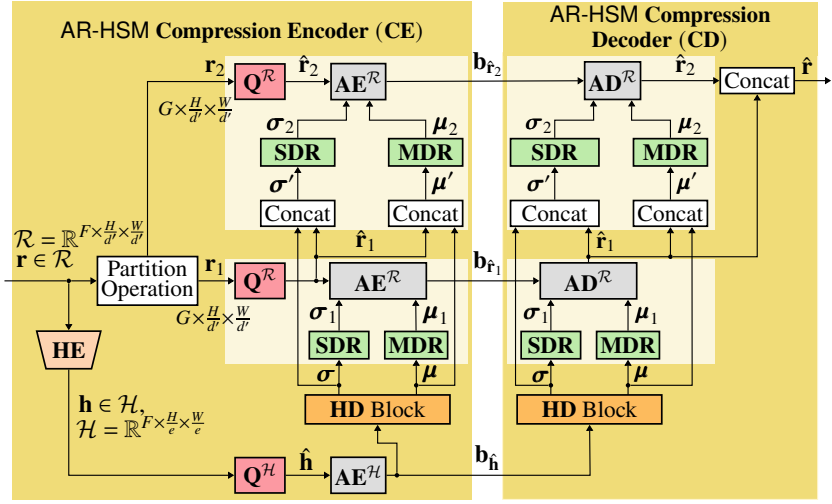}}
        \caption{ \textbf{Proposed} \textbf{\textsf{AR-HSM} source codec} network architecture with compression encoder $\mathbf{CE}$ and compression decoder $\mathbf{CD}$, utilized in Fig.\ \ref{fig:high_level_overview}. Details of the hyperprior encoder ($\mathbf{HE}$) are shown in Fig.\ \ref{fig:he_hsd_hmd}(a), details of the \textbf{conventional} and \textbf{proposed} hyperprior decoder ($\mathbf{HD}$) blocks are shown in Figs.\ \ref{fig:hyperprior_block}(a) and Figs.\ \ref{fig:hyperprior_block}(b), respectively. Further details of the standard deviation refinement block $\mathbf{SDR}$ and mean refinement block $\mathbf{MDR}$ are shown in Fig.\ \ref{fig:sr_mr}. }
        \label{fig:litcm_codec}
    \end{figure*}

    Inspired by the recent learned image compression source codec \cite{liu2023learned}, we propose an \textit{autoregressive} hyperprior mean and standard deviation (\textsf{AR-HSM}) source codec for bottleneck feature compression, where standard deviation and mean parameters from the hyperprior are used to perform \textit{channel-wise autoregressive entropy modeling of the latent representation $\mathbf{r} \in \mathbb{R}^{F \times \frac{H}{d^{\prime}} \times \frac{W}{d^{\prime}}}$ to capture the inter-channel dependencies within $\mathbf{r}$ and enable very low bit\-rates, while further improving the rate-distortion performance of the distributed semantic segmentation approaches}. Different to the learned image compression source codec \cite{liu2023learned}, which employs a mixed high-complexity CNN and transformer-based network architecture for the $\mathbf{HD}$ block, we propose a low-complexity $\mathbf{HD}$ block (cf. Fig.\ \ref{fig:hyperprior_block}(a)) specifically designed for distributed semantic segmentation. Fig.\ \ref{fig:litcm_codec} illustrates the detailed network architectures of compression encoder $\mathbf{CE}$ and compression decoder $\mathbf{CD}$ employed in the proposed \textsf{AR-HSM} source codec. Similar to the \textsf{HSM} source codec described in Section \ref{subsubsec:hyp_ms_src_codec}, both, $\mathbf{CE}$ and $\mathbf{CD}$, follow a hyperprior architecture for entropy modeling of $\mathbf{r}$.
    The compression \textit{encoder} $\mathbf{CE}$ with parameters $\bm{\theta}^{\mathrm{CE}}$ in Figure \ref{fig:litcm_codec} (left) receives the latent representation $\mathbf{r}$ and applies a $\mathrm{partition \ operation}$, which partitions $\mathbf{r}$ along the channel dimension into $K \in \mathbb{N}$ equally-sized partitions containing consecutive channels. The $\mathrm{partition \ operation}$ is defined as $\mathcal{V}: \mathbb{R}^{F \times \frac{H}{d^{\prime}} \times \frac{W}{d^{\prime}}} \rightarrow \mathbb{R}^{K \times G \times \frac{H}{d^{\prime}} \times \frac{W}{d^{\prime}}}$ and $\mathcal{V}_{k} (\mathbf{r}) = (\mathbf{r}_{k})$ with partition $\mathbf{r}_{k} \in \mathbb{R}^{G \times \frac{H}{d^{\prime}} \times \frac{W}{d^{\prime}}}$, partition index $k \in \mathcal{K}=\{1,\dots,K\}$, number of partitions $K=2$, and number of kernels in a partition $G=\frac{F}{K}$. Note that different to \cite{liu2023learned}, we do not employ latent residual blocks in the proposed source codec, as they are only beneficial if the number of partitions fulfills $K\gg2$. The first partition $\mathbf{r}_{1}$ is quantized using $\mathbf{Q}^{\mathcal{R}}$ to compute its quantized representation $\hat{\mathbf{r}}_{1}=\mathbf{Q}^{\mathcal{R}}(\mathbf{r}_{1})$, which is passed along with the refined standard deviation $\boldsymbol{\sigma}_1=\mathbf{SDR}(\boldsymbol{\sigma};\bm{\theta}_{1}^{\mathrm{SDR}}) \in \mathbb{R}^{G \times \frac{H}{d^{\prime}} \times \frac{W}{d^{\prime}}}$, and the mean $\boldsymbol{\mu}_1\!=\!\mathbf{MDR}(\boldsymbol{\mu};\bm{\theta}_{1}^{\mathrm{MDR}}) \in \mathbb{R}^{G \times \frac{H}{d^{\prime}} \times \frac{W}{d^{\prime}}}$ parameters \cite{liu2023learned} to an arithmetic encoder $\mathbf{AE}^{\mathcal{R}}$ \cite{AE} to generate the first partition bitstream $\mathbf{b}_{\hat{\mathbf{r}}_1}=\mathbf{CE}^{(\hat{\mathbf{r}}_1)}(\mathbf{r}_1; \bm{\theta}^{\mathrm{CE}})$. Here, $\mathbf{SDR}$ and $\mathbf{MDR}$ denote the standard deviation and mean refinement blocks with parameters $\bm{\theta}_{1}^{\mathrm{SDR}}$ and $\bm{\theta}_{1}^{\mathrm{MDR}}$, respectively. Further, $\boldsymbol{\sigma} \in \mathbb{R}^{F_1 \times \frac{H}{d^{\prime}} \times \frac{W}{d^{\prime}}}$ and $\boldsymbol{\mu} \in \mathbb{R}^{F_1 \times \frac{H}{d^{\prime}} \times \frac{W}{d^{\prime}}}$ with number of kernels $F_1$, represent the standard deviation and mean parameters produced by the proposed $\mathbf{HD}$ block for $\mathbf{r}_1$, respectively, cf. Fig.\ \ref{fig:hyperprior_block}(b).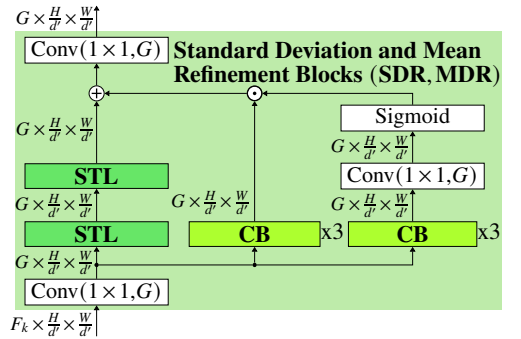
\begin{wrapfigure}{r}{0.5\textwidth}
            \hspace{-2.5em} 
                \resizebox{1.13\linewidth}{!}{\usetikzlibrary{shapes.geometric, arrows.meta, calc,backgrounds}

\definecolor{encoder_decoder}{RGB}{213, 232, 212}
\definecolor{rectangle}{RGB}{255, 245, 204}
\definecolor{bottleneck_color}{RGB}{230, 221, 184}
\definecolor{_decoder}{RGB}{255, 204, 255} 
\definecolor{conv}{RGB}{255, 255, 255}
\definecolor{quant_color}{RGB}{232, 123, 142}
\definecolor{ae_color}{RGB}{221, 221, 221}
\definecolor{_rectangle}{RGB}{221, 221, 221}
\definecolor{source_color}{RGB}{200, 201, 164}
\definecolor{proposed}{RGB}{103, 230, 103}
\definecolor{hyper_color}{RGB}{255, 219, 77}
\definecolor{lsd_color}{RGB}{194, 245, 193}
\definecolor{ce_color}{RGB}{230, 221, 184}
\definecolor{cb}{RGB}{173,255,47}
\definecolor{latent_block_color}{RGB}{190, 235, 170}

\tikzstyle{encoder} = [trapezium,
    trapezium angle=55,
    trapezium stretches=true,
    minimum width=7.95cm,
    minimum height=5.2cm,
    trapezium right angle=90,
    trapezium left angle=90,
    shape border rotate=180,
    text centered,
    ]

\tikzstyle{process} = [rectangle,minimum width=5cm, minimum height=1cm, text
centered, draw=black, text=black,line width = 1pt, fill=conv]
\tikzstyle{arrow} = [-Triangle, line width=1pt]
\tikzstyle{label} = [text width= 0.2cm, align=center]
\tikzstyle{waypoint}=[fill,circle,minimum size=5.0pt,inner sep=0pt]
\tikzstyle{dots}=[fill,circle,minimum size=3pt,inner sep=0pt]

\newcommand{\CircularPlus}{%
    \tikz[baseline=0.25ex, line width=1.5, scale=0.2]{
        \draw[black, fill=black] (0,-1) -- (0,1);  
        \draw[black, fill=black] (-1,0) -- (1,0);  
    }%
}

\newcommand{\CircularOdot}{
  \tikz[baseline=0.25ex, scale=0.24]{
    \draw[line width=0.5] (0,0) circle (1.2);
    \fill (0,0) circle (0.3);
  }
}


\begin{tikzpicture} [node distance = 1.5cm,font=\fontsize{27}{29}\selectfont]

\coordinate (hyperprior_encoder) at (0,0);

\node (latent_decoder) [encoder ,minimum width=19.7cm, minimum height=11.3cm,  fill=latent_block_color] at (hyperprior_encoder) { } ;

\node (heading) [above of=hyperprior_encoder,xshift=2.8cm,yshift=3.4cm, font=\fontsize{27.5}{29.5}\selectfont]  {$\mathbf{Standard \ Deviation \ and \ Mean }$};

\node (sub_heading) [below of=heading,xshift=0.4cm,yshift=0.45cm, font=\fontsize{27.5}{29.5}\selectfont]  {$\mathbf{Refinement \ Blocks \ (SDR,MDR)}$};

\node (latent_conv1) [process, xshift=-0.9cm,yshift=0.75cm,minimum width=5.8cm,minimum height=0.9cm,font=\fontsize{25.5}{27.5}\selectfont] at (latent_decoder.south west) {$\mathrm{Conv(1 \times 1, }G)$};

\node (swin_transformer_layer) [process, xshift=0cm,yshift=0.85cm,minimum width=5.8cm,minimum height=1.0cm, above of=latent_conv1, fill=proposed]  {$\mathbf{STL}$ };

\node (swin_transformer_layer_2) [process, xshift=0cm,yshift=0.85cm,minimum width=5.8cm,minimum height=1.0cm, above of=swin_transformer_layer,fill=proposed]  {$\mathbf{STL}$   };

\node (plus) [draw, circle, minimum size=0.1cm,yshift=1.85cm,fill=conv,above of=swin_transformer_layer_2, line width=1,inner sep=-0.3pt] {\CircularPlus};

\node (output_conv) [process, xshift=0cm,yshift=0.25cm,minimum width=5.8cm,minimum height=0.9cm, above of=plus,font=\fontsize{25.5}{27.5}\selectfont] {$\mathrm{Conv(1 \times 1, }G)$};

\node (convolutional_block_1) [process, xshift=4.9cm,yshift=0.0cm,minimum width=5.3cm,minimum height=1.0cm, right of=swin_transformer_layer, fill=cb]  {$\mathbf{CB}$ };

\node (cross) [draw, circle, minimum size=0.05cm,yshift=4.2cm,fill=conv,above of=convolutional_block_1, line width=0.8,inner sep=-2.4pt] {\CircularOdot};

\node (convolutional_block_4) [process, xshift=4.9cm,yshift=0cm,minimum width=5.2cm,minimum height=1.0cm, right of=convolutional_block_1, fill=cb]  {$\mathbf{CB}$};

\node (label) [right of=convolutional_block_1,xshift=1.65cm,yshift=0.1cm]  {$\mathrm{x3}$};
\node (label) [right of=convolutional_block_4,xshift=1.6cm,yshift=0.1cm]  {$\mathrm{x3}$};



\node (2nd_conv) [process, xshift=0cm,yshift=0.85cm,minimum width=5.8cm,minimum height=0.9cm, above of=convolutional_block_4,font=\fontsize{25.5}{27.5}\selectfont] {$\mathrm{Conv(1 \times 1, }G)$};

\node (sigmoid) [process, xshift=0cm,yshift=0.85cm,minimum width=5.8cm,minimum height=1.0cm, above of=2nd_conv,font=\fontsize{25.5}{27.5}\selectfont] {$\mathrm{Sigmoid}$};

\node (way_layer)[waypoint,yshift=-0.4cm, xshift=0cm, above of= latent_conv1]  {};

\node (way_block1)[waypoint,yshift=0.25cm, xshift=0cm, below of= convolutional_block_1]  {};

\draw[arrow] ($(latent_conv1.south)+(0,-1.25cm)$)  to node[midway, right,xshift=-3.65cm, yshift=-0.15cm,font=\fontsize{22}{24}\selectfont] {$F_k \times \frac{H}{d^{\prime}} \times \frac{W}{d^{\prime}}$}  (latent_conv1.south);
\draw[arrow] (latent_conv1)  to node[midway, right,xshift=-3.45cm, yshift=0cm,font=\fontsize{22}{24}\selectfont] {$G \times \frac{H}{d^{\prime}} \times \frac{W}{d^{\prime}}$}  (swin_transformer_layer);
\draw[arrow] (swin_transformer_layer)  to node[midway, right,xshift=-3.45cm, yshift=0cm,font=\fontsize{22}{24}\selectfont] {$G \times \frac{H}{d^{\prime}} \times \frac{W}{d^{\prime}}$}  (swin_transformer_layer_2);
\draw[arrow] (swin_transformer_layer_2)  to node[midway, right,xshift=-3.4cm, yshift=0.1cm,font=\fontsize{22}{24}\selectfont] {$G \times \frac{H}{d^{\prime}} \times \frac{W}{d^{\prime}}$}  (plus);
\draw[arrow] (plus)  to node[midway, right,xshift=0.1cm, yshift=0cm,font=\LARGE] {}  (output_conv);
\draw[line width=1] (way_layer)  to node[midway, right,xshift=0.1cm, yshift=0cm,font=\LARGE] {}  (way_block1);
\draw[arrow] (way_block1)  to node[midway, right,xshift=0.1cm, yshift=0cm,font=\LARGE] {}  (convolutional_block_1);
\draw[arrow] (convolutional_block_1)  to node[midway, left,xshift=-0.08cm, yshift=-1.67cm,font=\fontsize{22}{24}\selectfont] {$G \times \frac{H}{d^{\prime}} \times \frac{W}{d^{\prime}}$}  (cross);
\draw[arrow] (convolutional_block_4) to node[midway, right,xshift=-3.45cm, yshift=0cm,font=\fontsize{22}{24}\selectfont] {$G \times \frac{H}{d^{\prime}} \times \frac{W}{d^{\prime}}$}  (2nd_conv);
\draw[arrow] (way_block1)  to node[midway, right,xshift=0.1cm, yshift=0cm,font=\LARGE] {} ++(6.40,0)   --  (convolutional_block_4.south);
\draw[arrow] (2nd_conv)  to node[midway, right,xshift=-3.45cm, yshift=0cm,font=\fontsize{22}{24}\selectfont] {$G \times \frac{H}{d^{\prime}} \times \frac{W}{d^{\prime}}$}  (sigmoid);
\draw[arrow] (sigmoid)  to node[midway, right,xshift=0.1cm, yshift=0cm,font=\LARGE] {} ++(0,1.)   --  (cross.east);
\draw[arrow] (cross) to (plus);
\draw[arrow] (output_conv)  to node[midway, right, xshift=-3.4cm, yshift=0.1cm, font=\fontsize{22}{24}\selectfont]  {$G \times \frac{H}{d^{\prime}} \times \frac{W}{d^{\prime}}$} ($([xshift=0cm]output_conv)+(0,1.8cm)$);

\end{tikzpicture}

}
                \caption{Standard deviation and mean refinement blocks ($\mathbf{SDR}$, $\mathbf{MDR}$) used in Fig.\ \ref{fig:litcm_codec}, adopted from \cite{liu2023learned}. \texttt{Swin} transformer layer ($\mathbf{STL}$) and convolutional block ($\mathbf{CB}$) are detailed in Figs.\ \ref{fig:stl_cb}(a) and (b), respectively.}
             \label{fig:sr_mr}
    \end{wrapfigure} Note that the channel-wise autoregressive entropy modeling of $\mathbf{r}$ increases the computational complexity and the number of learnable parameters compared to the baseline \cite{nazirjd,nazir2025efficient} and \textsf{HSM} source codecs.\begin{wrapfigure}{l}{0.5\textwidth}
            \hspace{-2.1em} 
                \resizebox{1.13\linewidth}{!}{\usetikzlibrary{shapes.geometric, arrows.meta, calc,backgrounds}

\definecolor{encoder_decoder}{RGB}{213, 232, 212}
\definecolor{rectangle}{RGB}{255, 245, 204}
\definecolor{bottleneck_color}{RGB}{230, 221, 184}
\definecolor{_decoder}{RGB}{255, 204, 255} 
\definecolor{conv}{RGB}{255, 255, 255}

\definecolor{quant_color}{RGB}{232, 123, 142}
\definecolor{ae_color}{RGB}{221, 221, 221}
\definecolor{_rectangle}{RGB}{221, 221, 221}
\definecolor{source_color}{RGB}{200, 201, 164}
\definecolor{hyper_color}{RGB}{255, 219, 77}
\definecolor{ce_color}{RGB}{230, 221, 184}
\definecolor{proposed}{RGB}{103, 230, 103}

\definecolor{cb}{RGB}{173,255,47}

\tikzstyle{encoder} = [trapezium,
    trapezium angle=55,
    trapezium stretches=true,
    minimum width=7.95cm,
    minimum height=5.2cm,
    trapezium right angle=90,
    trapezium left angle=90,
    shape border rotate=180,
    text centered,
    ]

\tikzstyle{process} = [rectangle,minimum width=5cm, minimum height=1cm, text
centered, draw=black, text=black,line width = 1pt, fill=conv]
 \tikzstyle{arrow} = [-Triangle, line width=1pt]
 \tikzstyle{label} = [text width= 0.2cm, align=center]
 \tikzstyle{waypoint}=[fill,circle,minimum size=3.5pt,inner sep=0pt]


\begin{tikzpicture} [node distance = 1.5cm,font=\fontsize{25}{27}\selectfont]

\node (stl) [encoder ,minimum width=7.3cm, minimum height=5.7cm, xshift=0cm, yshift=-0.35cm,  fill=proposed]{ } ;

\node (heading) [above of=stl,xshift=-2.5cm,yshift=0.9cm]  {$\mathbf{STL}$};

\node (window_transformer_blk) [process, yshift=1.03cm,minimum width=4.5cm,minimum height=0.9cm,font=\fontsize{24.5}{26.5}\selectfont] at (stl.south) {\shortstack{\text{Window-based}  \\ [0.1em] \text{Transformer Block} \\ [-0.01em]   } };
 
\node (shifted_window_transformer_blk) [process, yshift=0.97cm,minimum width=4.5cm,minimum height=0.9cm, above of=window_transformer_blk,font=\fontsize{24.5}{26.5}\selectfont]  {\shortstack{\text{(Shifted)} \\ \text{Window-based}  \\ [0.1em] \text{Transformer Block} \\ [-0.01em]   } };
 
\draw[arrow] (shifted_window_transformer_blk)  to node[midway, right, xshift=-3.15cm, yshift=0.42cm, font=\fontsize{20}{22}\selectfont]  {$G \times \frac{H}{d^{\prime}} \times \frac{W}{d^{\prime}}$} ($([xshift=0cm]shifted_window_transformer_blk)+(0,3.4cm)$); 

 \draw[arrow] ($([xshift=0cm]window_transformer_blk)+(0,-2.3cm)$)  to node[midway, right, xshift=-3.15cm, yshift=-0.1cm, font=\fontsize{20}{22}\selectfont]  {$G \times \frac{H}{d^{\prime}} \times \frac{W}{d^{\prime}}$} (window_transformer_blk);


\node[align=center, anchor=north,xshift=-0.2cm,yshift=-0.8cm] 
   at ($(stl.south) + (0,-0.6)$)
   {\parbox{7.8cm}{
     (a) \texttt{Swin} transformer layer ($\mathbf{STL}$)
 }};

\node (cb) [encoder ,minimum width=5.9cm, minimum height=7.20cm, xshift=0cm, yshift=0.7cm,  fill=cb, right=9.3cm of stl.west]  { } ;
\node (heading) [above of=cb,xshift=-2.15cm,yshift=1.65cm]  {$\mathbf{CB}$};
\node (cb_conv1) [process, xshift=0cm,yshift=0.7cm,minimum width=5.5cm,minimum height=0.8cm,font=\fontsize{23.5}{25.5}\selectfont] at (cb.south) {$\mathrm{Conv(1 \times 1, }G)$};

\node (cb_conv1_relu) [process, xshift=0cm,yshift=-0.23cm,minimum width=5.5cm,minimum height=1.0cm, above of=cb_conv1,font=\fontsize{23.5}{25.5}\selectfont]  {$\mathrm{ReLU}$};

\node (cb_conv2) [process, xshift=0cm,yshift=-0.21cm,minimum width=5.5cm,minimum height=0.8cm, above of=cb_conv1_relu,font=\fontsize{23.5}{25.5}\selectfont]  {$\mathrm{Conv(3 \times 3, }G)$};
\node (cb_conv2_relu) [process, xshift=0cm,yshift=-0.23cm,minimum width=5.5cm,minimum height=1.0cm, above of=cb_conv2,font=\fontsize{23.5}{25.5}\selectfont]  {$\mathrm{ReLU}$};

\node (cb_conv3) [process, xshift=0cm,yshift=-0.23cm,minimum width=5.5cm,minimum height=0.8cm, above of=cb_conv2_relu,font=\fontsize{23.5}{25.5}\selectfont]{$\mathrm{Conv(1 \times 1, }G)$};
\draw[arrow] (cb_conv3)  to node[midway, right, xshift=-3.15cm, yshift=0.41cm,  font=\fontsize{20}{22}\selectfont]  {$G \times \frac{H}{d^{\prime}} \times \frac{W}{d^{\prime}}$} ($([xshift=0cm]cb_conv3)+(0,2.6cm)$); 
  \draw[arrow] ($([xshift=0cm]cb_conv1)+(0,-1.97cm)$)  to node[midway, right, xshift=-3.15cm, yshift=-0.1cm,  font=\fontsize{20}{22}\selectfont]  {$G \times \frac{H}{d^{\prime}} \times \frac{W}{d^{\prime}}$} (cb_conv1); 


\node[align=center, anchor=north,xshift=-0.2cm,,yshift=-0.8cm] 
   at ($(cb.south) + (0,-0.6)$)
   {\parbox{8.4cm}{
     (b) Convolutional block ($\mathbf{CB}$)
   }};
 
\end{tikzpicture}

}
                \caption{\texttt{Swin} transformer layer $(\mathbf{STL})$ and convolutional block $(\mathbf{CB})$ employed in Fig.\ \ref{fig:sr_mr}, adopted from \cite{liu2023learned}.}
             \label{fig:stl_cb}
    \end{wrapfigure}
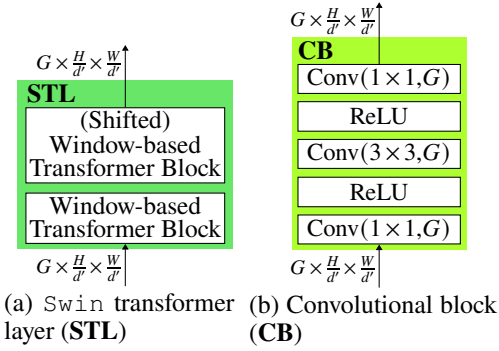~However, the autoregressive formulation enables explicit modeling of inter-channel dependencies in $\mathbf{r}$, which is necessary to enable very low bitrates and an improved RD trade-off \cite{liu2023learned}.

    In Figure \ref{fig:sr_mr}, the detailed network architectures of $\mathbf{SDR}$ and $\mathbf{MDR}$ blocks that we adopt from \cite{liu2023learned} are illustrated. They utilize a combination of \texttt{Swin} transformer layers $\mathbf{STL}$ \cite{liu2021swin} and convolutional blocks $\mathbf{CB}$ \cite{liu2023learned}, which are applied in parallel to facilitate both global and local feature extraction. Two $\mathbf{STLs}$ are utilized, with each $\mathbf{STL}$ comprising a standard window-based transformer block and a shifted window-based transformer block  to extract global features, as shown in Fig.\ \ref{fig:stl_cb}(a). In parallel to $\mathbf{STL}$, a total of six convolutional blocks are employed, with each $\mathbf{CB}$ consisting of three convolutional layers, as shown in Fig.\ \ref{fig:stl_cb}(b). Three $\mathbf{CB}$ blocks are employed in parallel to modulate the output of the remaining three $\mathbf{CB}$ blocks. The modulation is necessary to activate important local information and is performed by applying a $1\times1$ convolution, followed by a sigmoid operation to the output of the three $\mathbf{CB}$ blocks. The output of sigmoid operation is element-wise multiplied with the other remaining three parallel $\mathbf{CB}$ block output and the result is added to the output of the second $\mathbf{STL}$, which is passed through a $1\times1$ convolutional layer to compute the output of the $\mathbf{SDR}$ and $\mathbf{MDR}$ blocks. The element-wise multiplication operator is denoted by $\odot$.

    Back to the $\mathbf{CE}$ and $\mathbf{CD}$ networks in Figure \ref{fig:litcm_codec}: Similar to the first partition $\mathbf{r}_1$, the second partition $\mathbf{r}_2 \in \mathbb{R}^{G \times \frac{H}{d^\prime} \times \frac{W}{d^\prime}}$ is quantized using a quantizer $\mathbf{Q}^{\mathcal{R}}$, resulting in the quantized representation $\hat{\mathbf{r}}_{2}=\mathbf{Q}^{\mathcal{R}}(\mathbf{r}_{2})$. To provide richer context for estimating the distribution parameters of $\mathbf{r}_2$, $\hat{\mathbf{r}}_{1}$ is concatenated with both $\boldsymbol{\sigma}$ and $\boldsymbol{\mu}$ to compute the updated standard deviation $\boldsymbol{\sigma}^{\prime} \in \mathbb{R}^{F_2 \times \frac{H}{d^\prime} \times \frac{W}{d^\prime}}$, and mean $\boldsymbol{\mu}^{\prime} \in \mathbb{R}^{F_2 \times \frac{H}{d^\prime} \times \frac{W}{d^\prime}}$ with number of kernels $F_2$ \cite{liu2023learned,minnen2020channel}. The refined standard deviation $\boldsymbol{\sigma}_2=\mathbf{SDR}(\boldsymbol{\sigma}^{\prime};\bm{\theta}^{\mathrm{SDR}}_2) \! \in \mathbb{R}^{G \times \frac{H}{d^{\prime}} \times \frac{W}{d^{\prime}}}$ and mean $\boldsymbol{\mu}_2\!=\!\mathbf{MDR}(\boldsymbol{\mu}^{\prime};\bm{\theta}_2^{\mathrm{MDR}}) \! \in \! \mathbb{R}^{G \times \frac{H}{d^{\prime}}  \times \frac{W}{d^{\prime}}}$ along with $\hat{\mathbf{r}}_2$ are sent to an arithmetic encoder $\mathbf{AE}^{\mathcal{R}}$ to generate the second partition bitstream $\mathbf{b}_{\hat{\mathbf{r}}_2}=\mathbf{CE}^{(\hat{\mathbf{r}}_2)}(\mathbf{r}_2; \bm{\theta}^{\mathrm{CE}})$. Consisting of $\mathbf{b}_{\hat{\mathbf{r}}_1}$ and $\mathbf{b}_{\hat{\mathbf{r}}_2}$, the latent bitstream is defined as $\mathbf{b}_{\hat{\mathbf{r}}}=(\mathbf{b}_{\hat{\mathbf{r}}_1}, \mathbf{b}_{\hat{\mathbf{r}}_2})$. Note that the weights $\bm{\theta}^{\mathrm{SDR}}_k$ and $\bm{\theta}^{\mathrm{MDR}}_k$ differ for both partitions. 

    The compression \textit{decoder} $\mathbf{CD}$ with parameters $\bm{\theta}^{\mathrm{CD}}$ shown in Figure \ref{fig:litcm_codec} (right) reconstructs the first partition through arithmetic decoder $\mathbf{AD}^{\mathcal{R}}$ along with the $\mathbf{HD}$, $\mathbf{SDR}$, and $\mathbf{MDR}$ blocks as $\hat{\mathbf{r}}_{1}= \mathbf{CD}\!\left(\mathbf{b}_{\hat{\mathbf{r}}_1},\mathbf{b}_{\hat{\mathbf{h}}};\bm{\theta}^{\mathrm{CD}}\right)$. Since the second partition relies on the first partition to provide the necessary context, it can only be reconstructed after the first partition has been reconstructed, dubbed by the original authors as being "autoregressive" \cite{liu2023learned,minnen2020channel}. Similar to the first partition, it is reconstructed with an  arithmetic decoder $\mathbf{AD}^{\mathcal{R}}$ along with the $\mathbf{HD}$, $\mathbf{SDR}$, and $\mathbf{MDR}$ blocks as $\hat{\mathbf{r}}_{2}=\mathbf{CD}\!\left(\mathbf{b}_{\hat{\mathbf{r}}_1},\mathbf{b}_{\hat{\mathbf{r}}_2},\mathbf{b}_{\hat{\mathbf{h}}};\bm{\theta}^{\mathrm{CD}}\right)$, which is concatenated with $\hat{\mathbf{r}}_1$ to reconstruct $\hat{\mathbf{r}}$.

    \section{Experimental Setup}
	\label{sec:experimental_setup}

    \subsection{Datasets}
    \label{subsec:dataset}

    In Table \ref{table:datasets}, the datasets and splits used in our experiments are listed. We provide results on both ADE20K \cite{ADE20K} and Cityscapes \cite{cityscapes} datasets. The ADE20K dataset  presents a significant challenge due to its diverse object categories, densely annotated scenes, and large number of classes. In total, it has $150$ classes spanning across both indoor and outdoor environments. The training and validation splits for ADE20K are denoted by $\mathcal{D}^{\mathrm{train}}_{\mathrm{ADE20K}}$, and $\mathcal{D}^{\mathrm{val}}_{\mathrm{ADE20K}}$, respectively. Furthermore, we also report results on Cityscapes, which focuses on urban scene understanding, especially in the context of autonomous driving. It contains high-resolution images with fine-grained pixel-level annotations across $19$ semantic classes. For Cityscapes, the training and validation splits are represented by $\mathcal{D}^{\mathrm{train}}_{\mathrm{CS}}$, and $\mathcal{D}^{\mathrm{val}}_{\mathrm{CS}}$, respectively. Both datasets enable us to comprehensively evaluate the rate-distortion performance of our proposed source codecs against the current state-of-the-art source codec for distributed semantic segmentation.

    \begin{table}[t!]
      \caption{\textbf{Datasets \& splits} used in our experiments.  }
      \label{table:datasets}
      \centering
      \begin{tabular}{@{}cccc@{}}
        \toprule
         \centering Dataset &  Official subsets & \#images & Symbol \\
        \midrule
       \centering \multirow{2}{*}{ADE20K \cite{ADE20K} }   & train & 20,210 & $\mathcal{D}_{\mathrm{ADE20K}}^{\mathrm{train}}$ \\  [0.4ex]
       \centering    & val & 2,000 & $\mathcal{D}_{\mathrm{ADE20K}}^{\mathrm{val}}$ \\ 
         \midrule
    \centering \multirow{2}{*}{Cityscapes \cite{cityscapes}}   & train & 2,975 & $\mathcal{D}_{\mathrm{CS}}^{\mathrm{train}}$ \\  [0.4ex]
      \centering    & val & 500 & $\mathcal{D}_{\mathrm{CS}}^{\mathrm{val}}$ \\
      \bottomrule
      \end{tabular}
    \end{table}

    \subsection{Experimental Design, Training, and Metrics}
    \label{subsec:experimental_design}

    To perform all our experiments, we utilize the \texttt{MMSegmentation} toolbox \cite{mmseg2020} along with the \texttt{CompressAI} \cite{begaint2020compressai} library. During training, random cropping is applied with an input resolution of $512 \,\times\, 512$ for ADE20K \cite{ADE20K} and $768 \,\times\, 768$ for Cityscapes \cite{cityscapes}. To evaluate the rate–distortion (RD) performance of our proposed source codecs, we employ four baselines: (1) a \textsf{no compression baseline}, without applying any source codec, (2) a (hyperprior source codec baseline) \textsf{HS} \cite{nazirjd,nazir2025efficient} based on the hyperprior architecture described in Section \ref{subsec:proposed_source_codecs}, (3) an inference-time ablated version of the proposed \textsf{AR-HSM} with $\boldsymbol{\mu}=\mathbf{0}$ denoted as \textsf{AR-HS} ($\boldsymbol{\mu}=\mathbf{0}$), where $\boldsymbol{\mu}$ is set to zero only at inference time without performing any additional retraining, (4) a reference (autoregressive hyperprior source codec) \textsf{AR-HS} method with the baseline $\mathbf{HD}$ (cf.\ Fig.\ \ref{fig:hyperprior_block}(a)) and $\mathbf{SDR}$ blocks (cf.\ Fig.\ \ref{fig:sr_mr}(a)). Note that \textsf{AR-HS} is a self-created reference method derived from an ablation of the proposed \textsf{AR-HSM} source codec, which does not include the mean prediction by the $\mathbf{HMD}$ block (cf.\ Fig.\ \ref{fig:hyperprior_block}(b)). Further, following prior works \cite{nazirjd,nazir2025efficient}, we utilize both convolutional neural networks (CNNs) and transformers \cite{nazirjd,nazir2025efficient} on the ADE20K and Cityscapes datasets. As CNNs, we select \texttt{ResNet(RN)-50}\cite{resnet50} as $\mathbf{E}$ with \texttt{DeepLabV3} \cite{deeplabv3} as the semantic segmentation decoder. Among the transformers, we chose four different variants of $\mathbf{E}$ from \texttt{SegFormer} \cite{Segformer}: \texttt{MiT-B0}, \texttt{MiT-B1}, \texttt{MiT-B2}, and \texttt{MiT-B5} with \texttt{SegDeformer} \cite{segdeformer} as the semantic segmentation decoder. Note that the \textsf{source codec baseline} employs the CNN-based \cite{nazirjd} and transformer-based \cite{nazir2025efficient} joint feature and task decoders $\mathbf{JD}$, respectively. The training hyperparameters are provided in Supplement Tables 5 and 6, with details in Supplement Section 1. All trainings and evaluations are conducted on an \texttt{NVIDIA H100} GPU.

  The expected bitrate during training can be defined as
    \begin{equation}
    J^{\mathrm{rate}} = \mathbb{E}_{\bf{x} \sim \mathrm{p_{train}}}  \biggl[ \frac{  \mathrm{-log_{2}} (\mathrm{P_{\bf{\hat{r}}}}(\bf{\hat{r}} | \bf{\hat{h}}))  \mathrm{-log_{2}} (\mathrm{P_{\bf{\hat{h}}}}(\bf{\hat{h}}))  }{H \cdot W} \biggr],
    \label{Eq:rate}
    \end{equation}
    with $\mathbb{E}_{\bf{x} \sim \mathrm{p_{train}} }$ denoting the expectation over a minibatch in the dataset, $\mathrm{P_{\bf{\hat{r}}}}$ and $\mathrm{P_{\bf{\hat{h}}}}$ representing the discrete probability distributions over $\bf{\hat{r}}$ and $\bf{\hat{h}}$, respectively.
    The distortion objective is a \begin{wrapfigure}{r}{0.5\textwidth}
            \hspace{-0.8em} 
              \resizebox{1.0\linewidth}{!}{\begin{tikzpicture}

\definecolor{darkgoldenrod1811370}{RGB}{181,137,0}
\definecolor{darkgreen}{RGB}{0,100,0}
\definecolor{lightgray}{RGB}{211,211,211}
\definecolor{lightgray204}{RGB}{204,204,204}
\definecolor{rosybrown163154141}{RGB}{163,154,141}
\definecolor{voilet}{RGB}{139,69,19}
\begin{axis}[
width=14cm,   
height=9cm,
name = bpp_vs_miou,
legend cell align={left},
legend cell align={left},
legend cell align={left},
scaled ticks=false,
tick align=outside,
tick pos=left,
xtick distance=0.2,
tick label style={color=black,font=\fontsize{19.5}{21.5}\selectfont},
label style={font=\fontsize{19.5}{21.5}\selectfont},
xlabel={bits per pixel (bpp) },
xmajorgrids,
xmin=0.05, xmax=0.9,
xtick style={color=black},
x tick label style={/pgf/number format/precision=10},
x grid style={gray!50, line width=1.25pt},
y grid style={gray!50, line width=1.25pt},
ylabel={\(\displaystyle \textrm{mIoU}\) (\%)},
ylabel style={yshift=-3pt},
xlabel style={yshift=4pt},
ymajorgrids,
ymin=30, ymax=35.5,
ytick style={color=black},
ytick={30,31,32,33,34,35},
yticklabels={
  \(\displaystyle 30\),
  \(\displaystyle 31\),
  \(\displaystyle 32\),
  \(\displaystyle 33\),
  \(\displaystyle 34\),
  \(\displaystyle 35\)
}
]

\addplot [
 name path=litcm,
 line width=1.5pt,
 green,
 mark=star,
 mark size=5,
 mark options={ line width=1.7, solid}
 ]
table [y ] {
x    y     
0.08 31.0  
0.12 32.7 
0.18 33.3 
0.44 34.5 
0.71 35.3 
};
\label{litcm_mitb0_ade20k}

\addplot [
name path=ours,
line width=1.5pt,
green,
mark=square,
mark size=3.5,
mark options={ line width=1.7, solid}
]
table [y ] {
x    y     
0.21 30.1   
0.27 32     
0.37 32.8  
0.45 33.7   
0.87 34.8  
};
\label{ours_mitb0_ade20k}

\path[name path=litcm_horiz]
 (axis cs:0.08,31.0) -- (axis cs:0.24,31.0);

\path[name path=ours_vertical]
 (axis cs:0.545,34.8 ) -- (axis cs:0.87,34.8);

\path [fill=orange!90, fill opacity=0.2] 
(axis cs:0.08,31.0) -- (axis cs:0.24,31.0) 
-- (axis cs:0.27,32) 
-- (axis cs:0.37,32.8) 
-- (axis cs:0.45,33.7) 
-- (axis cs:0.6,34.1)
-- (axis cs:0.87,34.8)
-- (axis cs:0.541,34.8) 
-- (axis cs:0.44,34.5) 
-- (axis cs:0.18,33.3) 
-- (axis cs:0.12,32.7) 
-- cycle;




\addplot [line width=1.5pt, gray , dashed]
table {%
1.00 44
};
\label{No_Compression_fake_legend_ade20k}

\addplot [line width=1.5pt,black,mark=star, mark size=5, green,  mark options={ line width=1.7, solid}]
table {%
1 34.8000030517578
};
\label{Ahuja_fake_legend_ade20k_bd}

\addplot [line width=1.5pt,black,mark=triangle, mark size=5,  mark options={ line width=1.7, solid}]
table {%
1 34.8000030517578
};
\label{Ahuja_fake_legend_ade20k2}

\addplot [line width=1.5pt,black,mark=o, mark size=3.7,  mark options={ line width=1.7, solid}]
table {%
1 34.8000030517578
};
\label{Ahuja_fake_legend_ade20k3}

\addplot [line width=1.5pt,black,mark=square, mark size=3.5, green, mark options={ line width=1.7, solid}]
table {%
1 34.8000030517578
};
\label{ours_fake_legend_bd}

\addplot [line width=1.5pt,  red]
table {%
0.4 34 
};
\label{ours_fake_legend_ade20k}

\addplot [line width=1.5pt,  green]
table {%
0.4 34 
};
\label{mit-b0}

\end{axis}

\node (r_1) [draw=none,fill=none,font=\fontsize{20}{22}\selectfont] at  ([xshift=-3.35cm,yshift=-0.15cm]bpp_vs_miou){\shortstack[l]{
     {\textcolor{red}{\textbf{BD-rate}=}} \\ \textcolor{red}{$-57.2\%$}
   }};

\node [draw,fill=white,font=\fontsize{20}{22}\selectfont] at (rel axis cs:0.67,0.23) {
  \shortstack[l]{
   \hspace*{0mm}\textbf{Source codec method}\\
    \ref*{Ahuja_fake_legend_ade20k_bd} \hspace{-0.01em}\textsf{AR-HSM (\textbf{Ours})} \\
    \ref*{ours_fake_legend_bd} \textsf{HS Baseline} \cite{nazir2025efficient}
  }
};
 \end{tikzpicture}}
                \caption{Illustration of our \textbf{mIoU-adapted BD-rate metric}, inspired from \cite{bjontegaard2001psnr}, employed in Table \ref{table:bd_rate_comparison} for the comparison of our proposed methods vs.\ various baselines.}
             \label{fig:bd_rate}
    \end{wrapfigure}
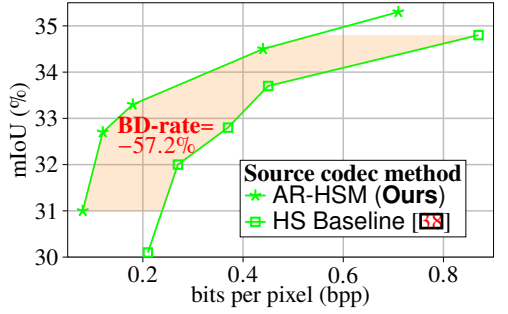cross-entropy loss
    \begin{equation}
    J^{\mathrm{dist}} = \mathbb{E}_{\bf{x} \sim \mathrm{p_{train}}}  \left[ \frac{1}{|\mathcal{I}|}  \sum_{\substack{i \in \mathcal{I} }} \sum_{\substack{s \in \mathcal{S} }} \Bar{y}_{i,s}\cdot\mathrm{log}(y_{i,s}) \right].
    \label{Eq:CE}
    \end{equation}
    Here, $ \Bar{\bm{y}}=(\Bar{y}_{i,s}) \in \{0,1 \}^{H \times W \times S}$ is the one-hot-encoded ground truth and we have $\forall i \in \mathcal{I}:\sum_{s \in \mathcal{S}}y_{i,s}=1$,\quad$\sum_{s \in \mathcal{S}}\Bar{y}_{i,s}=1$. By combining (\ref{Eq:rate}) and (\ref{Eq:CE}), we obtain the RD trade-off in our total loss as follows: 
    \begin{equation}
    J = \alpha \cdot J^{\mathrm{dist}} + (1 - \alpha) \cdot  J^{\mathrm{rate}},
    \label{Eq:RD}
    \end{equation}
    controlled by the hyperparameter $\alpha \in (0,1)$. Note that the baseline (\textsf{HS}), reference method (\textsf{AR-HS}), and the proposed source codecs (\textsf{HSM}, \textsf{AR-HSM}), employ exactly the same objective function defined in (\ref{Eq:RD}).

To evaluate the performance of our proposed source codecs, we report the rate-distortion (RD) performance. The rate is defined following (\ref{Eq:rate}) as bits per pixel (bpp), while the distortion is measured using the mIoU metric \cite{Segformer,segdeformer,nazirjd,ahuja2023neural} to evaluate the performance of the distributed semantic segmentation. Furthermore, we adopt the Bjøntegaard delta (BD)-rate metric \cite{bjontegaard2001psnr} and adapt it to our purposes to quantify the average percentage bitrate savings between the two RD curves at an identical mIoU level. As shown in Fig.\ \ref{fig:bd_rate}, our adaptation of the BD-rate calculation is illustrated using two representative RD curves from Fig.\ \ref{fig:result_comparison}(a), where the shaded area between both curves is used to compute the average BD-rate savings of \textsf{AR-HSM}. Following \cite{barman2024bj}, for ease of interpretation, BD-rate is expressed as a percentage, which corresponds to the average relative bitrate savings required to achieve equivalent mIoU performance as the baseline method \textit{over a common mIoU range}. Note that the BD‑rate is computed relative to the \textsf{HS} baseline and is applicable to all other RD curves shown in Fig.~\ref{fig:result_comparison}. Additionally, we compare the computational complexity of our approach against the prior methods by reporting the number of floating-point operations per image (FLOPs), and number of parameters. The FLOPs and mIoU are measured at the resolutions of $2048 \times 1024 \times 3$ and $512 \times 512 \times 3$ for Cityscapes and ADE20K, respectively. 
Here, the reported percentage corresponds to the average relative bitrate savings required to achieve equivalent mIoU performance as the baseline method, averaged over a common mIoU range.

\section{Experimental Results and Discussion}
\label{sec:experimental_results_discussion}

\begin{figure}[t!]
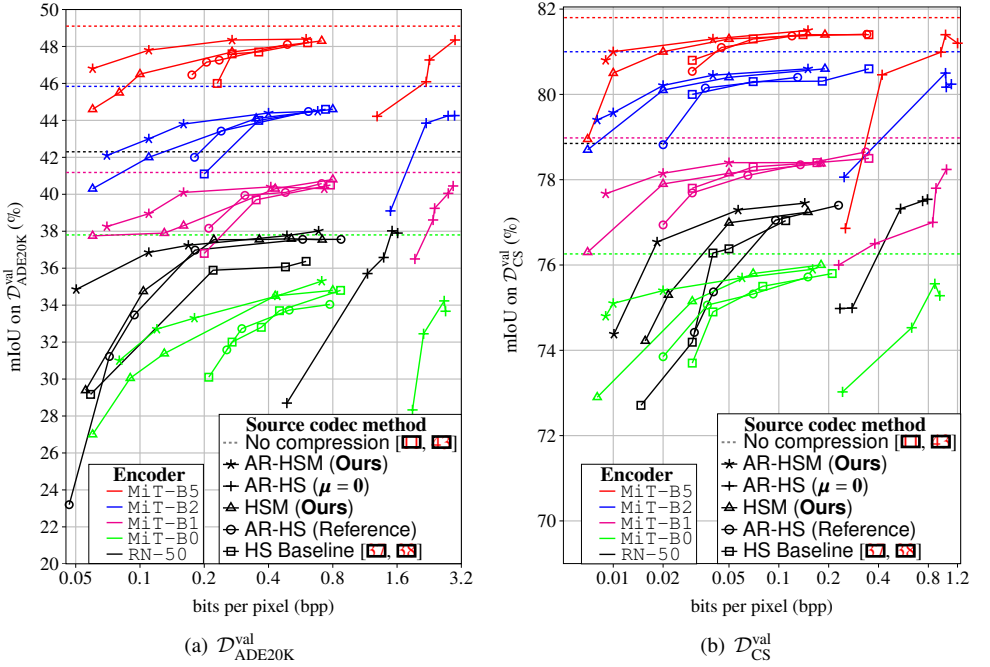

    \centering
    \hspace{-0.6em}
    \subfigure[$\mathcal{D}_{\mathrm{ADE20K}}^{\mathrm{val}}$]{%
         \resizebox{0.495\linewidth}{!}{\input{Figures/results/ade20k}}
        \label{fig:ade20k_results}
    }
    \subfigure[$\mathcal{D}_{\mathrm{CS}}^{\mathrm{val}}$]{%
         \resizebox{0.495\linewidth}{!}{\input{Figures/results/cs}}
        \label{fig:cityscapes_results}
    }

    \caption{Performance (mIoU) of the \textbf{ \textbf{proposed} {\normalfont \textsf{HSM}} and {\normalfont \textsf{AR-HSM}} source codecs vs.\ the state-of-the-art {\normalfont \textsf{HS}} source codec} \cite{nazirjd,nazir2025efficient}, the ablated variant {\normalfont \textsf{AR-HS}} ($\boldsymbol{\mu}=\mathbf{0}$), reference method {\normalfont \textsf{AR-HS}}, and no compression for distributed semantic segmentation on (a) $\mathcal{D}_{\mathrm{ADE20K}}^{\mathrm{val}}$ and on (b) $\mathcal{D}_{\mathrm{CS}}^{\mathrm{val}}$.}
    \label{fig:result_comparison}
\end{figure}

In Figure \ref{fig:result_comparison}, we present the rate-distortion (RD) performance of our proposed \textsf{HSM} and \textsf{AR-HSM} source codecs vs.\ the current state-of-the-art (SOTA) \textsf{HS} baseline in distributed semantic segmentation \cite{nazirjd,nazir2025efficient}, the ablated variant \textsf{AR-HS} with common zero-mean assumption ($\boldsymbol{\mu}=\mathbf{0}$), and the reference method \textsf{AR-HS}. Fig.\ \ref{fig:result_comparison}(a) reports the results on ADE20K \cite{ADE20K}, while Fig.\ \ref{fig:result_comparison}(b) presents the results on Cityscapes \cite{cityscapes}, respectively. Note that the bits per pixel (bpp) axis is shown on a logarithmic scale to improve readability across a wide range of bitrates. We also present qualitative results on both the Cityscapes and ADE20K datasets in Supplement Section 3, Figures 12 and 13. All results, including ours, the \textsf{no compression} \cite{segdeformer} baseline, and the so-far SOTA \textsf{HS} baseline in distributed semantic segmentation \cite{nazirjd,nazir2025efficient} are averaged over three different random seeds.    

In Figure \ref{fig:result_comparison}(a) we observe on ADE20K that, expectedly, the mIoU gets larger with larger encoder sizes (\texttt{MiT-B0} $<$ ... $<$ \texttt{MiT-B5}), while the \texttt{RN-50} encoder delivers suboptimal results, as its performance is below \texttt{MiT-B1}, but its model size is larger. Looking deeper into each of the encoder curve bundles (any color), we identify two overall weak methods, which are the so-far SOTA \textsf{HS} baseline \cite{nazirjd,nazir2025efficient} and our \textsf{AR-HS} reference method, on which we build upon with our specific contributions. Our first proposed method \textsf{HSM} exceeds both of these baselines/references for all investigated bitrates and encoder backbones. This confirms our hypothesis that predicting the standard deviation \textit{and the mean} of the latent representation $\hat{\mathbf{r}}$ during inference is advantageous compared to just assuming a zero mean as is common in literature \cite{nazirjd,nazir2025efficient,ahuja2023neural}. As was displayed in Figure \ref{fig:mean_distribution}, the \textsf{HS} baseline is far from showing a mean variance of zero.

\textit{Our proposed autoregressive method \textsf{AR-HSM}, which also predicts both mean $\boldsymbol{\mu}$ and standard deviation $\boldsymbol{\sigma}$ performs even better than \textsf{HSM} at all bitrates.} In this context, our ablation experiment with the mean explicitly set to zero ($\boldsymbol{\mu}=0$) in inference of \textsf{AR-HSM} reveals that in this case much larger bitrates result. \textit{This confirms the significant effectiveness of our proposed mean predictor as depicted in Figure \ref{fig:he_hsd_hmd}(b).} This holds both on \texttt{MiT-Bn} encoders and on the \texttt{RN-50} encoder.

In Figure \ref{fig:result_comparison}(b), we see all earlier observations confirmed also on Cityscapes, which hints at some generalization property of our proposed \textsf{HSM} and \textsf{AR-HSM} methods.

    \begin{table}[t!]
      \caption{\textbf{BD-rate} savings measured with an input resolution of $512 \times 512$ for $\mathcal{D}^{\mathrm{val}}_{\mathrm{ADE20K}}$ and $2048\times1024$ for $\mathcal{D}^{\mathrm{val}}_{\mathrm{CS}}$ for the proposed \textsf{HSM} and \textsf{AR-HSM} vs. the state-of-the-art (\textsf{HS}) baseline \cite{nazirjd, nazir2025efficient}, the ablated variant {\normalfont \textsf{AR-HS}} ($\boldsymbol{\mu}=0$), and the reference method (\textsf{AR-HS}). BD-rate is reported relative to the \textsf{HS} baseline. In each table segment, best results bold, second-best underlined.}
      \label{table:bd_rate_comparison}
      \centering
      \setlength{\tabcolsep}{2.5pt} 
      \begin{tabular}{llrr}
        \toprule
        \multirow{2}{*}{\rotatebox{90}{\textbf{Enc.}}} & \multirow{2}{*}{\textbf{Source Codec}} & \multicolumn{1}{c}{\textbf{ADE20K} $\mathcal{D}_{\mathrm{ADE20K}}^{\mathrm{val}}$} & \multicolumn{1}{c}{\textbf{Cityscapes} $\mathcal{D}_{\mathrm{CS}}^{\mathrm{val}}$} \\
        & & \multicolumn{1}{c}{\makecell{BD-rate $\downarrow$ \\[-0.7pt] \hspace{-4pt}(\%)}} & \multicolumn{1}{c}{\makecell{BD-rate $\downarrow$ \\[-0.7pt] \hspace{-4pt}(\%)}}  \\
        \midrule
        \multirow{4}{*}{\rotatebox{90}{\texttt{RN-50}}} &  \textsf{HS Baseline} \cite{nazirjd} & $0.0$ & $0.0$ \\
                                            &  \textsf{HSM (ours)} & $-\underline{17.6}$ & $-\underline{41.0}$ \\
                                            & \textsf{AR-HS ($\boldsymbol{\mu}=0$)}  & $+706.9$   & $+621.9$ \\  
                                            & \textsf{AR-HS (reference)}  & $-7.4$   & $+15.5$ \\  
                                            &  \textsf{AR-HSM (ours)} & $\mathbf{-74.3}$ & $\mathbf{-67.3}$ \\
        \midrule
        \midrule
        \multirow{4}{*}{\rotatebox{90}{\texttt{MiT-B0}}} &  \textsf{HS Baseline} \cite{nazir2025efficient} & $0.0$ & $0.0$ \\
                                            &  \textsf{HSM (ours)} & $-\underline{45.5}$ & $-\underline{51.2}$ \\
                                            & \textsf{AR-HS ($\boldsymbol{\mu}=0$)}  & $+660.5$   & $+1559.2$ \\  
                                            & \textsf{AR-HS (reference)}  & $-4.1$   & $-17.0$ \\  
                                            &  \textsf{AR-HSM (ours)} & $\mathbf{-57.2}$ & $\mathbf{-71.5}$ \\
        \midrule
        \midrule
        \multirow{4}{*}{\rotatebox{90}{\texttt{MiT-B1}}} &  \textsf{HS Baseline} \cite{nazir2025efficient} & $0.0$ & $0.0$ \\
                                            &  \textsf{HSM (ours)} & $-\underline{19.5}$ & $-\underline{4.8}$ \\
                                            & \textsf{AR-HS ($\boldsymbol{\mu}=0$)}  & $+754.9$   & $+1900.1$ \\  
                                            & \textsf{AR-HS (reference)}  & $-15.0$   & $+31.9$ \\  
                                            &  \textsf{AR-HSM (ours)} & $\mathbf{-40.1}$ & $\mathbf{-61.0}$ \\
        \midrule
        \midrule
        \multirow{4}{*}{\rotatebox{90}{\texttt{MiT-B2}}} &  \textsf{HS Baseline} \cite{nazir2025efficient} & $0.0$ & $0.0$ \\
                                            &  \textsf{HSM (ours)} & $-\underline{39.7}$ & $-\underline{61.6}$ \\
                                            & \textsf{AR-HS ($\boldsymbol{\mu}=0$)}  & $+635.5$   & $+1169.0$ \\  
                                            & \textsf{AR-HS (reference)}  & $-23.4$   & $-26.9$ \\  
                                            &  \textsf{AR-HSM (ours)} & $\mathbf{-57.9}$ & $\mathbf{-71.7}$ \\
        \midrule    
        \midrule
        \multirow{4}{*}{\rotatebox{90}{\texttt{MiT-B5}}} &  \textsf{HS Baseline} \cite{nazir2025efficient} & $0.0$ & $0.0$ \\
                                            &  \textsf{HSM (ours)} & $-\underline{41.5}$ & $-\underline{37.2}$ \\
                                            & \textsf{AR-HS ($\boldsymbol{\mu}=0$)}  & $+772.4$   & $+2105.3$ \\  
                                            & \textsf{AR-HS (reference)}  & $-11.4$   & $+12.6$ \\  
                                            &  \textsf{AR-HSM (ours)} & $\mathbf{-72.8}$ & $\mathbf{-62.3}$ \\
        \midrule    
        \bottomrule
      \end{tabular}
    \end{table}

In Table \ref{table:bd_rate_comparison}, we investigate the BD-rate \cite{bjontegaard2001psnr} savings on ADE20K $\mathcal{D}_{\mathrm{ADE20K}}^{\mathrm{val}}$ and Cityscapes $\mathcal{D}_{\mathrm{CS}}^{\mathrm{val}}$ datasets for our proposed \textsf{HSM} and \textsf{AR-HSM} source codecs vs.\ the SOTA \textsf{HS} baseline source codec \cite{nazirjd,nazir2025efficient} in distributed semantic segmentation, the ablated variant \textsf{AR-HS} ($\boldsymbol{\mu}=\mathbf{0}$), and the reference method \textsf{AR-HS}. We report results for all backbones \texttt{RN-50}, and \texttt{MiT-B0}, ..., \texttt{MiT-B5} separately in five table segments.

Starting with our reference method \textsf{AR-HS}, we observe that for \texttt{MiT-B0} and \texttt{MiT-B2} it turns out to be better than the \textsf{HS} baseline, while for the other encoders there is a mixed picture with datasets ADE20K and Cityscapes. Concerning our proposed methods, however, we observe that \textit{for all encoder backbones and both datasets}, there is a consistent improvement of \textsf{HSM} (all underlined BD-rates) over the so-far SOTA \textsf{HS} baseline, while our \textsf{AR-HSM} (all bold BD-rates) even exceeds our \textsf{HSM} proposal. This is consistent with our findings in Figure \ref{fig:result_comparison}. \textit{Accordingly, we claim a new SOTA in distributed semantic segmentation for the top-ranked method \textsf{AR-HSM}.}

\begin{figure}[t!]
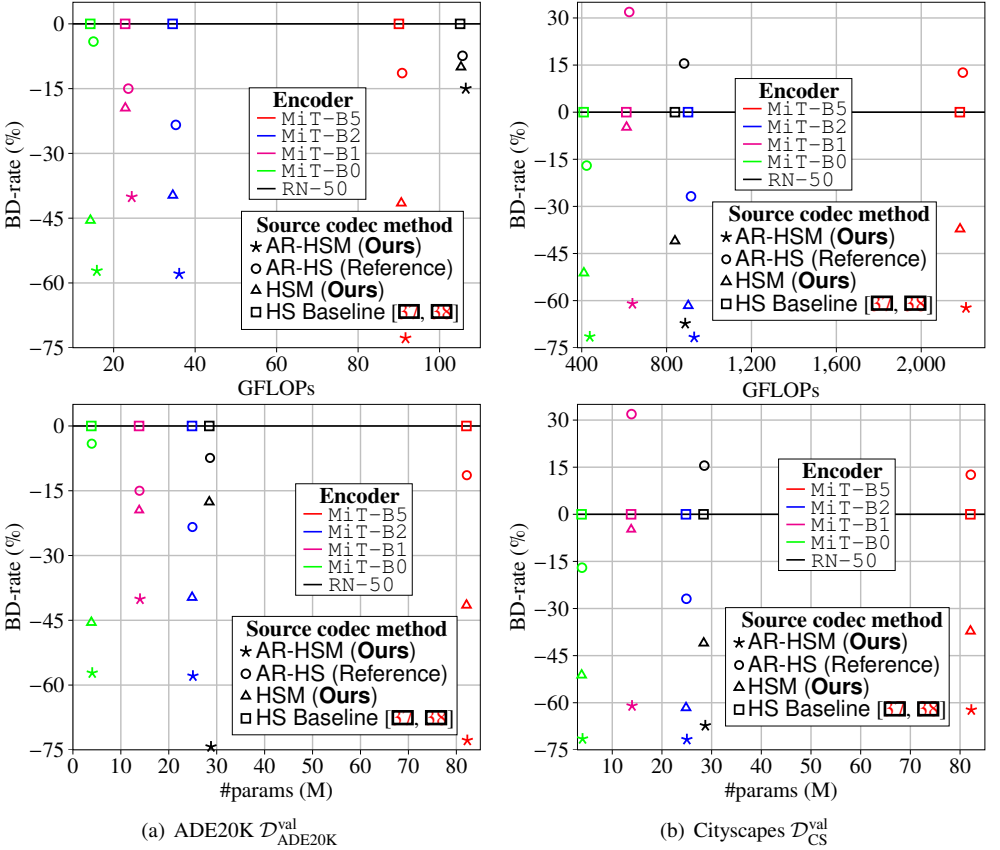

     \hspace{-0.95em}
      \subfigure[ADE20K $\mathcal{D}_{\mathrm{ADE20K}}^{\mathrm{val}}$\label{fig:bdrate_ade20k_results}]{
        \resizebox{0.499\linewidth}{!}{\input{Figures/results/gflops_params_vs_bd_rate_ade20k}}
      }
      \subfigure[Cityscapes $\mathcal{D}_{\mathrm{CS}}^{\mathrm{val}}$\label{fig:bdrate_cityscapes_results}]{
        \resizebox{0.499\linewidth}{!}{\input{Figures/results/gflops_params_vs_bd_rate_cs}}
      }
      \caption{\textbf{BD-rate} savings of our proposals \textsf{HSM} and \textsf{AR-HSM} in comparison to the \textsf{HS} baseline and the \textsf{AR-HSM} reference, in dependence on total GFLOPs (top two plots) and $\#$params (lower two plots), measured with an input resolution of $512\times512$ for $\mathcal{D}^{\mathrm{val}}_{\mathrm{ADE20K}}$ and $2048\times1024$ for $\mathcal{D}^{\mathrm{val}}_{\mathrm{CS}}$.}
      \label{fig:bdrate_results_diagram}
    \end{figure}

    In Figure \ref{fig:bdrate_results_diagram}, we show a detailed comparison of BD-rate \cite{bjontegaard2001psnr} savings in dependence on total GFLOPs (top two plots) and number of parameters $\#$params (lower two plots) for the proposed \textsf{HSM} and \textsf{AR-HSM} source codecs vs.\ the state-of-the-art \textsf{HS} source codec \cite{nazirjd,nazir2025efficient}, and the reference method \textsf{AR-HS}. Note that we omit the BD-rate savings of the ablated version \textsf{AR-HS} ($\boldsymbol{\mu}=\mathbf{0}$) in Figure \ref{fig:bdrate_results_diagram}, since it provides significantly worse BD-rate savings compared to \textsf{HSM} and \textsf{AR-HSM} source codecs (cf.\ Table \ref{table:bd_rate_comparison}). Further, Fig.\ \ref{fig:bdrate_results_diagram}(a) demonstrates the results on $\mathcal{D}_{\mathrm{ADE20K}}^{\mathrm{val}}$, while Fig.\ \ref{fig:bdrate_results_diagram}(b) reports the results on  $\mathcal{D}_{\mathrm{CS}}^{\mathrm{val}}$ \cite{cityscapes} datasets, respectively. The horizontal black lines mark the \textsf{HS} baseline, which is reference to all BD-rate computations, and accordingly results in BD-rate $= 0$. 
  
In Figure \ref{fig:bdrate_results_diagram}(a), we observe on ADE20K (lower image resolution) the expected increase of GFLOPS and $\#$params when going from \texttt{MiT-B0} towards \texttt{MiT-B5}. The \texttt{RN-50} backbone shows the highest computational demand, however, at a moderate network size comparable to \texttt{MiT-B2}. An important observation is that for a certain encoder backbone, \textit{all investigated methods have a very similar demand of GFLOPS and model parameters}, as we see the markers of a given color are almost stacked on top of each other. More detailed GFLOPS and numbers of parameters are provided in Supplement Section 2, Tables 7 and 8. This confirms that our proposed \textsf{HSM} and \textsf{AR-HSM} methods hardly require any such extra resources.

While the BD-rate rank order of \textsf{HS} baseline, our \textsf{HSM} (better), and our \textsf{AR-HSM} (best) is naturally preserved, we nevertheless observe interesting trade-offs among our proposed methods. Let's take our \textsf{AR-HSM} \texttt{MiT-B1} approach (magenta-colored star, BD-rate $=-40.1\%$, $24.4$ GFLOPS) and compare it to the here better (BD-rate $=-45.5\%$) and less complex ($14.3$ GFLOPS) HSM approach: We see that, w.r.t.\ a certain optimization criterion (here GFLOPS) for a practical application, also our proposed non-autoregressive HSM method can be advantageous. Note that---of course---a certain mIoU degradation comes along with this. The same observation can be made within \textsf{AR-HSM} approaches (\texttt{MiT-B0} vs.\ \texttt{MiT-B2}), where BD-rate is roughly the same ($\approx -57\%$), but both GFLOPS and model size can be drastically reduced by just choosing the \texttt{MiT-B0} backbone.

In Figure \ref{fig:bdrate_results_diagram}(b), we observe on Cityscapes (higher image resolution) again that three of the in total six backbones in conjunction with the reference \textsf{AR-HS} method even lead to higher bitrates. Concerning our proposed approaches, the conclusions are very similar to ADE20K, however, \textit{on large image resolutions, our finally proposed \textsf{AR-HSM} seems to be even more consistently strong in BD-rate ($< -60\%$) for all investigated backbones.} This is naturally seen in figures both for GFLOPS and computational complexity. 

Note that we also report edge device, cloud, total computational complexities (edge device plus cloud) and latencies in Supplement Section 2, Tables 7, 8, and 9 for completeness.
    
\section{Conclusions}
\label{sec:conclusions}

In this work, we address the challenge of achieving an improved rate–distortion (RD) trade-off in distributed semantic segmentation, particularly in the low-bitrate regime. Current state-of-the-art (SOTA) methods \cite{nazirjd,nazir2025efficient} rely heavily on hyperprior-based source codecs with a zero-mean assumption, which limits their ability to accurately model the latent feature distribution, resulting in a suboptimal RD trade-off. To overcome this limitation, we proposed two novel source codecs: the hyperprior standard deviation and mean (\textsf{HSM}) source codec and its autoregressive extension (\textsf{AR-HSM}). Our proposed \textsf{HSM} method does not follow the zero-mean assumption, instead, it explicitly predicts during inference both the mean and the standard deviation of the latent representation, which significantly improves the RD trade-off in the low-bitrate regime. Building upon this, \textsf{AR-HSM} incorporates channel-wise autoregressive modeling of the latent representation to capture inter-channel correlations, which further improves the RD performance at very low bitrates. Both \textsf{HSM} and \textsf{AR-HSM} exceed the so-far SOTA on ADE20K and Cityscapes datasets on various backbones and model sizes, confirmed by strong mIoU performance and bitrate savings down to very low bitrates (0.03 ... 0.2 bits per pixel), thereby marking a new SOTA in distributed semantic segmentation. Note that distributed semantic segmentation allows for highly bitrate-efficient cloud-based collection of data from edge devices such as vehicles, while strictly abiding to general data protection laws, as neither pedestrians nor number plates are recognizable.

\bibliography{egbib}
\end{document}


\maketitle

In the supplementary material, we provide in Section 1 a detailed overview of the training settings and hyperparameters required to reproduce all of our results.~Further, in Section~2, we compare the computational complexity incurred on the edge device, in the cloud, as well as the total overall computational complexity of the proposed source codecs vs.\ various baselines. Finally, in Section 3, we provide a qualitative comparison of our proposed source codecs vs.\ various baselines on both ADE20K \cite{ADE20K} and Cityscapes \cite{cityscapes} datasets.


\section{Training Details}

In this section, we provide a detailed description of the employed hyperparameters. For the ease of use and scalability, we integrate the \texttt{Compress‑AI} library \cite{begaint2020compressai} into the \texttt{MMSegmen\-tation} toolbox \cite{mmseg2020}, resulting in a unified framework that is used to conduct all training and evaluation experiments. 

We provide hyperparameter details to enable reproduction of the baselines and proposed source codec results for both settings: (1) a convolutional neural network (CNN)–based setting consisting of a \texttt{ResNet-50} \cite{resnet50} encoder and a \texttt{DeepLabV3} decoder \cite{deeplabv3}, and (2) a transformer-based setting employing \texttt{MiT-Bn} \cite{Segformer} as encoder with \texttt{SegDeformer} \cite{segdeformer} as decoder. The corresponding hyperparameters are reported in Tables~\ref{table:cnn_hyperparameters} and~\ref{table:trans_hyperparameters}, respectively. Following the prior works \cite{nazirjd,nazir2025efficient}, we employed two optimizers: a \texttt{main optimizer} and an \texttt{auxiliary optimizer}. The \texttt{auxiliary optimizer} updates the learnable \texttt{quantiles} parameter of the \texttt{Entropy Bottleneck} \cite{begaint2020compressai}, whereas remaining parameters are optimized by the \texttt{main optimizer}. Further, we employ identical training settings for the proposed hyperprior standard deviation and mean (\textsf{HSM}), and autoregressive hyperprior standard deviation and mean (\textsf{AR-HSM}) source codecs.

\setcounter{table}{4}
\begin{table}[t!]
  \caption{\textbf{Hyperparameters used for CNN-based training} of the so-far SOTA \textsf{HS} source codec, the reference source codec \textsf{AR-HS}, and the proposed \textsf{HSM}, \textsf{AR-HSM} source codecs on the ADE20K and Cityscapes datasets. Here, \texttt{main} and \texttt{aux} denote the hyperparameters used for \texttt{main optimizer} and \texttt{auxiliary optimizer}, respectively.}
  \centering
  \vspace{1pt}
  \begin{tabular}{@{}lcc@{}}
    \toprule
     \centering \textbf{Hyperparameter} &  ADE20K & Cityscapes  \\
    \midrule
     $\#$ of training iterations  & $160,000$  & $80,000$  \\ 
     Batch size   & $16$ & $8$  \\ 
     Random crop  & $512 \times 512$ & $768 \times 768$   \\ 
     Initial learning rate (\texttt{main} and \texttt{aux})  & $1\cdot10^{-3}$ & $1\cdot10^{-3}$  \\ 
     Learning rate schedule (\texttt{main} and \texttt{aux})  & polynomial & polynomial \\ 
     Optimizer (\texttt{main} and \texttt{aux}) & Adam & Adam \\
     Clip grad type (\texttt{aux})   & norm  & norm  \\ 
     Clip grad value (\texttt{aux})   & $1.0$  & $1.0$  \\ 
     Optimizer parameters $\beta_{1}$, $\beta_{2}$ (\texttt{main})   & $0.9, 0.999$ & $0.9, 0.999$  \\ 
     Weight decay (\texttt{main})   & $0$ & $0$  \\ 
  \bottomrule
  \end{tabular}
\label{table:cnn_hyperparameters}
\end{table}

\begin{table}[t!]
  \caption{\textbf{Hyperparameters used for transformer-based training} of the so-far SOTA \textsf{HS} source codec, the reference source codec \textsf{AR-HS}, and the proposed \textsf{HSM}, \textsf{AR-HSM} source codecs on the ADE20K and Cityscapes datasets. Here, \texttt{main} and \texttt{aux} denote the hyperparameters used for \texttt{main optimizer} and \texttt{auxiliary optimizer}, respectively.}
  \centering
  \vspace{1pt}
  \begin{tabular}{@{}lcc@{}}
    \toprule
     \centering \textbf{Hyperparameter} &  ADE20K & Cityscapes  \\
    \midrule
     $\#$ of training iterations  & $160,000$  & $160,000$  \\ 
     Batch size   & $16$ & $8$  \\ 
     Random crop  & $512 \times 512$ & $768 \times 768$   \\ 
     Initial learning rate encoder (\texttt{main})   & $6\cdot10^{-5}$ & $6\cdot10^{-5}$  \\ 
     Initial learning rate decoder (\texttt{main})   & $6\cdot10^{-4}$ & $6\cdot10^{-4}$  \\ 
     Initial learning rate (\texttt{aux})  & $1\cdot10^{-3}$ & $1\cdot10^{-3}$  \\ 
     Learning rate schedule (\texttt{main} and \texttt{aux})  & polynomial & polynomial \\ 
     Optimizer (\texttt{main}) & AdamW & AdamW \\
     Optimizer (\texttt{aux}) & Adam & Adam \\
     Clip grad type (\texttt{aux})   & norm  & norm  \\ 
     Clip grad value (\texttt{aux})   & $1.0$  & $1.0$  \\ 
     Optimizer parameters $\beta_{1}$, $\beta_{2}$ (\texttt{main})   & $0.9, 0.999$ & $0.9, 0.999$  \\ 
     Weight decay (\texttt{main})   & $0.01$ & $0.01$  \\ 
  \bottomrule
  \end{tabular}
\label{table:trans_hyperparameters}
\end{table}

\begin{table}[p!]
      \caption{\textbf{Comparison of total edge device ($\mathbf{E}+\mathbf{FE}+\mathbf{CE}$), and total cloud ($\mathbf{CE}+\mathbf{JD}$) GFLOPs and $\#$params (M)} measured with an input resolution of $512 \times 512$ for $\mathcal{D}_{\mathrm{ADE20K}}^{\mathrm{val}}$ and $2048\!\times\!1024$ for $\mathcal{D}_{\mathrm{CS}}^{\mathrm{val}}$ for the proposed {\normalfont \textsf{HSM}} (cf.\ Fig. 2) and {\normalfont \textsf{AR-HSM}} (cf.\ Fig. 5) vs.\ the state-of-the-art (\textsf{HS}) source codec \cite{nazirjd,nazir2025efficient}, the ablated variant (\textsf{AR-HS} ($\mathbf{\boldsymbol{\mu}=0}$)) and the reference method (\textsf{AR-HS}).}
      \label{table:total_edge_cost}
      \centering
      \setlength{\tabcolsep}{2.5pt} 
      \renewcommand{\arraystretch}{1.} 
        \vspace{3pt}
      \begin{tabular} {llrrrrr}
        \toprule
         & \multirow{3}{*}{\textbf{Encoder}} & \multirow{3}{*}{\textbf{Source Codec}} &  \multicolumn{2}{c}{\textbf{ADE20K} $\mathcal{D}_{\mathrm{ADE20K}}^{\mathrm{val}}$} & \multicolumn{2}{c}{\textbf{Cityscapes} $\mathcal{D}_{\mathrm{CS}}^{\mathrm{val}}$} \\
        \cmidrule(lr){4-5} \cmidrule(lr){6-7}
         & & &  \makecell{GFLOPs} & \makecell{$\#$params \\[-2pt] \hspace{6pt}(M)} & \makecell{GFLOPs } & \makecell{$\#$params \\[-2pt] \hspace{6pt}(M)} \\  
        \midrule
         \multirow{5}{*}{\rotatebox{90}{Edge} }  & \multirow{5}{*}{{\texttt{RN-50}} } &  \textsf{HS Baseline}  \cite{nazirjd} & ${100.00}$ & $23.74$ & $ {801.00}$  & $ {23.74}$ \\  
            & & \textsf{HSM (ours)}  & $ {100.20}$   & $ {23.77}$ & $ {801.20}$  & $ {23.77}$ \\
            & & \textsf{AR-HS ($\boldsymbol{\mu}=0$)}  & $100.40$   & $23.89$ & $803.00$  & $23.89$   \\
            & & \textsf{AR-HS (reference)} & $100.40$   & $23.89$ & $803.00$  & $23.89$  \\  
            & & \textsf{AR-HSM (ours)}  & ${101.00}$   & ${24.08}$ & $805.00$  & $24.08$ \\ 
        \midrule
       \multirow{5}{*}{\rotatebox{90}{Cloud} } & \multirow{5}{*}{{\texttt{RN-50}} } &  \textsf{HS Baseline}  \cite{nazirjd} 
          & $ {5.04}$ & $ {4.71}$ & $ {38.15}$  & $ {4.63}$ \\  
          &  & \textsf{HSM (ours)}  & $ {5.05}$   & $ {4.72}$ & $ {38.23}$  & $ {4.64}$ \\
          &  & \textsf{AR-HS ($\boldsymbol{\mu}=0$)}  & $5.23$   & $ {4.72}$ & $79.43$  & $ {4.64}$   \\
          &  & \textsf{AR-HS (reference)} & $5.23$   & $ {4.72}$ & $79.43$  & $ {4.64}$    \\  
          &  & \textsf{AR-HSM (ours)}  & ${5.43}$   & ${4.73}$ & $82.71$  & $4.65$ \\ 
       \midrule
       \midrule
         \multirow{20}{*}{\rotatebox{90}{Edge} } &   \multirow{5}{*}{{\texttt{MiT-B0}} } &  \textsf{HS Baseline}  \cite{nazirjd,nazir2025efficient} & $ {9.02}$ & $ {3.77}$ & $ {209.00}$  & $ {3.77}$ \\  
            & & \textsf{HSM (ours)}  & $ {9.04}$   & $ {3.79}$ & $ {210.00}$  & $ {3.79}$ \\
            & & \textsf{AR-HS ($\boldsymbol{\mu}=0$)}  & $9.41$   & $3.86$ & $216.00$  & $3.86$   \\
            & & \textsf{AR-HS (reference)} & $9.41$   & $3.86$ & $216.00$  & $3.86$   \\ 
            & & \textsf{AR-HSM (ours)}  & $9.83$   & $3.96$ & $224.00$  & $3.97$ \\ 
        \cmidrule(lr){2-7}
        & \multirow{5}{*}{{\texttt{MiT-B1}} } &  \textsf{HS Baseline}  \cite{nazirjd,nazir2025efficient} & $ {17.58}$ & $ {13.74}$ & $ {410.00}$  & $ {13.74}$ \\  
           & & \textsf{HSM (ours)}  & $ {17.60}$   & $ {13.76}$ & $ {411.00}$  & $ {13.76}$ \\
           & & \textsf{AR-HS ($\boldsymbol{\mu}=0$)}  & $17.97$   & $13.82$ & $417.00$  & $13.82$   \\
           & & \textsf{AR-HS (reference)} & $17.97$   & $13.82$ & $417.00$  & $13.82$  \\  
           & & \textsf{AR-HSM (ours)}  & $18.39$   & $13.93$ & $425.00$  & $13.93$ \\ 
        \cmidrule(lr){2-7}
            & \multirow{5}{*}{{\texttt{MiT-B2}} } &  \textsf{HS Baseline}  \cite{nazirjd,nazir2025efficient} & $ {29.23}$ & $ {24.78}$ & $ {701.00}$  & $ {24.78}$ \\  
            & & \textsf{HSM (ours)}  & $ {29.26}$   & $ {24.80}$ & $ {702.00}$  & $ {24.80}$ \\
            & & \textsf{AR-HS ($\boldsymbol{\mu}=0$)}  & $29.63$   & $24.87$ & $708.00$  & $24.87$   \\
            & & \textsf{AR-HS (reference)} & $29.63$   & $24.87$ & $708.00$  & $24.87$   \\  
            & & \textsf{AR-HSM (ours)}  & ${30.04}$   & ${24.97}$ & $716.00$  & $24.97$ \\ 
        \cmidrule(lr){2-7}
            & \multirow{5}{*}{{\texttt{MiT-B5}} } &  \textsf{HS Baseline}  \cite{nazirjd,nazir2025efficient} & $ {84.78}$ & $ {82.03}$ & $ {1982.00}$  & $ {82.03}$ \\  
            & & \textsf{HSM (ours)}  & $ {84.80}$   & $ {82.05}$ & $ {1983.00}$  & $ {82.05}$ \\
            & & \textsf{AR-HS ($\boldsymbol{\mu}=0$)}  & $85.18$   & $82.11$ & $1989.00$  & $82.11$   \\
            & & \textsf{AR-HS (reference)}  & $85.18$   & $82.11$ & $1989.00$  & $82.11$  \\  
            & & \textsf{AR-HSM (ours)}  & ${85.59}$   & ${82.22}$ & $1997.00$  & $82.22$ \\ 
        \midrule
           \multirow{5}{*}{\rotatebox{90}{Cloud} } & \multirow{5}{*}{{\texttt{MiT-Bn}} } &  \textsf{HS Baseline}  \cite{nazirjd,nazir2025efficient} & $ {5.24}$ & $ {0.06}$ & $ {200.00}$  & $ {0.05}$ \\  
            & & \textsf{HSM (ours)}  & $ {5.26}$   & $ {0.06}$ & $ {200.20}$  & $ {0.05}$ \\
            & & \textsf{AR-HS ($\boldsymbol{\mu}=0$)}  & $5.63$   & $0.07$ & $207.00$  & $ {0.05}$   \\
            & & \textsf{AR-HS (reference)} & $5.63$   & $0.07$ & $207.00$  & $ {0.05}$   \\
            & & \textsf{AR-HSM (ours)}  & $6.05$   & $0.07$ & $214.00$  & $ {0.05}$ \\ 
         \midrule
        \bottomrule
      \end{tabular}
    \end{table}

\section{Computational Complexity}

In this section, we provide a detailed comparison of the total overall computational complexity, as well as the computational complexities incurred on the edge device and in the cloud, measured in terms of giga floating‑point operations (GFLOPs) and the number of learnable parameters (M). Further, we also provide end-to-end latencies of our proposed source codecs.

\begin{table}[p!]
      \caption{\textbf{Comparison of overall total ($\mathbf{E}+\mathbf{FE}+\mathbf{CE}+\mathbf{CD}+\mathbf{JD}$) GFLOPs and $\#$params (M)} measured with an input resolution of $512 \times 512$ for $\mathcal{D}_{\mathrm{ADE20K}}^{\mathrm{val}}$ and $2048\!\times\!1024$ for $\mathcal{D}_{\mathrm{CS}}^{\mathrm{val}}$ for the proposed {\normalfont \textsf{HSM}} (cf.\ Fig. 2) and {\normalfont \textsf{AR-HSM}} (cf.\ Fig. 5) vs.\ the state-of-the-art (\textsf{HS}) source codec \cite{nazirjd,nazir2025efficient}, the ablated variant (\textsf{AR-HS} ($\mathbf{\boldsymbol{\mu}=0}$)) and the reference method (\textsf{AR-HS}).}
      \label{table:total_overall_cost}
      \centering
      \setlength{\tabcolsep}{2.5pt} 
      \renewcommand{\arraystretch}{1.} 
        \vspace{3pt}
      \begin{tabular} {llrrrrr}
        \toprule
         \multirow{3}{*}{\textbf{Encoder}} & \multirow{3}{*}{\textbf{Source Codec}} &  \multicolumn{2}{c}{\textbf{ADE20K} $\mathcal{D}_{\mathrm{ADE20K}}^{\mathrm{val}}$} & \multicolumn{2}{c}{\textbf{Cityscapes} $\mathcal{D}_{\mathrm{CS}}^{\mathrm{val}}$} \\
        \cmidrule(lr){3-4} \cmidrule(lr){5-6}
         & &  \makecell{GFLOPs} & \makecell{$\#$params \\[-2pt] \hspace{6pt}(M)} & \makecell{GFLOPs } & \makecell{$\#$params \\[-2pt] \hspace{6pt}(M)} \\  
        \midrule
            \multirow{5}{*}{{\texttt{RN-50}} } &  \textsf{HS Baseline}  \cite{nazirjd}
            & $ {105.04}$ & $ {28.45}$ & $ {839.15}$  & $ {28.37}$ \\  
            & \textsf{HSM (ours)}  & $ {105.25}$   & $ {28.49}$ & $ {839.43}$  & $ {28.41}$ \\
            & \textsf{AR-HS ($\boldsymbol{\mu}=0$)}  & $105.63$   & $28.61$ & $882.43$  & $28.53$   \\
            & \textsf{AR-HS (reference)}  & $105.63$   & $28.61$ & $882.43$  & $28.53$   \\  
            & \textsf{AR-HSM (ours)}  & ${106.43}$   & ${28.81}$ & $887.71$  & $28.73$ \\ 
        \midrule
        \midrule
            \multirow{5}{*}{{\texttt{MiT-B0}} } &  \textsf{HS Baseline}  \cite{nazirjd,nazir2025efficient} 
            & $ {14.26}$ & $ {3.84}$ & $ {409.00}$  & $ {3.82}$ \\  
            & \textsf{HSM (ours)}  & $ {14.30}$   & $ {3.85}$ & $ {410.00}$  & $ {3.84}$ \\
            & \textsf{AR-HS ($\boldsymbol{\mu}=0$)}  & $15.04$   & $3.93$ & $423.00$  & $3.91$   \\
            & \textsf{AR-HS (reference)} & $15.04$   & $3.93$ & $423.00$  & $3.91$    \\ 
            & \textsf{AR-HSM (ours)}  & $15.88$   & $4.03$ & $438.00$  & $4.02$ \\ 
        \midrule
        \midrule
            \multirow{5}{*}{{\texttt{MiT-B1}} } &  \textsf{HS Baseline}  \cite{nazirjd,nazir2025efficient} 
            & $ {22.82}$ & $ {13.80}$ & $ {610.00}$  & $ {13.79}$ \\  
            & \textsf{HSM (ours)}  & $ {22.86}$   & $ {13.83}$ & $ {611.20}$  & $ {13.81}$ \\
            & \textsf{AR-HS ($\boldsymbol{\mu}=0$)}  & $23.60$   & $13.89$ & $624.00$  & $13.87$   \\
            & \textsf{AR-HS (reference)}  & $23.60$   & $13.89$ & $624.00$  & $13.87$   \\  
            & \textsf{AR-HSM (ours)}  & $24.44$   & $14.00$ & $639.00$  & $13.98$ \\ 
        \midrule
        \midrule
            \multirow{5}{*}{{\texttt{MiT-B2}} } &  \textsf{HS Baseline}  \cite{nazirjd,nazir2025efficient} 
            & $ {34.47}$ & $ {24.85}$ & $ {901.00}$  & $ {24.83}$ \\  
            & \textsf{HSM (ours)}  & $ {34.52}$   & $ {24.87}$ & $ {902.00}$  & $ {24.85}$ \\
            & \textsf{AR-HS ($\boldsymbol{\mu}=0$)}  & $35.26$   & $24.94$ & $915.00$  & $24.92$   \\
            & \textsf{AR-HS (reference)} & $35.26$   & $24.94$ & $915.00$  & $24.92$   \\  
            & \textsf{AR-HSM (ours)}  & ${36.09}$   & $25.04$ & $930.00$  & $25.02$ \\ 
        \midrule
        \midrule
            \multirow{5}{*}{{\texttt{MiT-B5}} } &  \textsf{HS Baseline}  \cite{nazirjd,nazir2025efficient}
            & $ {90.02}$ & $ {82.10}$ & $ {2182.00}$  & $ {82.08}$ \\  
            & \textsf{HSM (ours)}  & $ {90.06}$   & $ {82.12}$ & $ {2183.20}$  & $ {82.10}$ \\
            & \textsf{AR-HS ($\boldsymbol{\mu}=0$)}  & $90.81$   & $82.18$ & $2196.00$  & $82.16$   \\
            & \textsf{AR-HS (reference)}   & $90.81$   & $82.18$ & $2196.00$  & $82.16$  \\  
            & \textsf{AR-HSM (ours)}  & ${91.64}$   & ${82.29}$ & $2211.00$  & $82.27$ \\ 
        \midrule
        \bottomrule
      \end{tabular}
    \end{table}

In Tables \ref{table:total_edge_cost}, and \ref{table:total_overall_cost}, we show a comparison of the edge device, cloud, and total overall computational complexities for both CNN and transformer-based settings, respectively. The comparison is performed on ADE20K $\mathcal{D}_{\mathrm{ADE20K}}^{\mathrm{val}}$ and Cityscapes $\mathcal{D}_{\mathrm{CS}}^{\mathrm{val}}$ datasets for the proposed \textsf{HSM}, and \textsf{AR-HSM} source codecs vs.\ the so-far SOTA hyperprior source codec baseline (\textsf{HS}) \cite{nazirjd,nazir2025efficient}, and the reference autoregressive hyperprior source codec (\textsf{AR-HS}) method. The reported metrics include the total edge device computational complexity $(\mathbf{E} + \mathbf{FE}+\mathbf{CE})$, the total cloud computational complexity $(\mathbf{CD}+\mathbf{JD})$, and the total overall computational complexity $(\mathbf{E}+\mathbf{FE}+\mathbf{CE}+\mathbf{CD}+\mathbf{JD})$ for all source codecs, cf. Fig.\ 1. Here, $\mathbf{E}$, $\mathbf{FE}$ and $\mathbf{CE}$ denote the image, feature and compression encoders, respectively, while $\mathbf{CD}$ and $\mathbf{JD}$ denote the compression and joint feature‑and‑task decoders, respectively.

In Table~\ref{table:total_edge_cost}, we observe that all investigated methods, including the proposed \textsf{HSM} and \textsf{AR-HSM}, have comparable computational complexity (GFLOPs and $\#$params) on both the edge and the cloud across both ADE20K and Cityscapes datasets. This indicates that the observed performance improvements in Figure 10 and Table 2 primarily come from the effective entropy modeling of the latent representations rather than the increased model computational complexity. Another important observation is that both edge and cloud components computational complexities remain well balanced, highlighting that the proposed methods do not shift the computational burden disproportionately to either side of the distributed architecture. Overall, these observations show that substantial improvements in rate-distortion performance can be achieved through better entropy modeling of the latent representation without significantly increasing the computational complexity of the distributed semantic segmentation framework, which is the main focus of our proposed source codecs.

In Table~\ref{table:total_overall_cost}, we see all earlier observations confirmed on the overall total computational complexity, which shows that our proposed \textsf{HSM} and \textsf{AR-HSM} source codecs does not significantly change the total computational complexity of the distributed semantic segmentation framework, making it suiteable for the real-world deployment.

\begin{table}[t!]
\caption{{\textbf{End-to-end latency (ms)} analysis for proposed {\normalfont \textsf{HSM}} and {\normalfont \textsf{AR-HSM}} source codecs vs.\ {\normalfont \textsf{HS baseline}} on an \texttt{MiT-B0} encoder.}}
\label{table:e2e_latency_comparison}
\centering
\setlength{\tabcolsep}{2.5pt} 
\renewcommand{\arraystretch}{1.} 
\vspace{3pt}
\begin{tabular}{lrrrr@{\hspace{0.4em}}crrr}
\toprule

\multirow{2}{*}{\textbf{Source Codec}}
& \multicolumn{4}{c}{\small$\mathcal{D}_{\mathrm{ADE20K}}^{\mathrm{val}}$}
& \multicolumn{4}{c}{\small$\mathcal{D}_{\mathrm{CS}}^{\mathrm{val}}$} \\

\cmidrule(lr){2-5}
\cmidrule(lr){6-9}

& $\tau^{\mathrm{E}}$
& $\tau^{\mathrm{TX}}$
& $\tau^{\mathrm{C}}$
& $\tau^{\mathrm{E2E}}$
& $\tau^{\mathrm{E}}$
& $\tau^{\mathrm{TX}}$
& $\tau^{\mathrm{C}}$
& $\tau^{\mathrm{E2E}}$ \\

\midrule

\textsf{HS Baseline}  \cite{nazir2025efficient}
& \textbf{70}
& 2.9
& \textbf{18.0}
& \underline{90.9}
& \textbf{804}
& 16.7
& \textbf{292}
& \underline{1,113} \\

\textsf{HSM (ours)}
& \underline{72}
& \textbf{0.4}
& \underline{18.2}
& \textbf{90.6}
& \underline{813}
& \underline{3.7}
& \underline{293}
& \textbf{1,110} \\

\textsf{AR-HSM (ours)}
& 117
& \underline{0.5}
& 38.2
& {155.7}
& 1,193
& \textbf{3.4}
& 461
& 1,657 \\

\bottomrule
\end{tabular}
\end{table}

In Table \ref{table:e2e_latency_comparison}, we further report the end-to-end latency $\tau^\mathrm{E2E}$ at low bitrates on both $\mathcal{D}_{\mathrm{ADE20K}}^{\mathrm{val}}$ and $\mathcal{D}_{\mathrm{CS}}^{\mathrm{val}}$. The end-to-end latency is defined as  $\tau^\mathrm{E2E}=\tau^\mathrm{E}+\tau^\mathrm{TX}+\tau^\mathrm{C}$, where $\tau^\mathrm{E}$, $\tau^\mathrm{TX}$, $\tau^\mathrm{C}$ denote the edge device, transmission, and cloud latencies, respectively. We measure $\tau^\mathrm{E}$ on an \texttt{NVIDIA Jetson Orin} and $\tau^\mathrm{C}$ on a workstation with an \texttt{NVIDIA Quadro GV100} GPU. As shown in Table \ref{table:e2e_latency_comparison}, $\tau^\mathrm{E2E}$ shows a clear trade-off: \textit{\textsf{HSM} maintains similar $\tau^\mathrm{E2E}$ as \textsf{HS} baseline, since its improved bitrate reduces $\tau^\mathrm{TX}$, while still improving RD trade-off (cf. Fig.~\textcolor{red}{10} and Tab.~\textcolor{red}{2})}. Further, \textsf{AR-HSM}  also keeps $\tau^\mathrm{TX}$ low and provides even stronger low-bitrate RD performance, but increases $\tau^\mathrm{E2E}$ due to autoregressive nature, making it suitable, when low bitrates are prioritized over latency.

\section{Qualitative Results}
\label{section:qualitative}

In this section, we present a qualitative comparison employing the \texttt{MiT-B1} encoder of our proposed \textsf{HSM} and \textsf{AR-HSM} source codecs against the so-far SOTA \textsf{HS} baseline source codec \cite{nazirjd} at low bitrates on both $\mathcal{D}^{\mathrm{ADE20K}}_{\mathrm{val}}$ and $\mathcal{D}^{\mathrm{CS}}_{\mathrm{val}}$. 

Figure \ref{fig:cs_low_qualitative} presents a qualitative comparison for seven samples (organized in rows) of our proposed \textsf{HSM} and \textsf{AR-HSM} source codecs vs.\ the SOTA \textsf{HS} baseline source codec \cite{nazirjd} \textit{at low bitrates on $\mathcal{D}^{\mathrm{CS}}_{\mathrm{val}}$ samples}. Sample-individual \textit{mIoU} performance is reported, while the \textit{bitrates} of the \textsf{HS} baseline \cite{nazirjd} and our proposed \textsf{HSM} and \textsf{AR-HSM} source codecs are given for the entire $\mathcal{D}^{\mathrm{CS}}_{\mathrm{val}}$ samples. In all the rows, we observe that both \textsf{HSM} and \textsf{AR-HSM} achieve higher mIoU performance, while attaining significantly lower bitrates than the \textsf{HS} baseline source codec. Note that the bitrates of our proposed source codecs are always kept lower than those of the \textsf{HS} baseline source codec. 

Figure \ref{fig:ade20k_low_qualitative} depicts a qualitative comparison of our proposed \textsf{HSM} and \textsf{AR-HSM} source codecs vs.\ the SOTA \textsf{HS} baseline source codec \cite{nazirjd} \textit{at low bitrates on $\mathcal{D}^{\mathrm{ADE20K}}_{\mathrm{val}}$ samples}. We observe that, across all rows, our \textsf{HSM} and \textsf{AR-HSM} methods achieve similar or slightly better performance than the \textsf{HS} baseline source codec. Important to note: Both, \textsf{HSM} and \textsf{AR-HSM}, operate at significantly lower bitrates than the \textsf{HS} baseline.

\setcounter{figure}{11}
\begin{figure}[t!]
\centering

\begin{minipage}[t]{0.2455\textwidth}
  \centering
  \includegraphics[width=\linewidth]{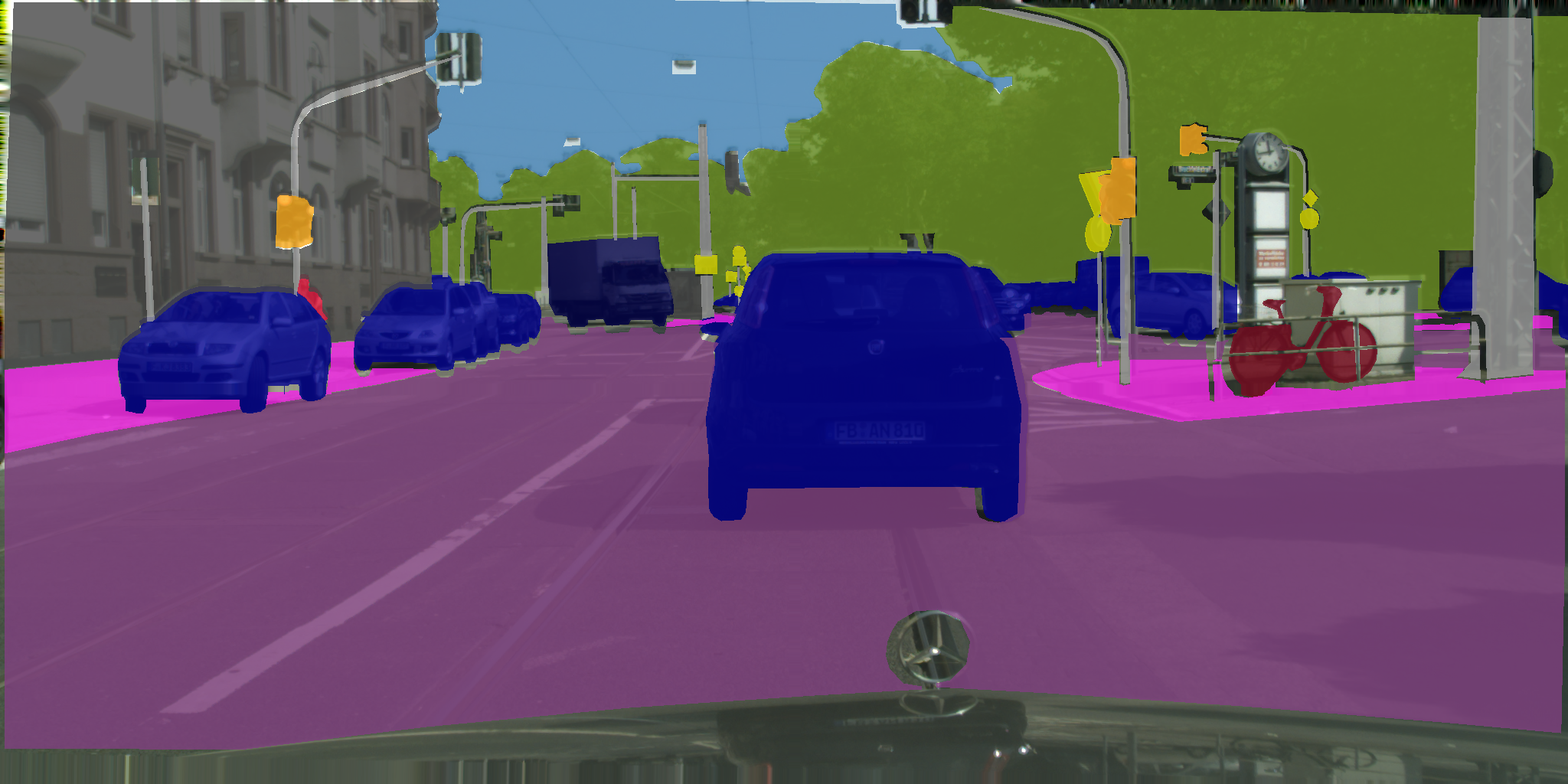}
  \includegraphics[width=\linewidth]{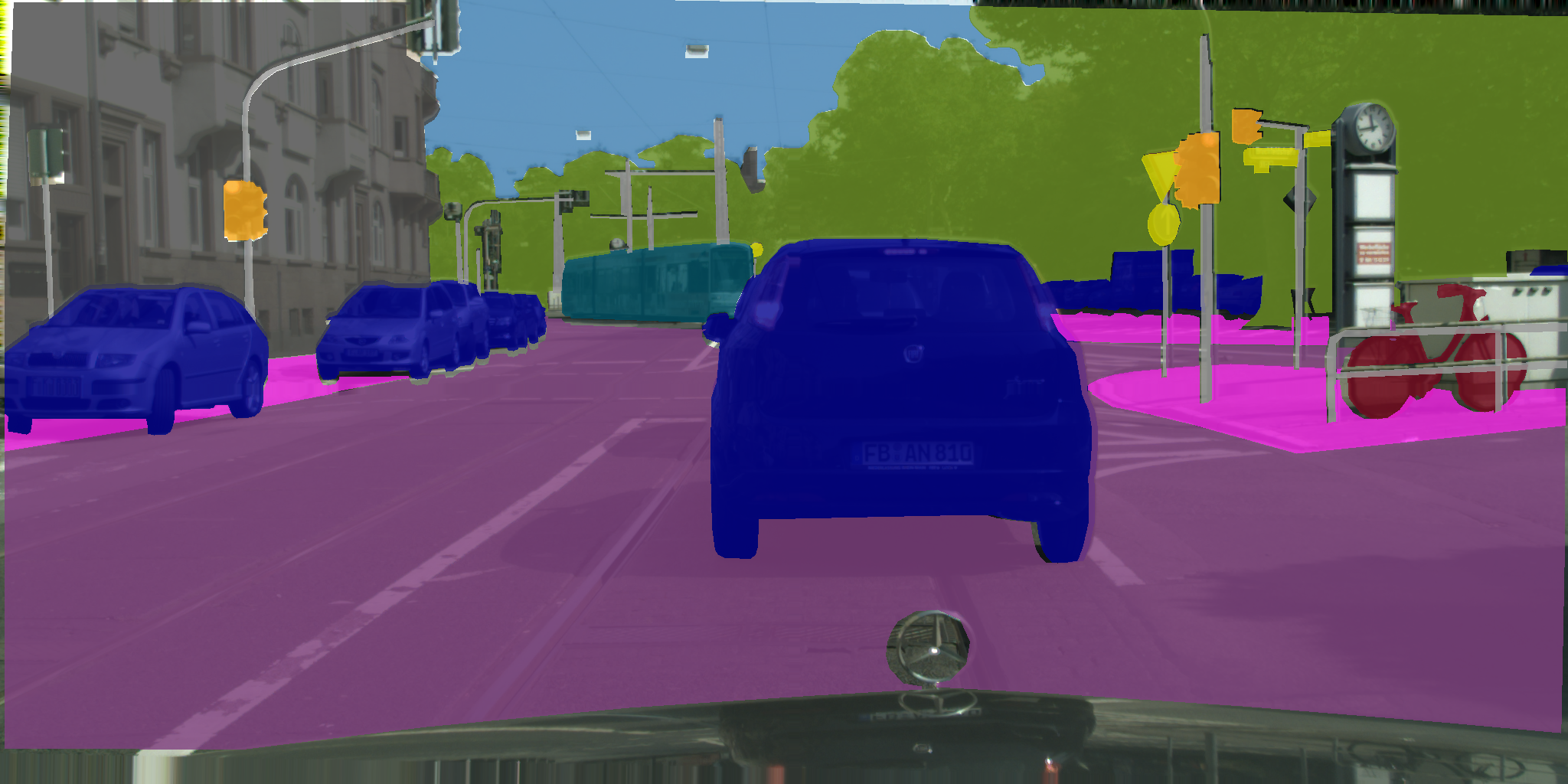}
  \includegraphics[width=\linewidth]{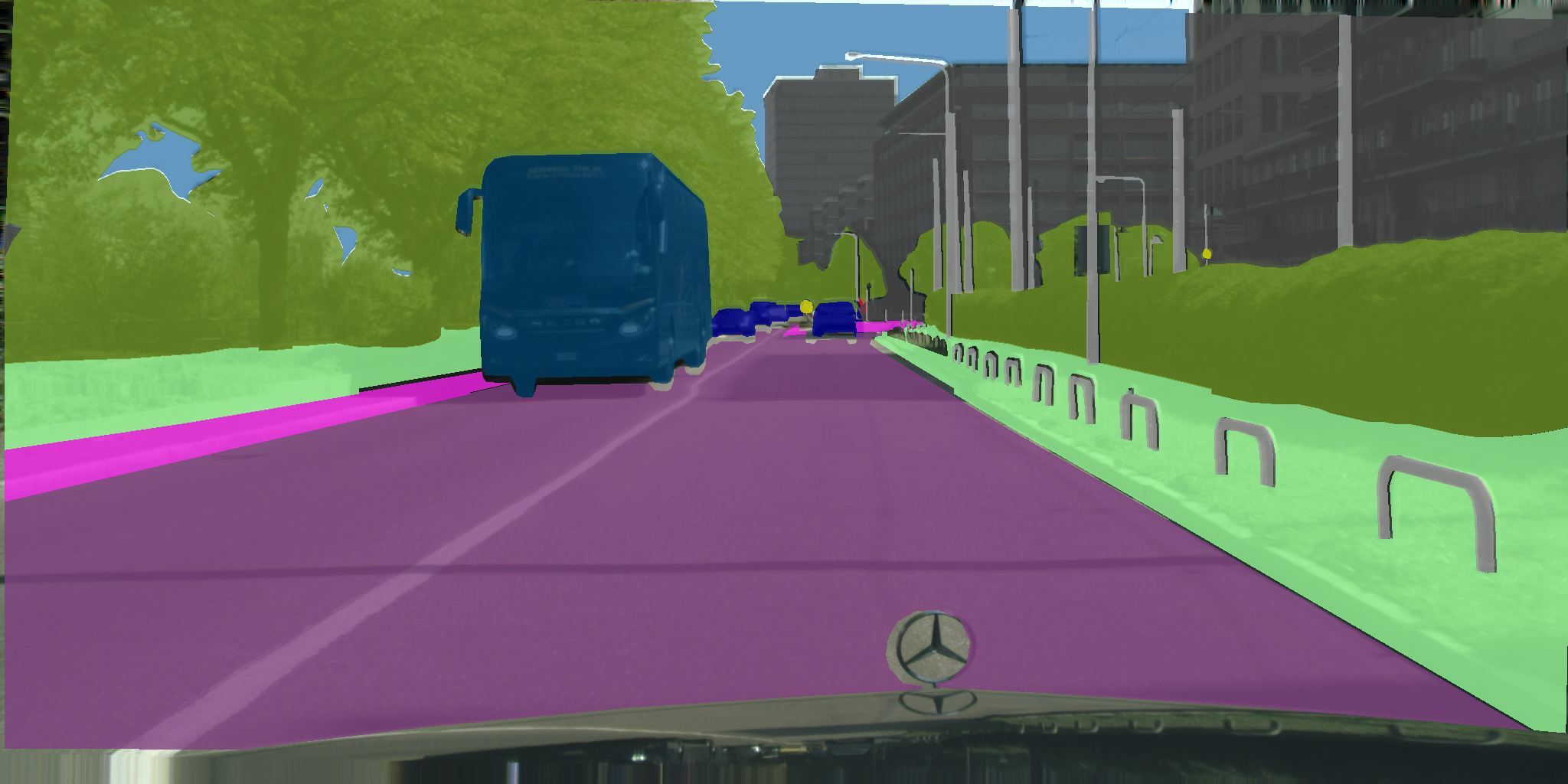}
  \includegraphics[width=\linewidth]{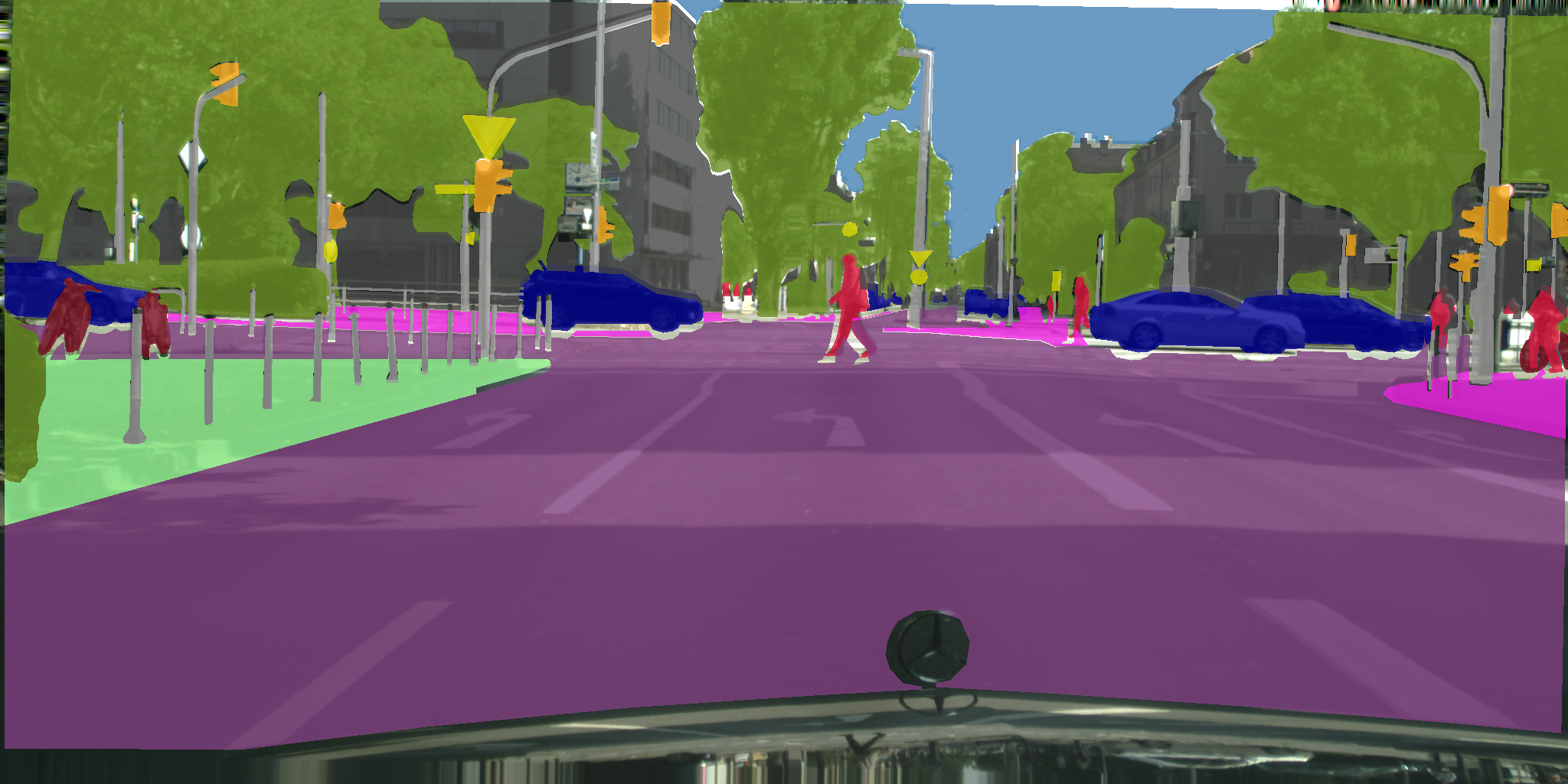}
  \includegraphics[width=\linewidth]{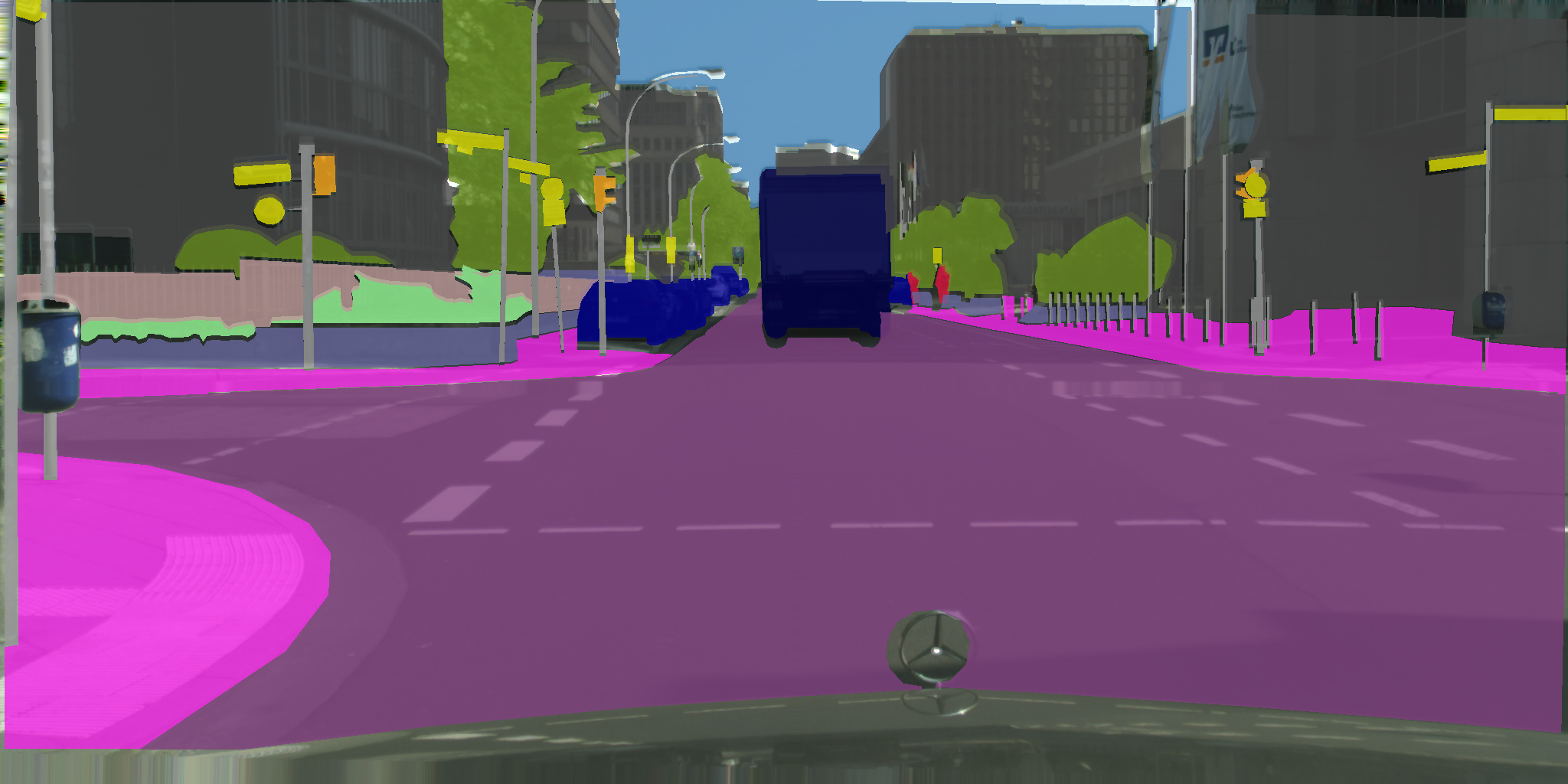}
  \includegraphics[width=\linewidth]{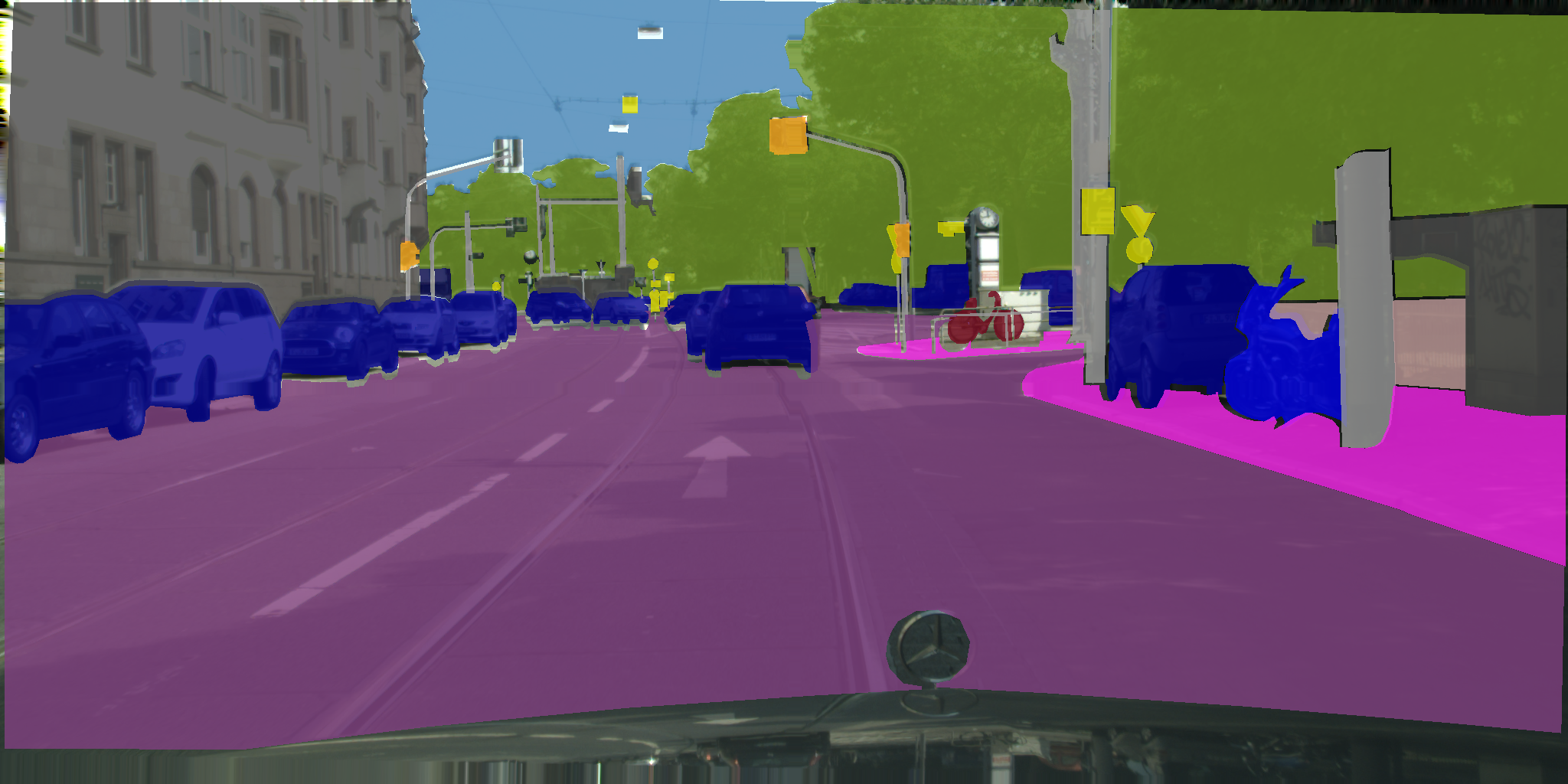}
    \includegraphics[width=\linewidth]{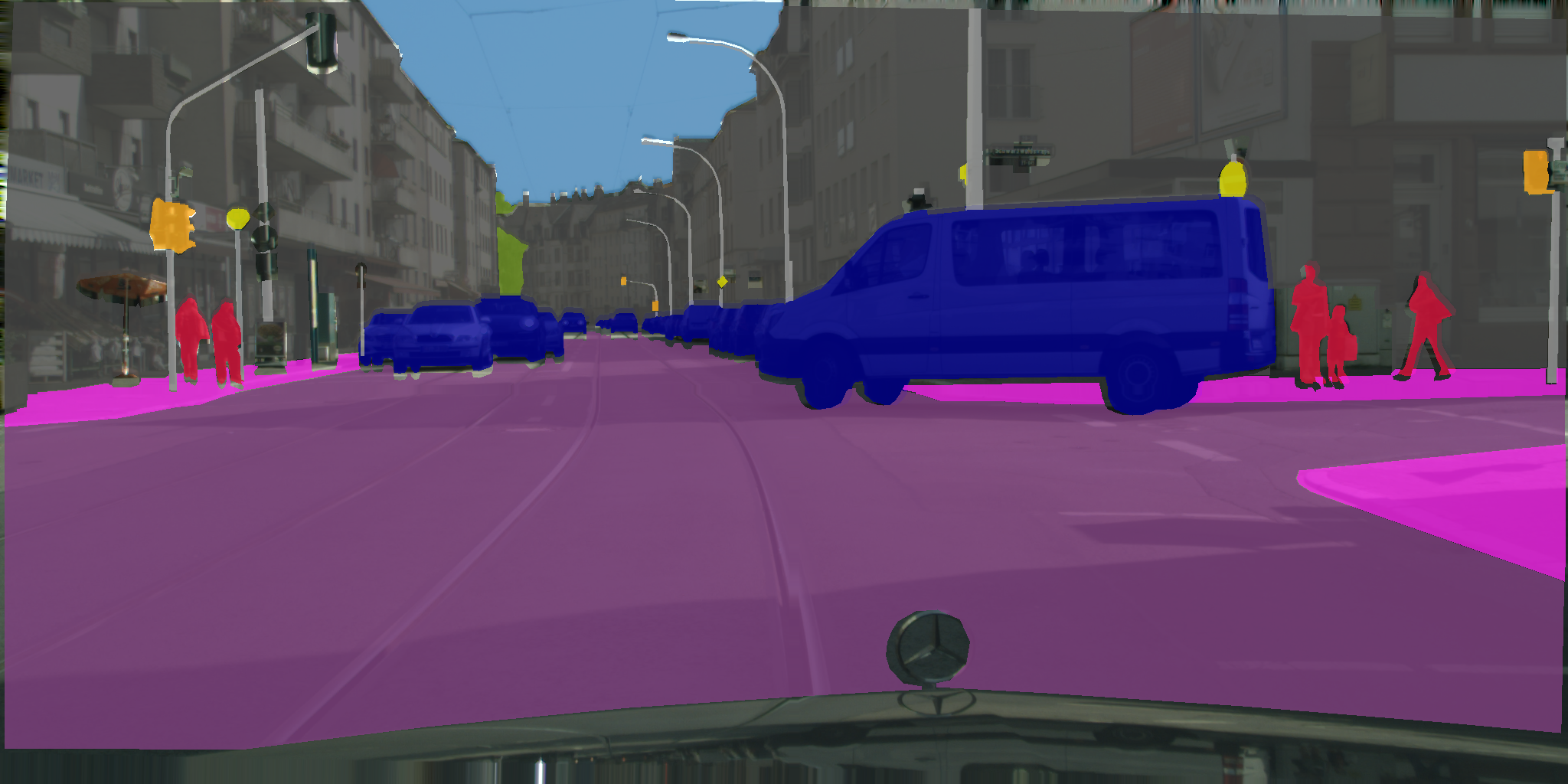}
  \vspace{0.2em}
  {\small a) Ground Truth}
\end{minipage}\hfill
%
\begin{minipage}[t]{0.2455\textwidth}
  \centering
  \includegraphics[width=\linewidth]{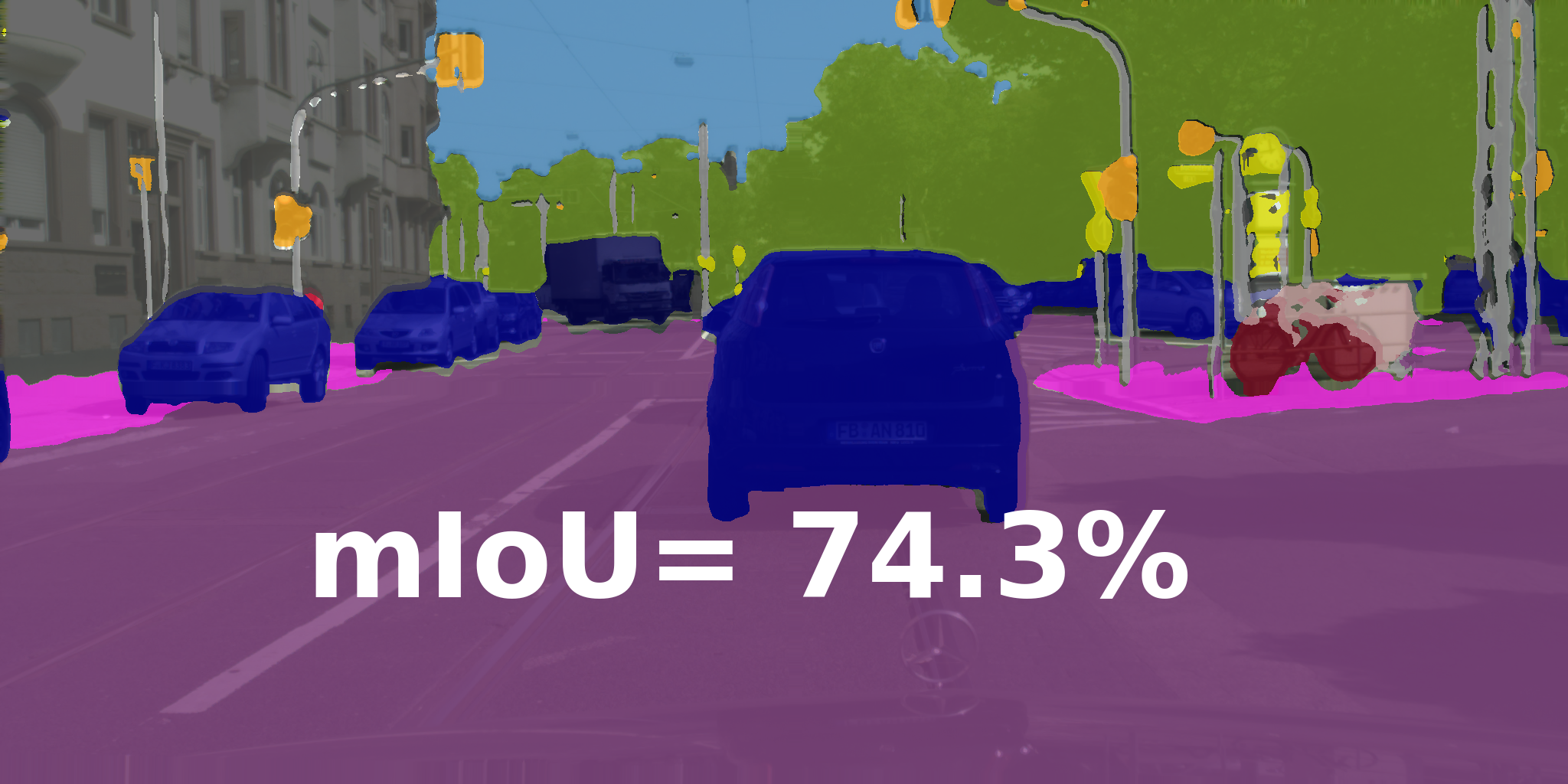}
  \includegraphics[width=\linewidth]{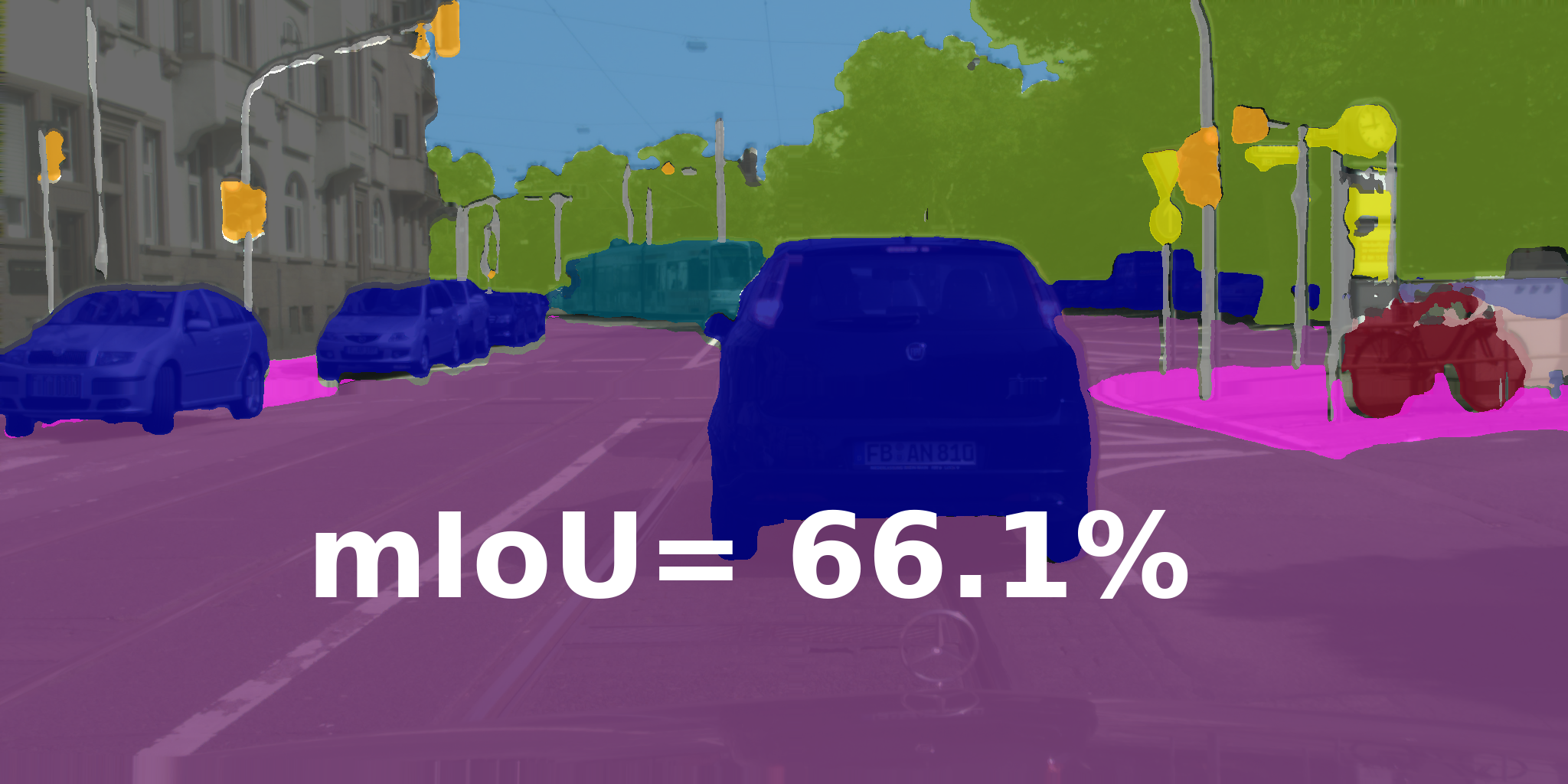}
  \includegraphics[width=\linewidth]{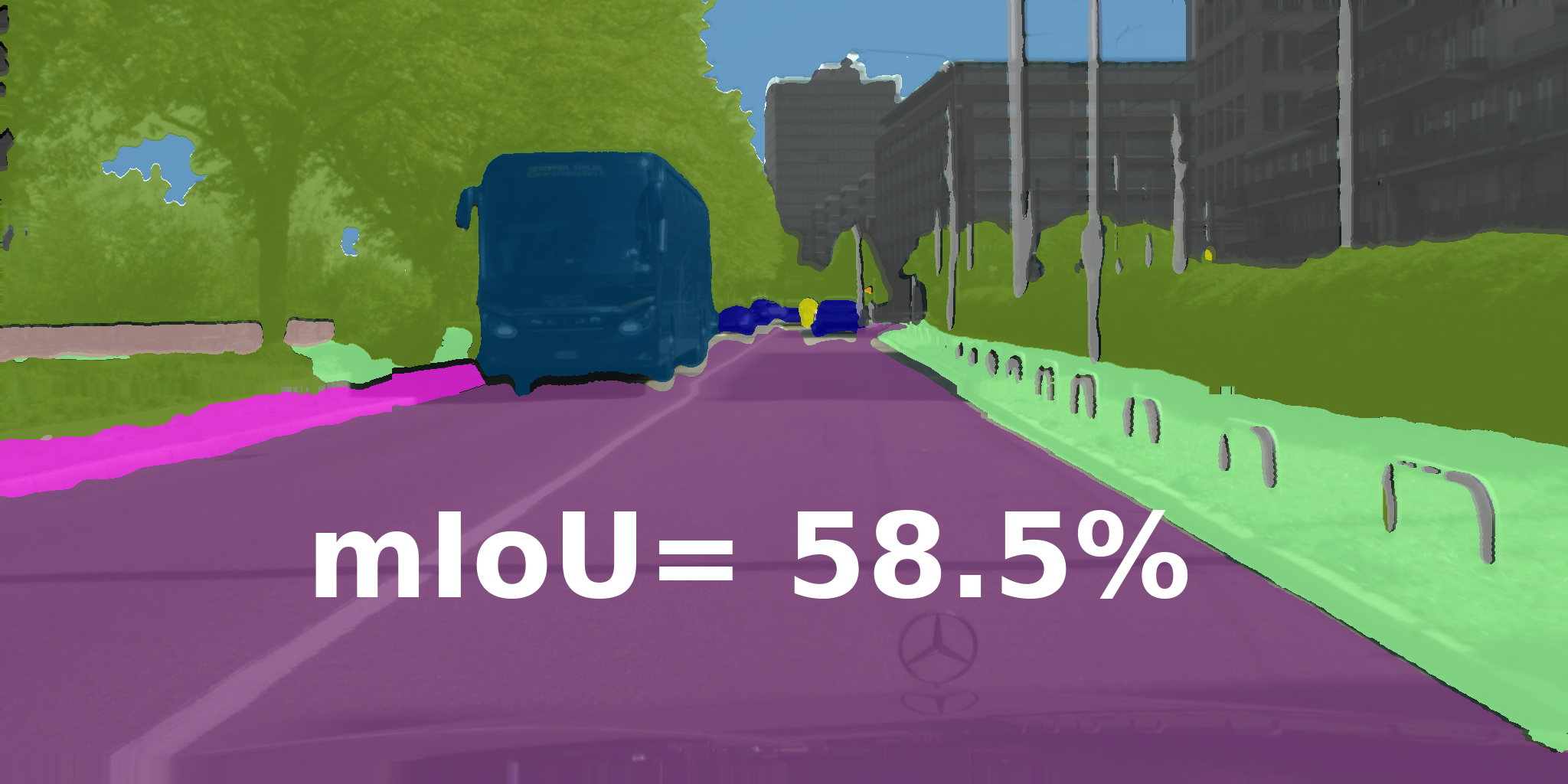}
  \includegraphics[width=\linewidth]{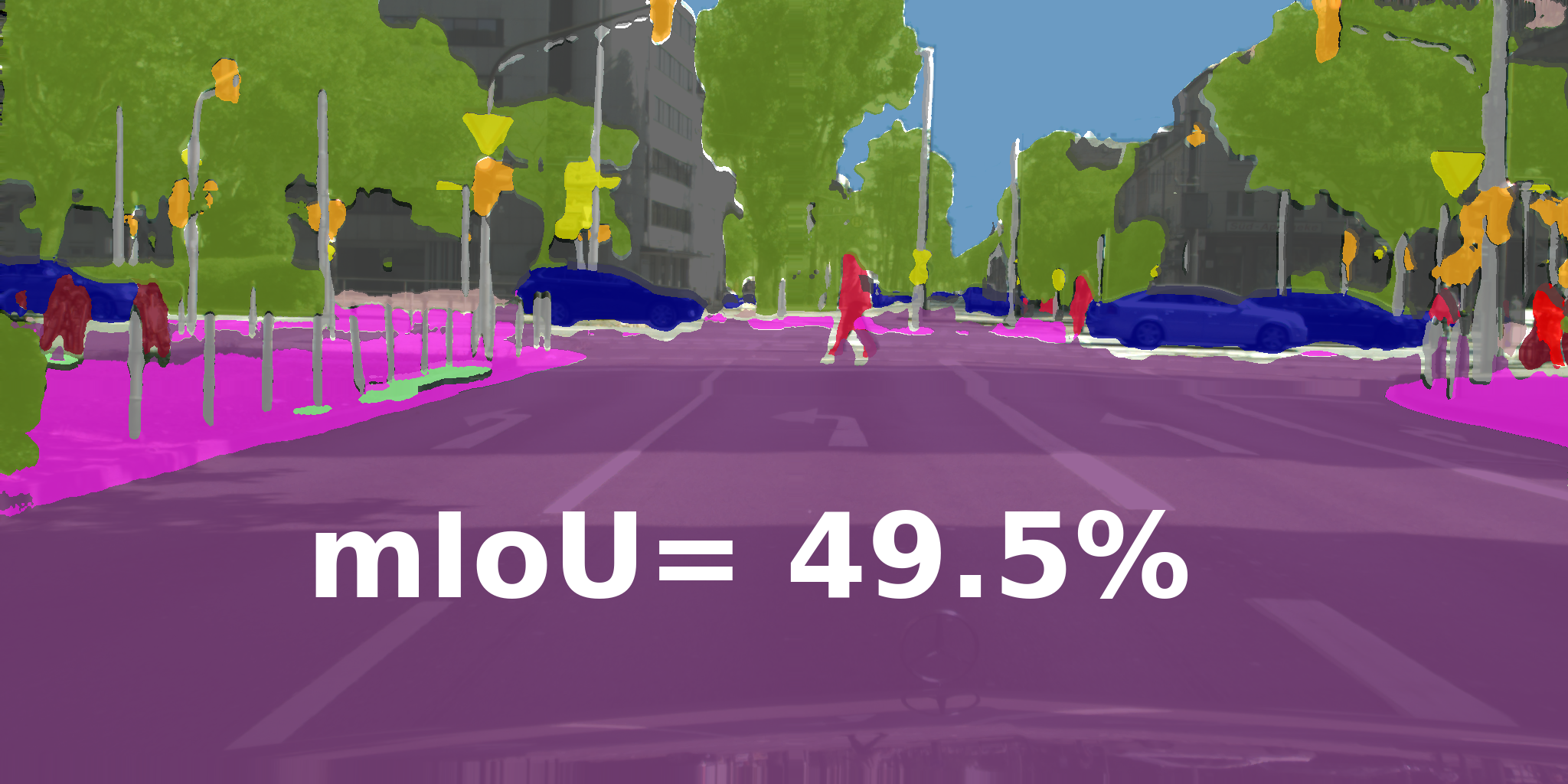}
  \includegraphics[width=\linewidth]{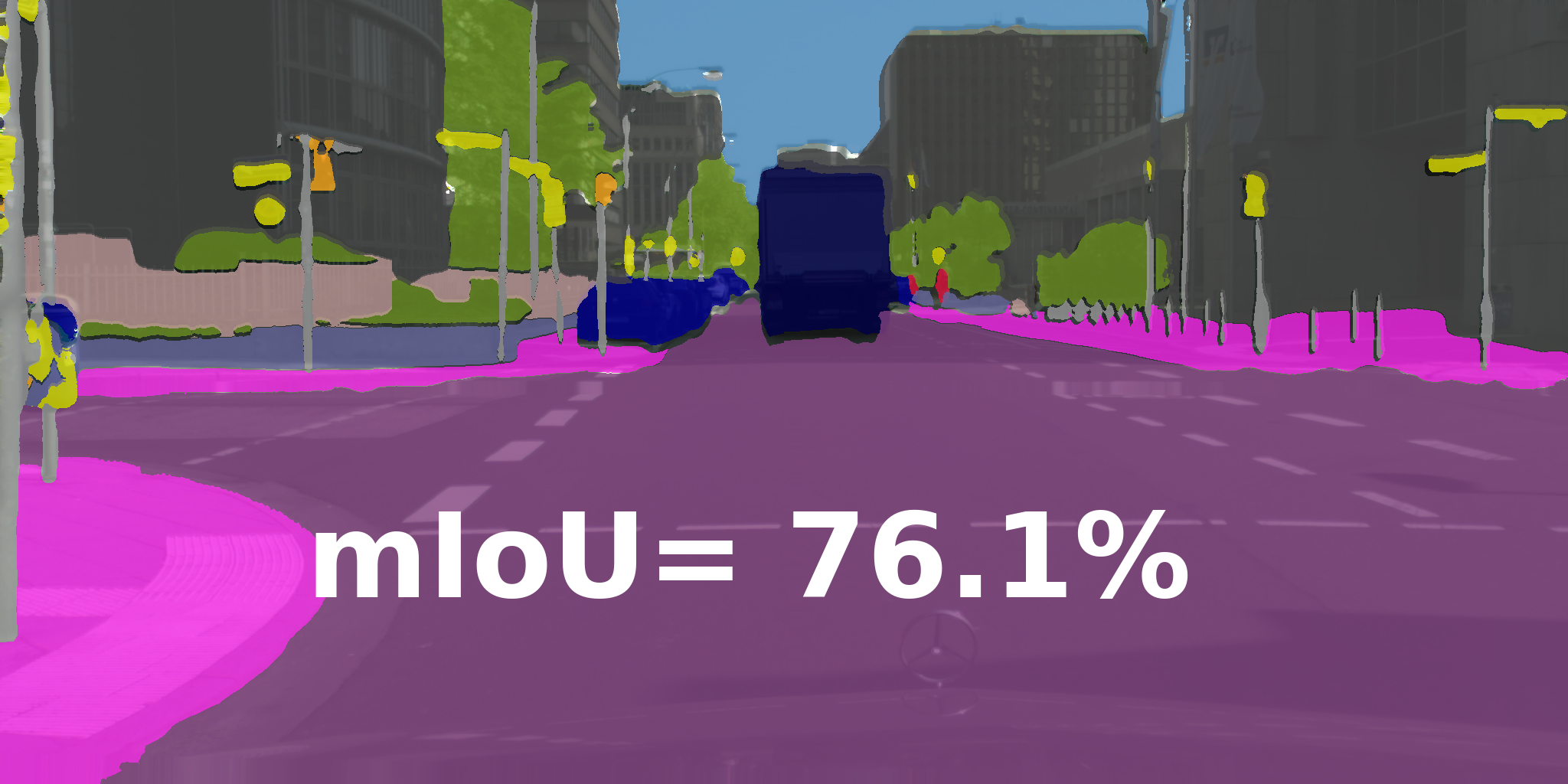}
  \includegraphics[width=\linewidth]{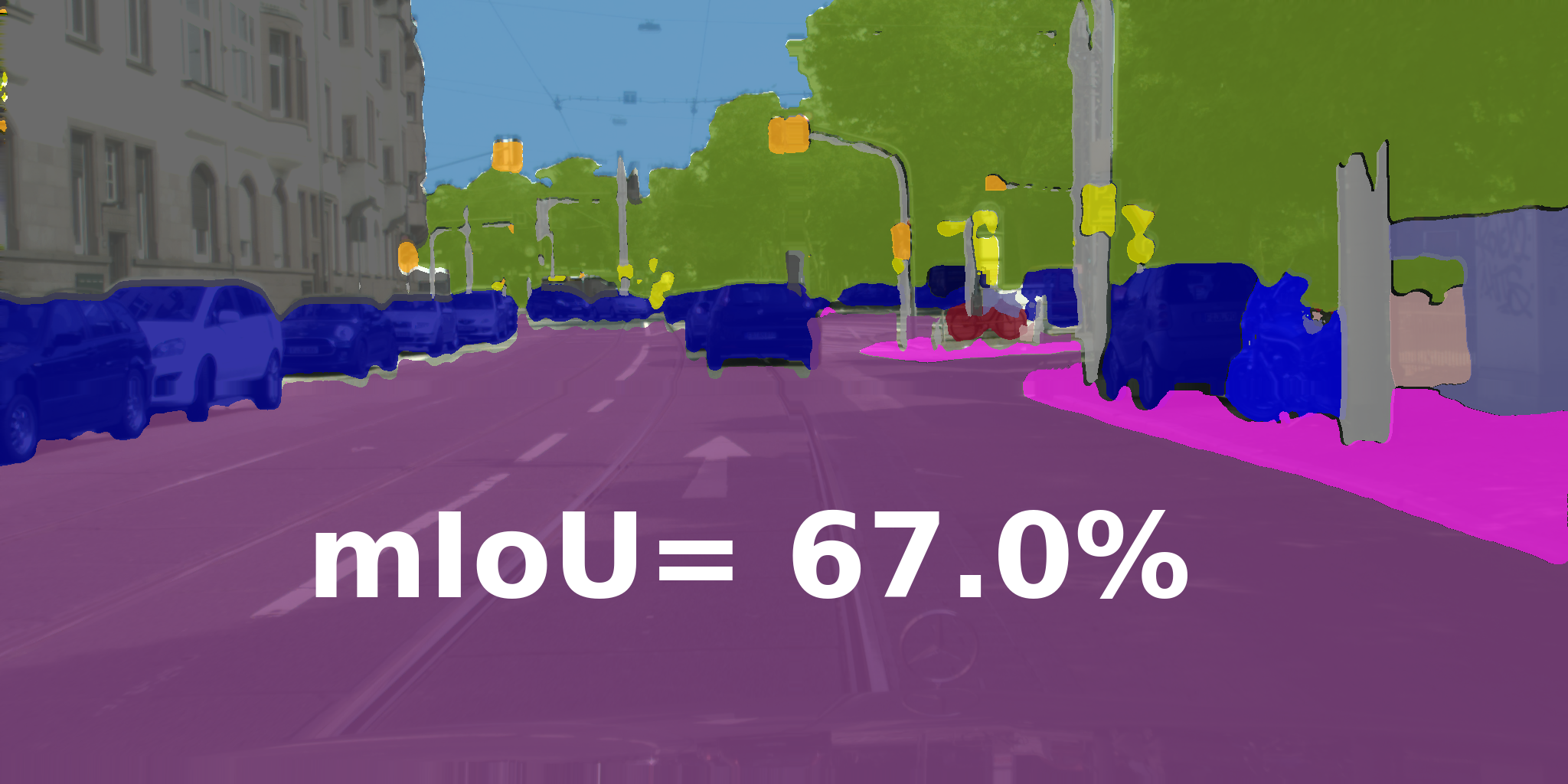}
    \includegraphics[width=\linewidth]{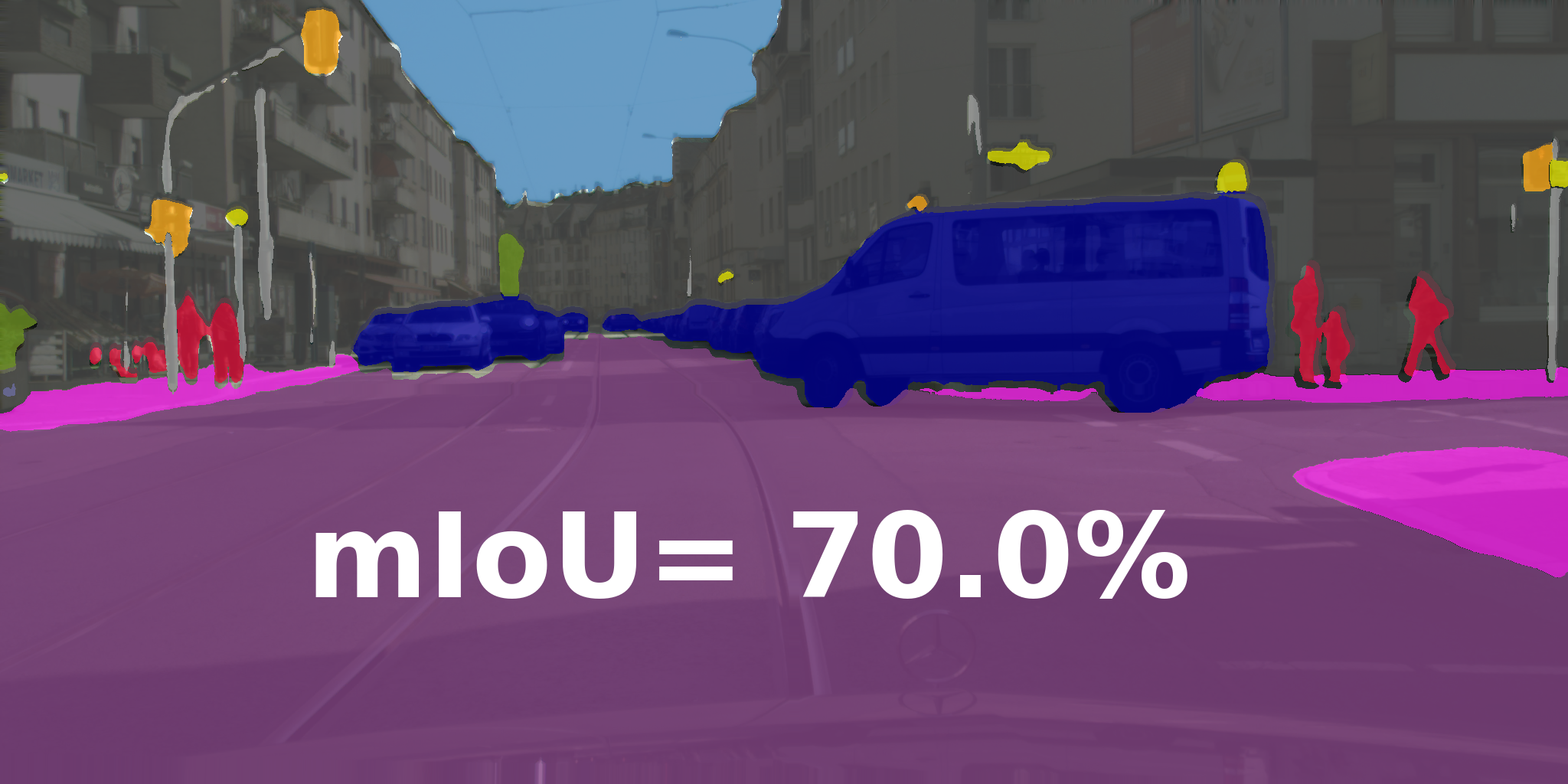}
  \vspace{0.2em}
  {\small b) \textsf{HS} ($0.035\,\mathrm{bpp}$)}
\end{minipage}\hfill
\begin{minipage}[t]{0.2455\textwidth}
  \centering
  \includegraphics[width=\linewidth]{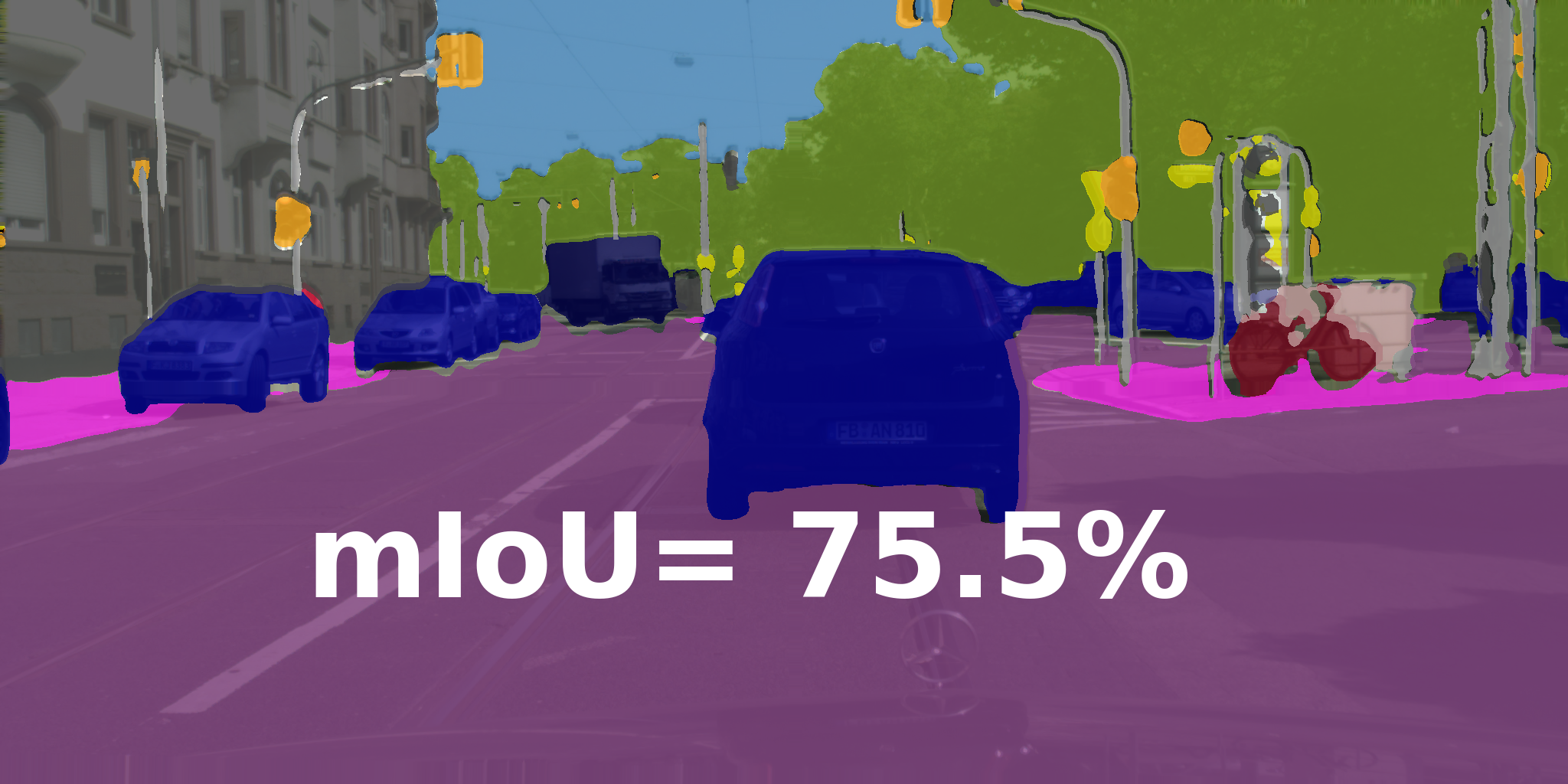}
  \includegraphics[width=\linewidth]{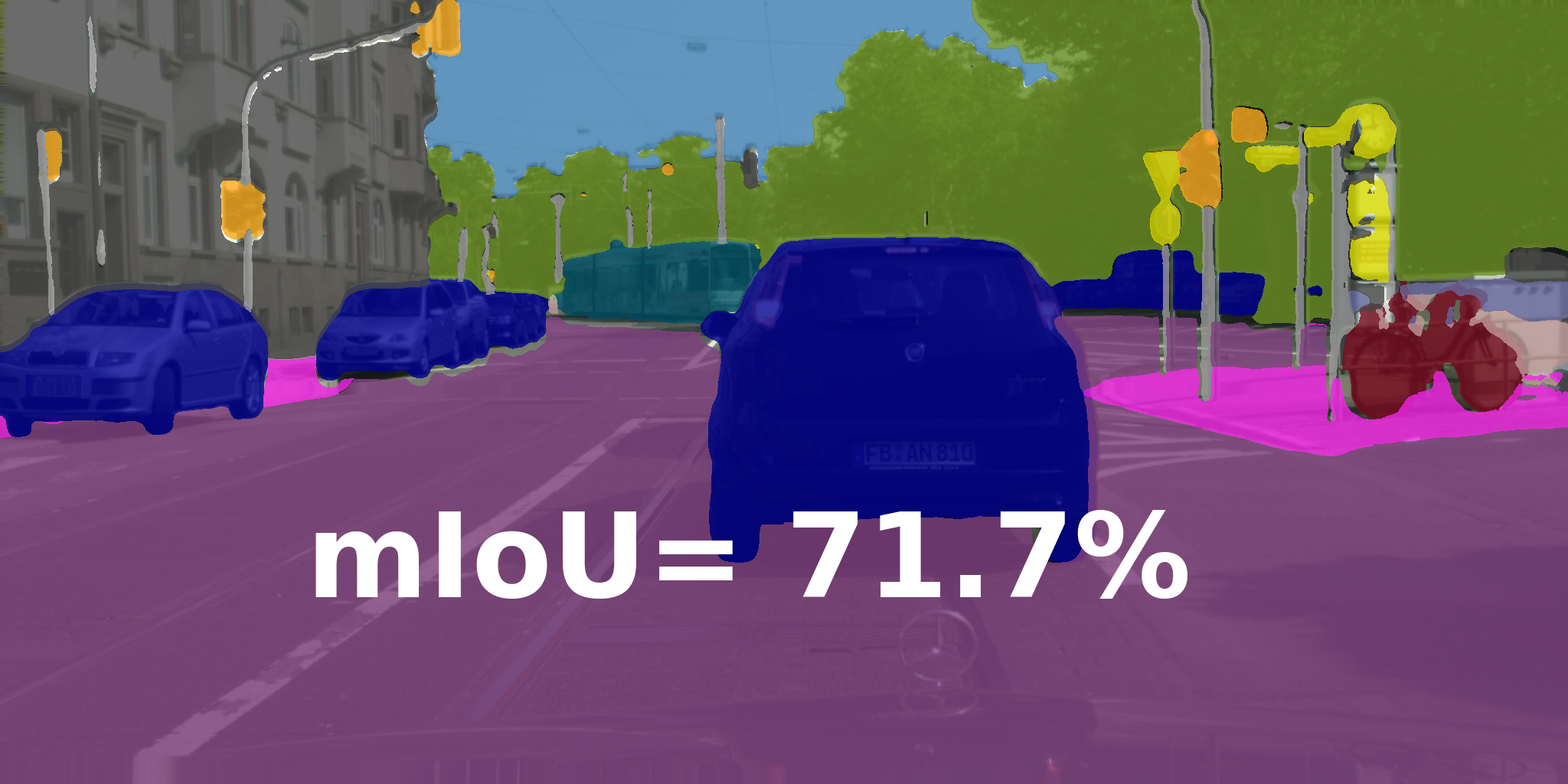}
  \includegraphics[width=\linewidth]{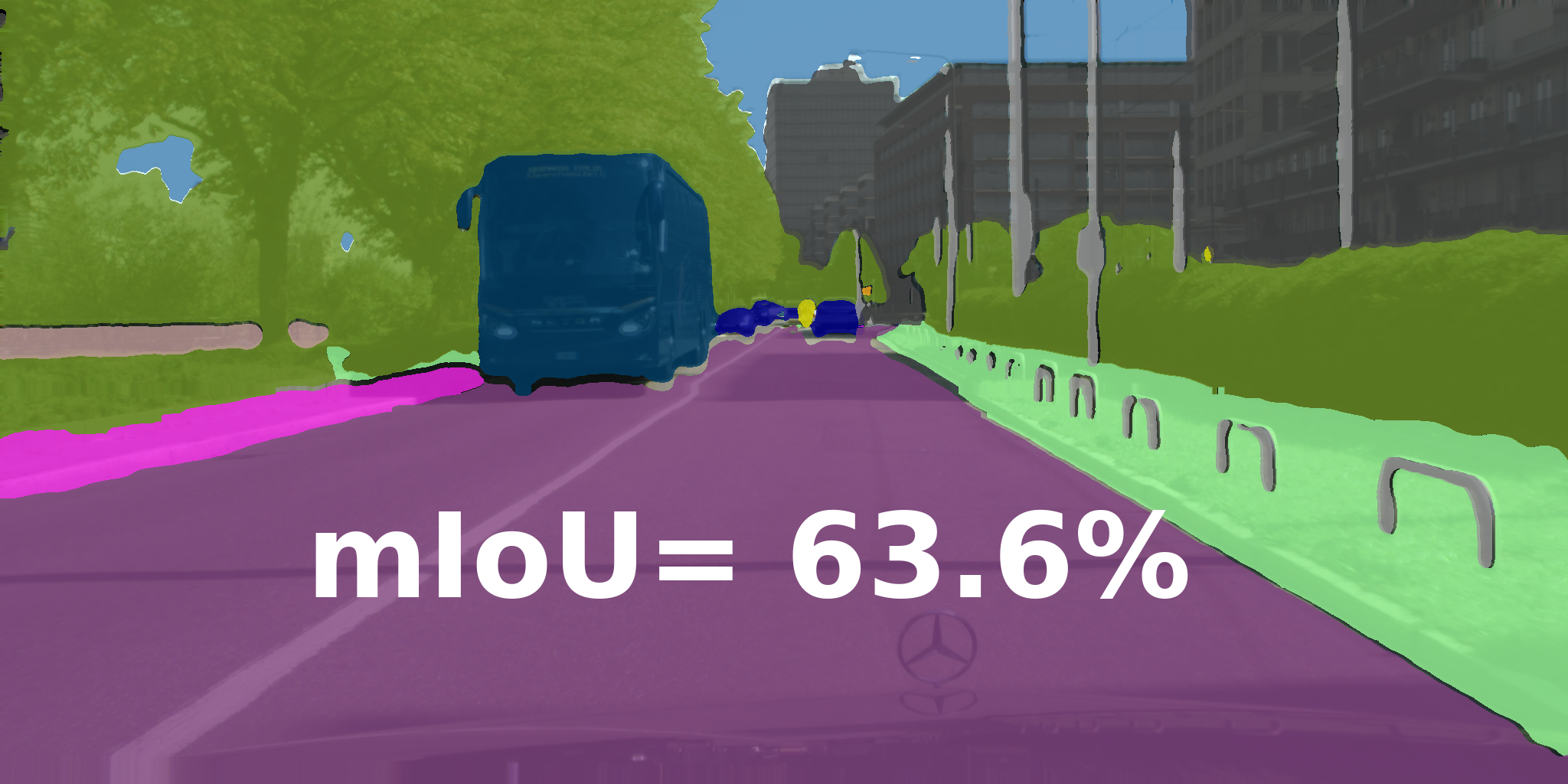}
  \includegraphics[width=\linewidth]{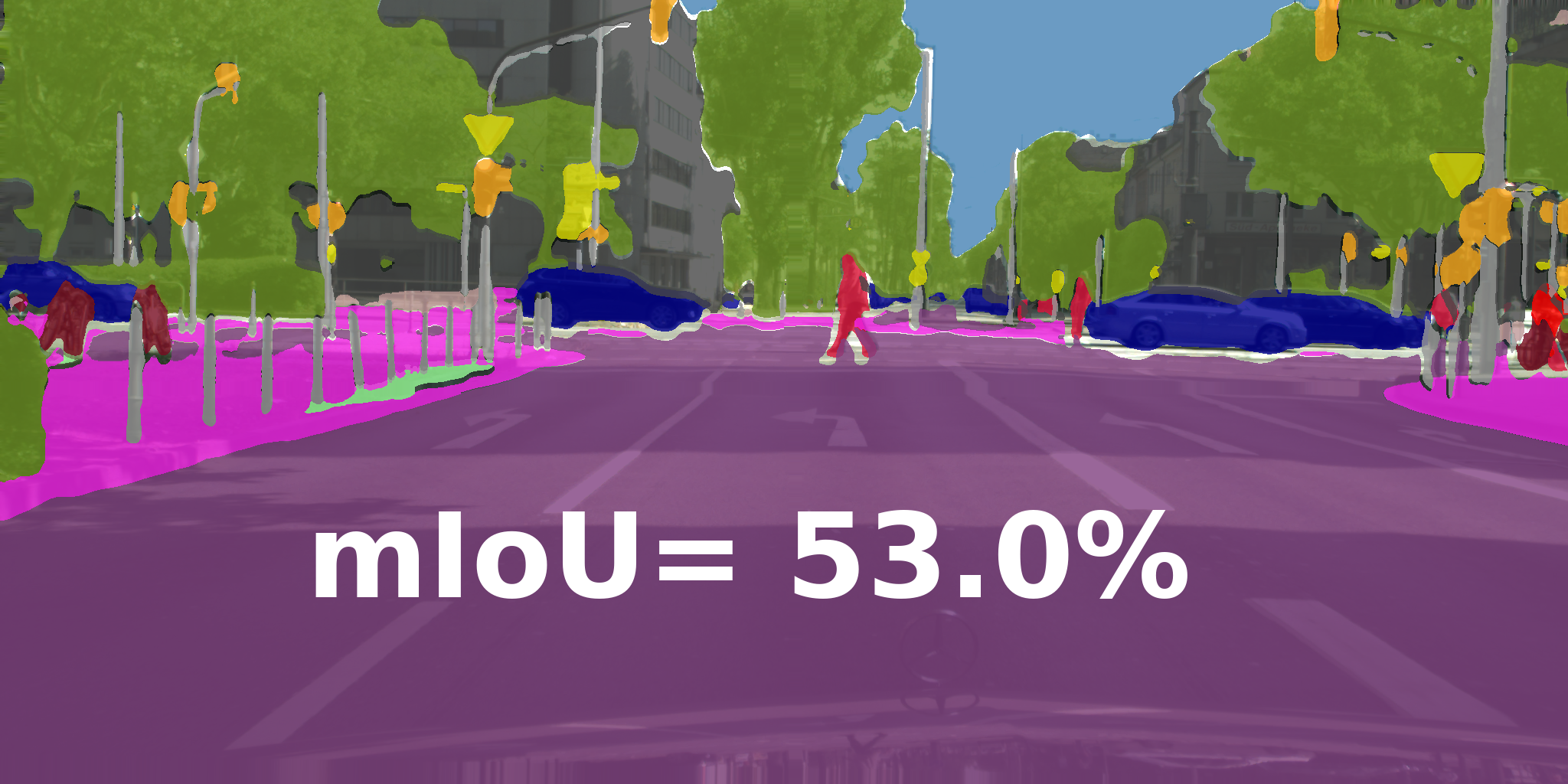}
  \includegraphics[width=\linewidth]{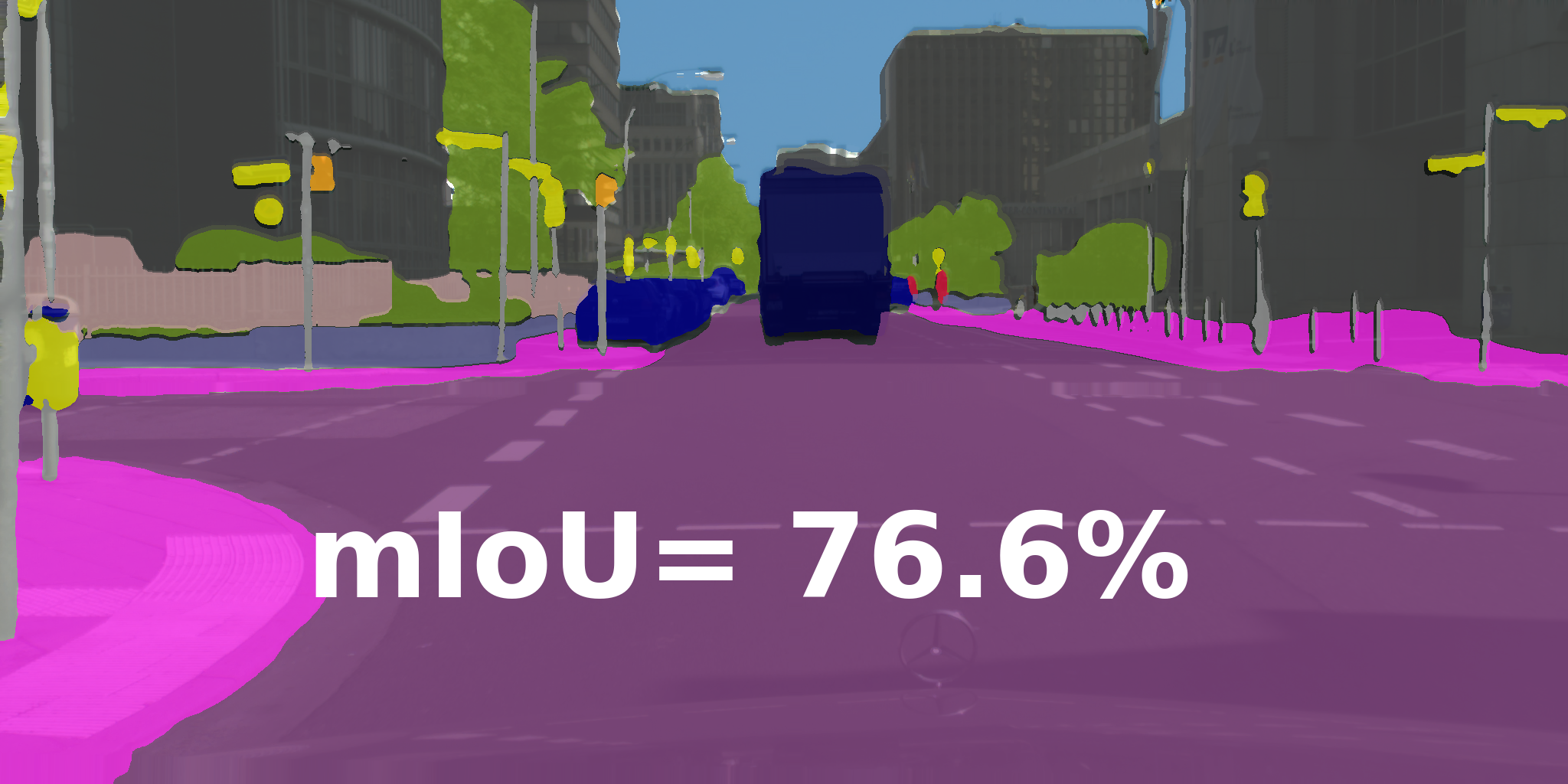}
  \includegraphics[width=\linewidth]{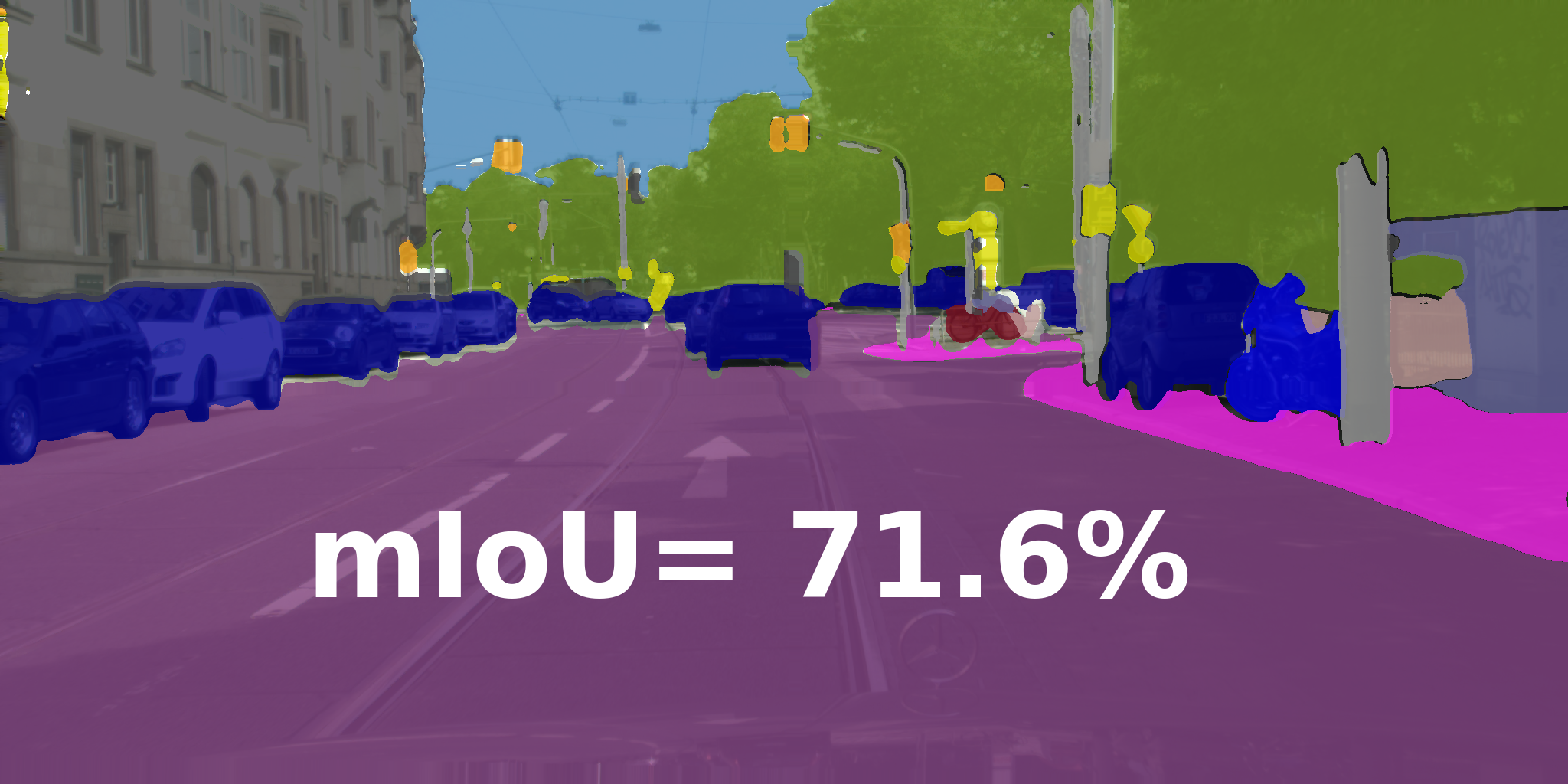}
    \includegraphics[width=\linewidth]{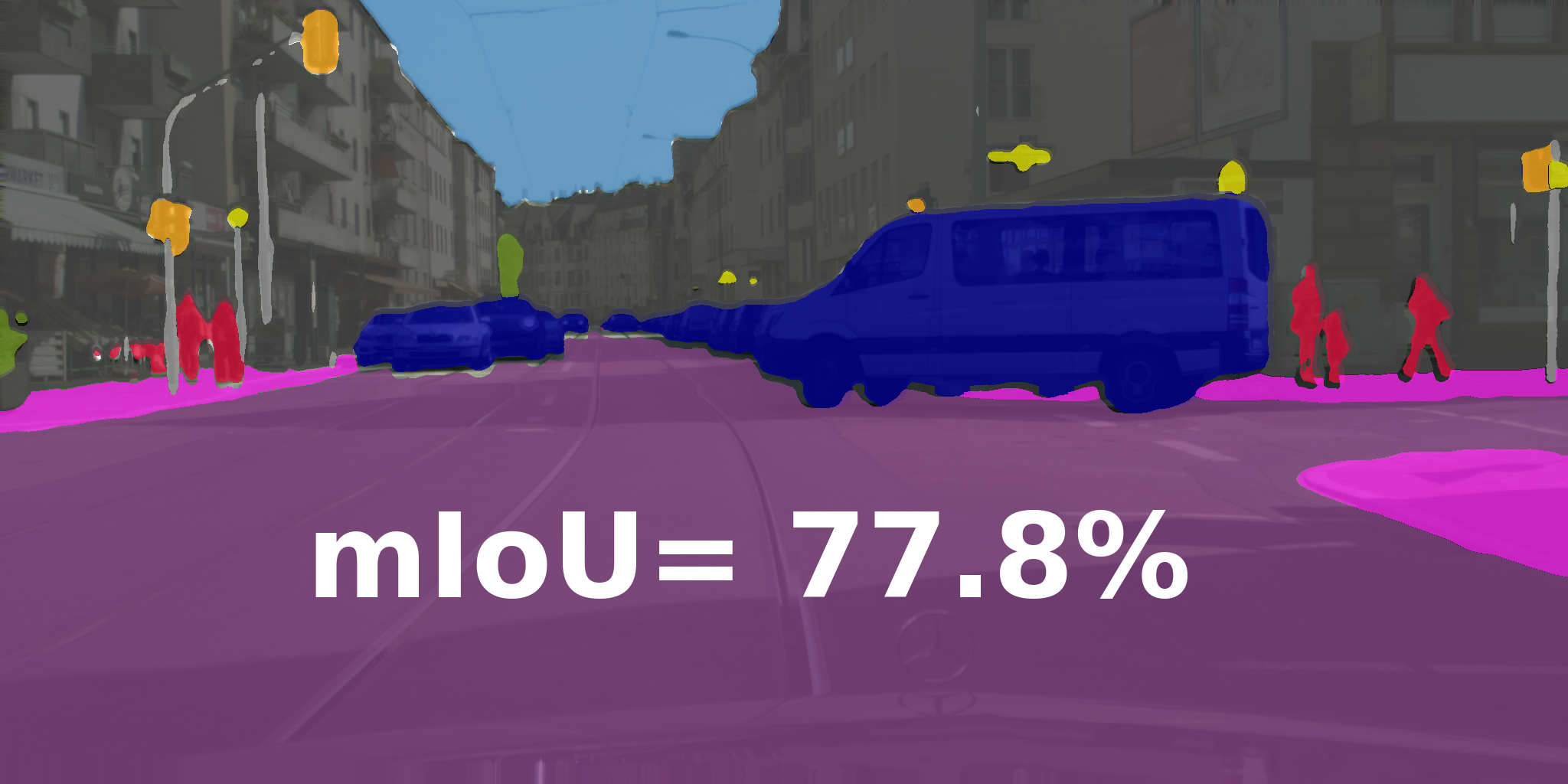}
  \vspace{0.2em}
  {\small c) \textsf{HSM} ($0.021\,\mathrm{bpp}$)}
\end{minipage}\hfill
%
\begin{minipage}[t]{0.2455\textwidth}
  \centering
  \includegraphics[width=\linewidth]{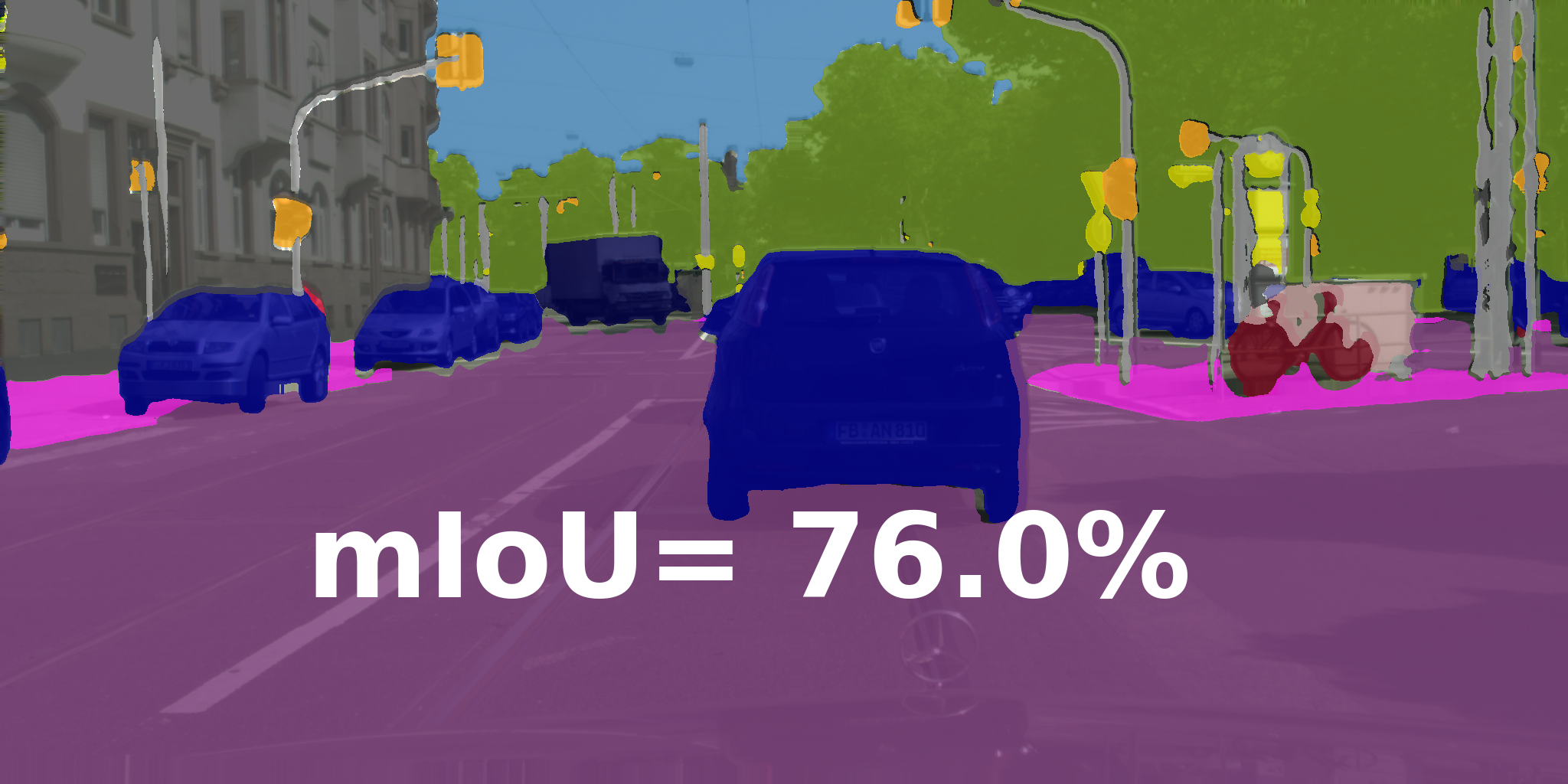}
  \includegraphics[width=\linewidth]{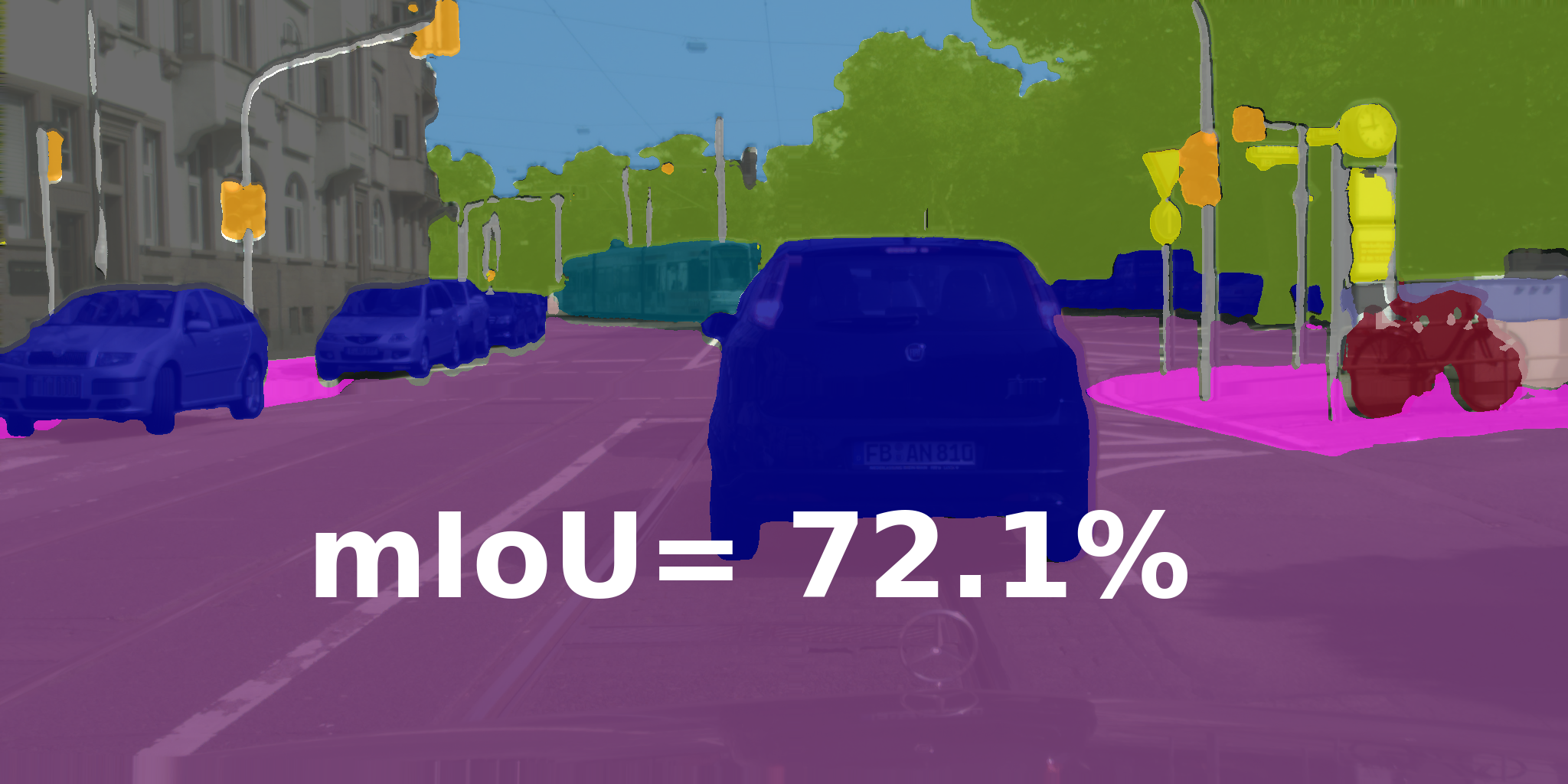}
  \includegraphics[width=\linewidth]{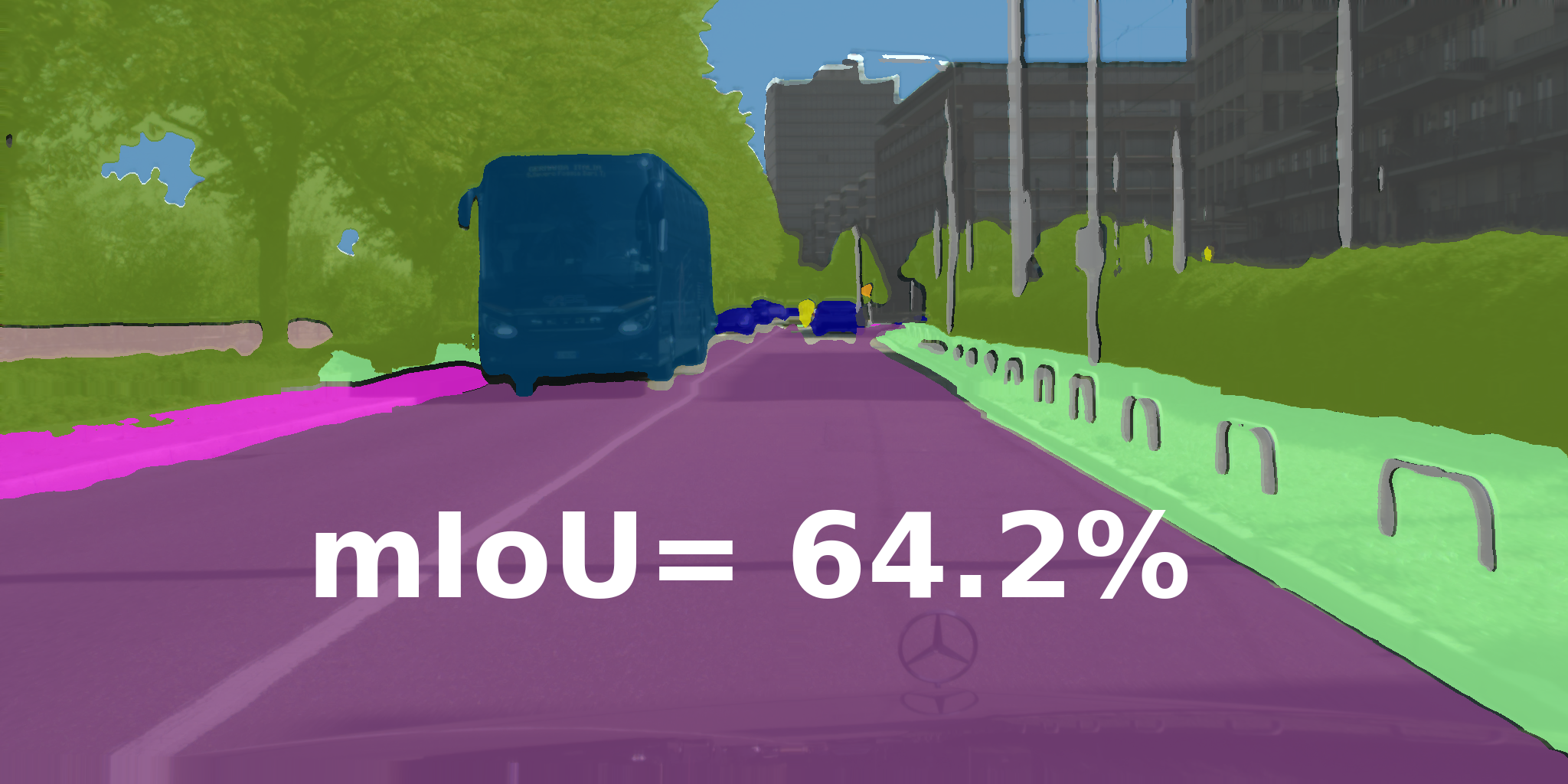}
  \includegraphics[width=\linewidth]{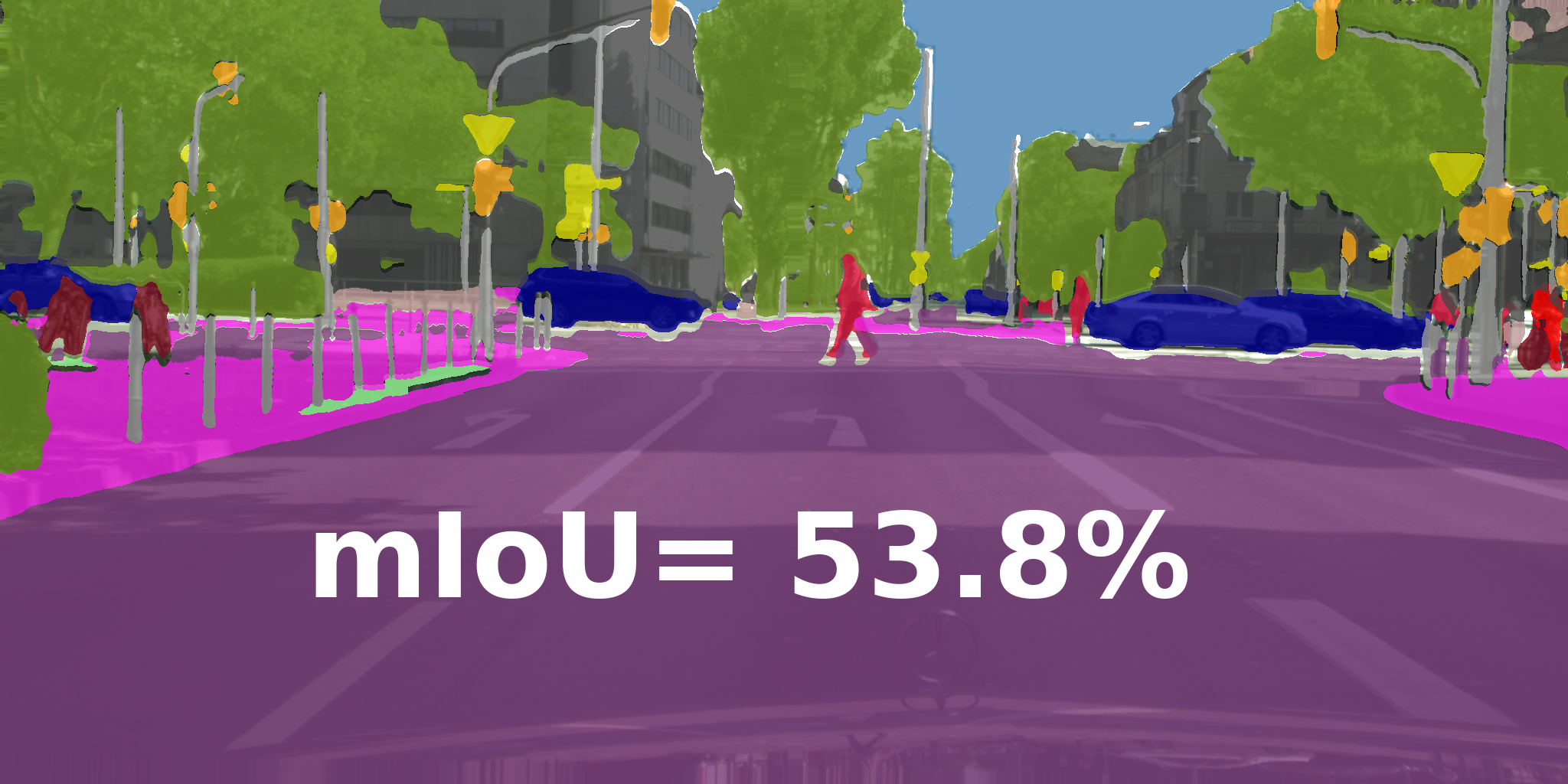}
  \includegraphics[width=\linewidth]{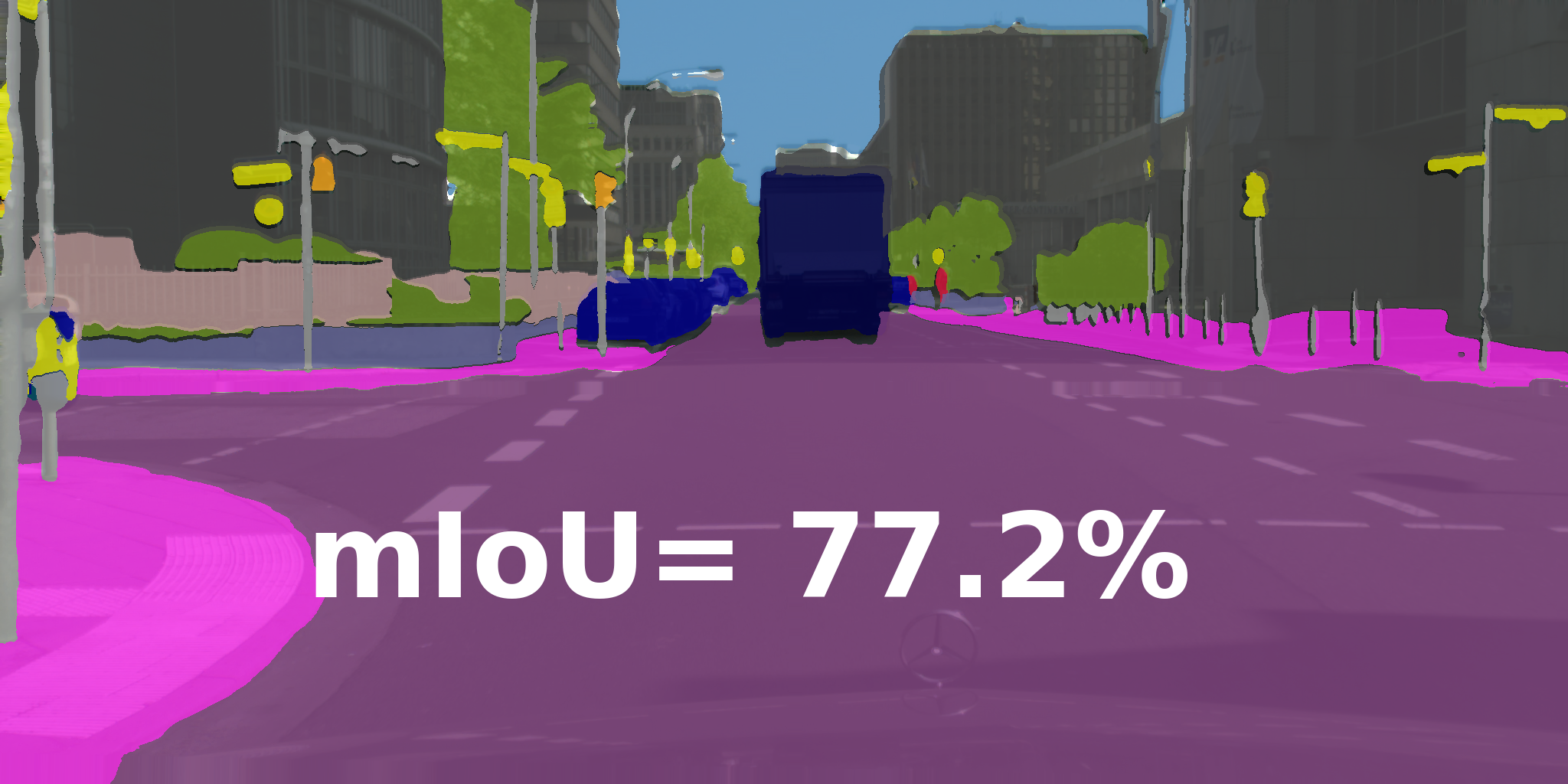}
  \includegraphics[width=\linewidth]{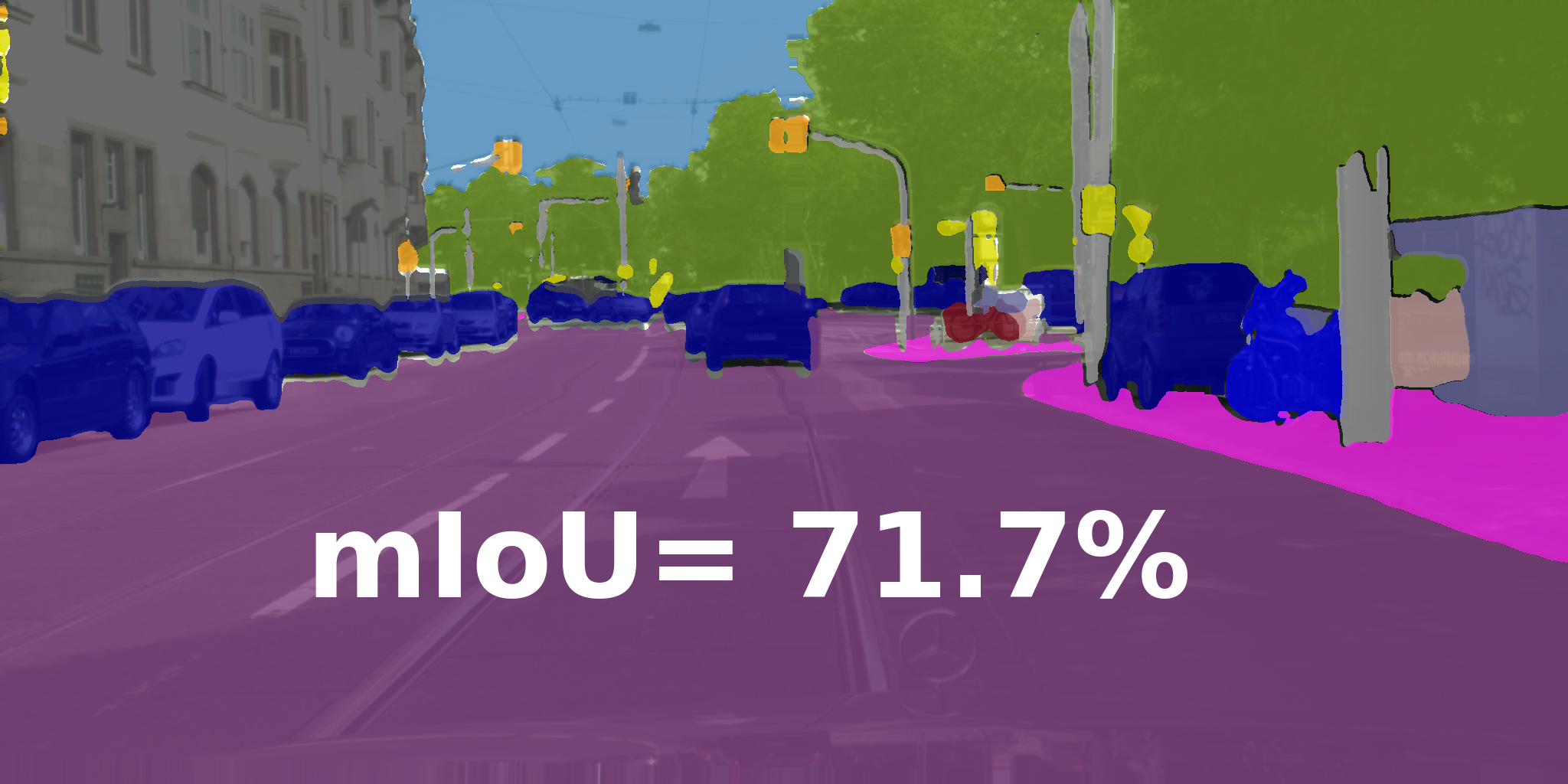}
 \includegraphics[width=\linewidth]{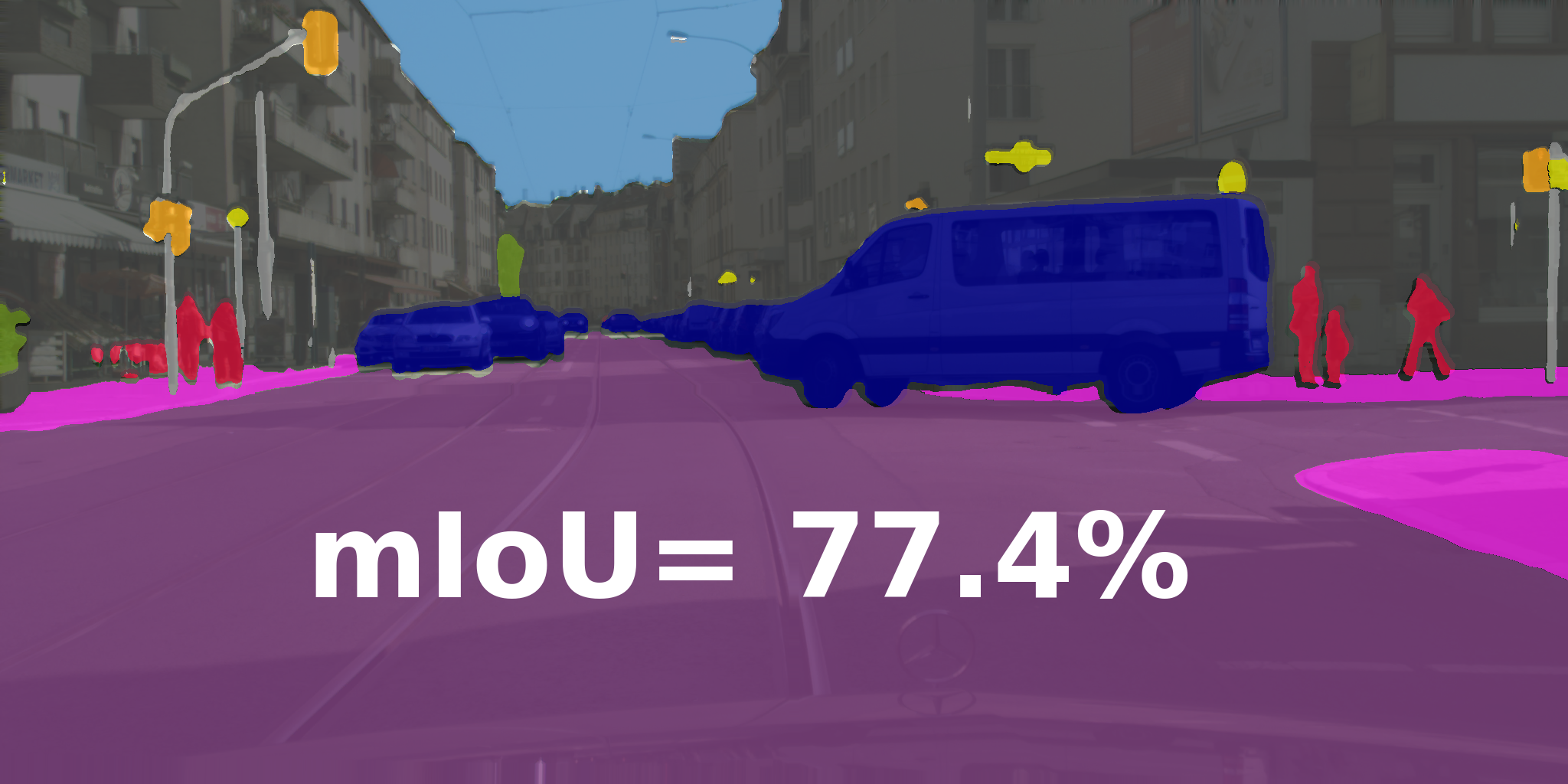}
  \vspace{0.2em}
  {\small d) \textsf{AR-HSM} ($0.009\,\mathrm{bpp}$)}
\end{minipage}

\caption{Qualitative comparison of the \textbf{proposed {\normalfont \textsf{HSM}} and {\normalfont \textsf{AR-HSM}} source codecs} against the so-far state-of-the-art {\normalfont \textsf{HS}} source codec on $\mathcal{D}^{\mathrm{CS}}_{\mathrm{val}}$ samples \textbf{at low bitrates.}}
\label{fig:cs_low_qualitative}
\end{figure}

\begin{figure}[t!]
\centering

\begin{minipage}[t]{0.2455\textwidth}
  \centering
    \includegraphics[width=\linewidth]{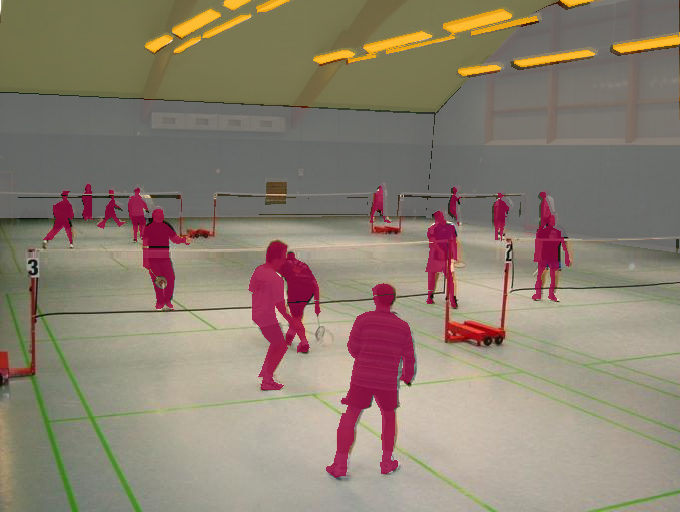}
  \includegraphics[width=\linewidth]{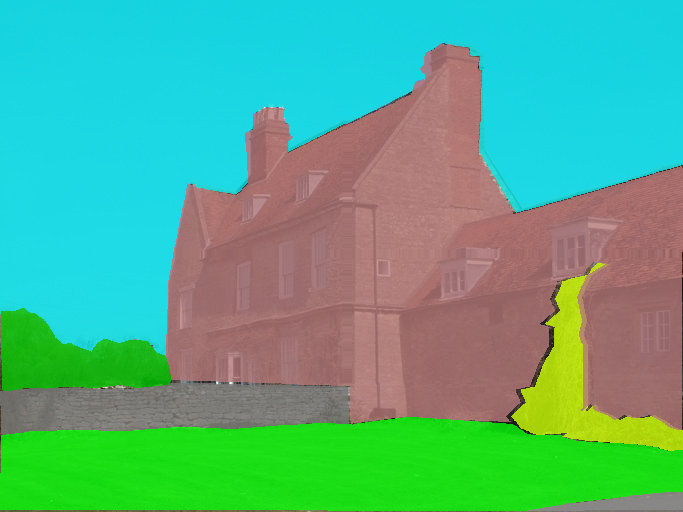}
  \includegraphics[width=\linewidth]{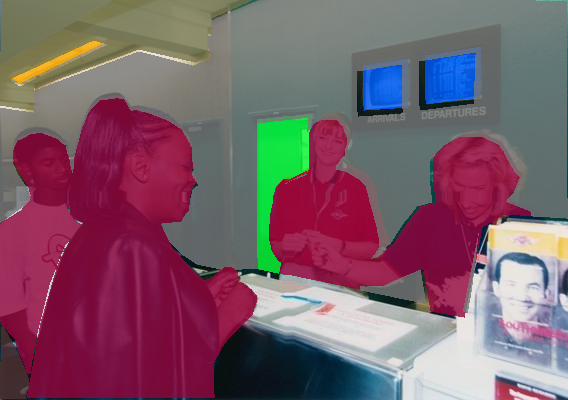}
  \includegraphics[width=\linewidth]{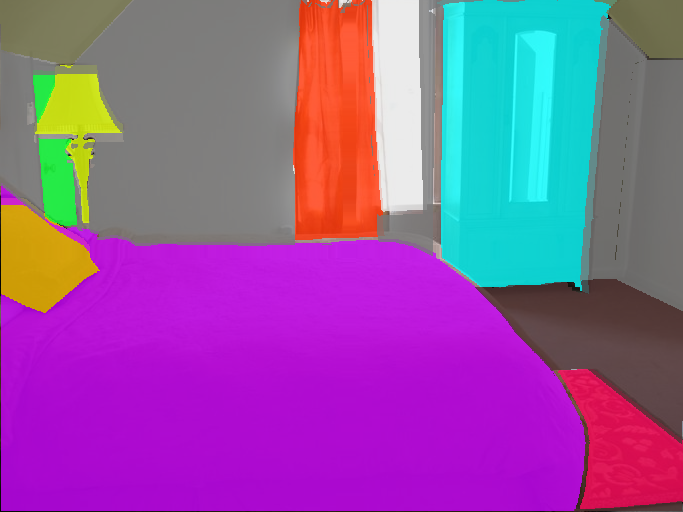}
  \includegraphics[width=\linewidth]{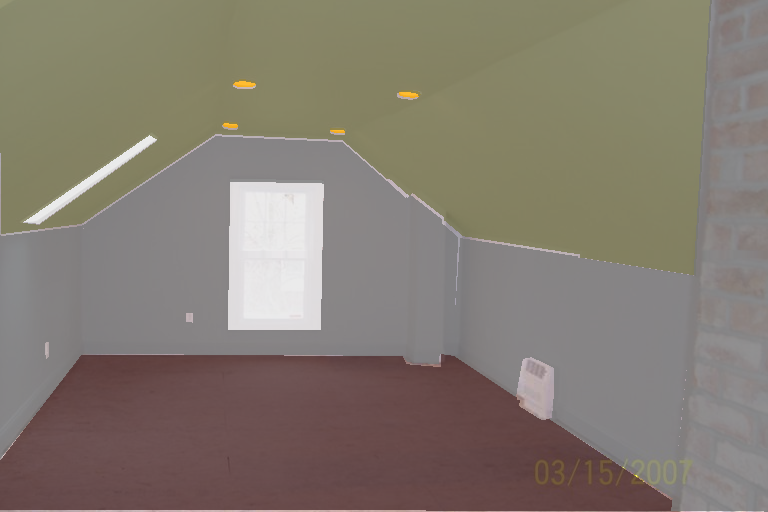}
  \includegraphics[width=\linewidth]{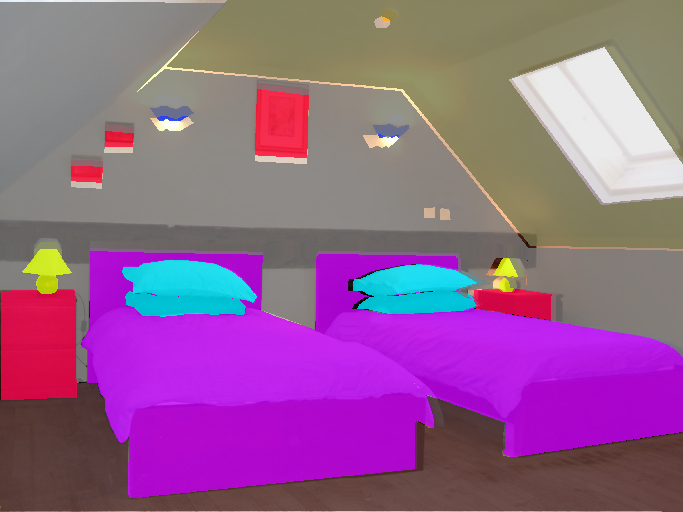}
    \includegraphics[width=\linewidth]{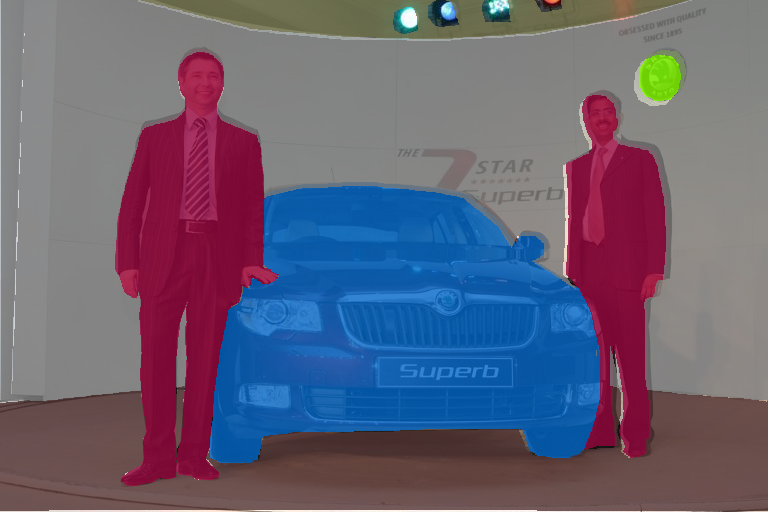}
  \vspace{0.2em}
  {\small a) Ground Truth}
\end{minipage}\hfill
%
\begin{minipage}[t]{0.2455\textwidth}
  \centering
    \includegraphics[width=\linewidth]{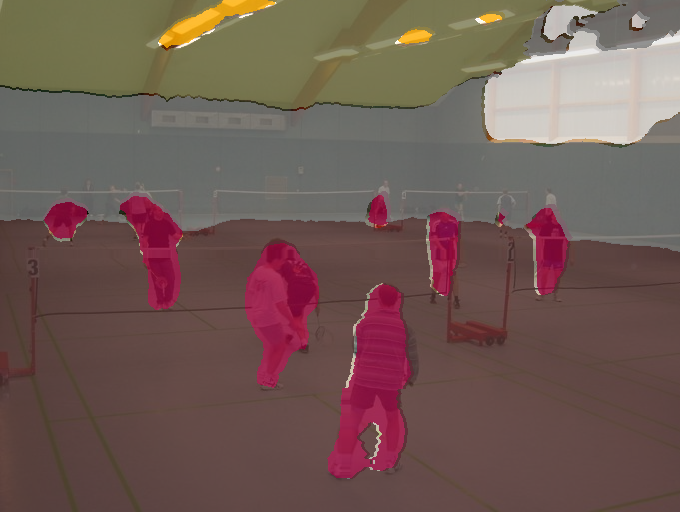}
  \includegraphics[width=\linewidth]{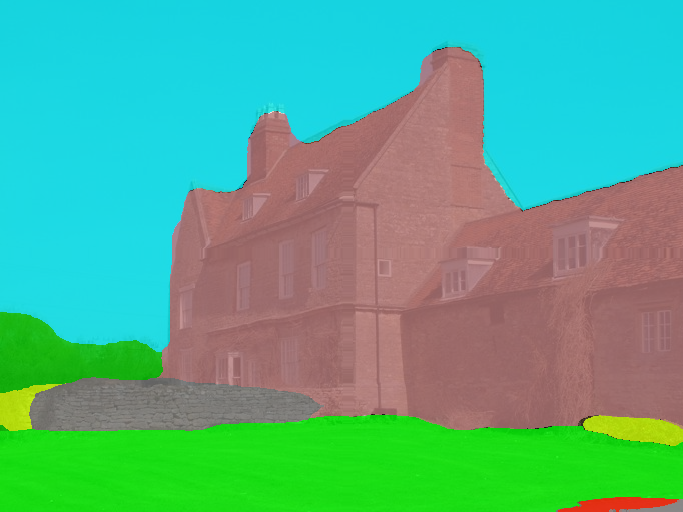}
  \includegraphics[width=\linewidth]{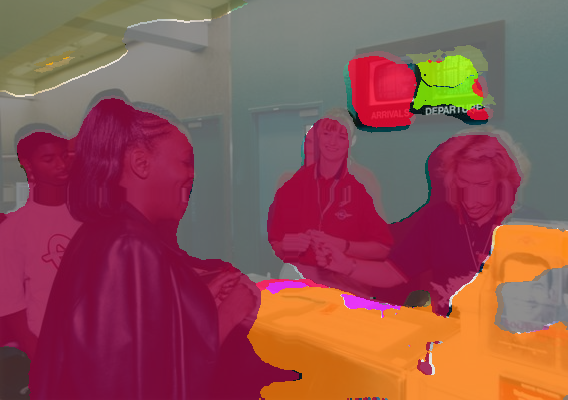}
  \includegraphics[width=\linewidth]{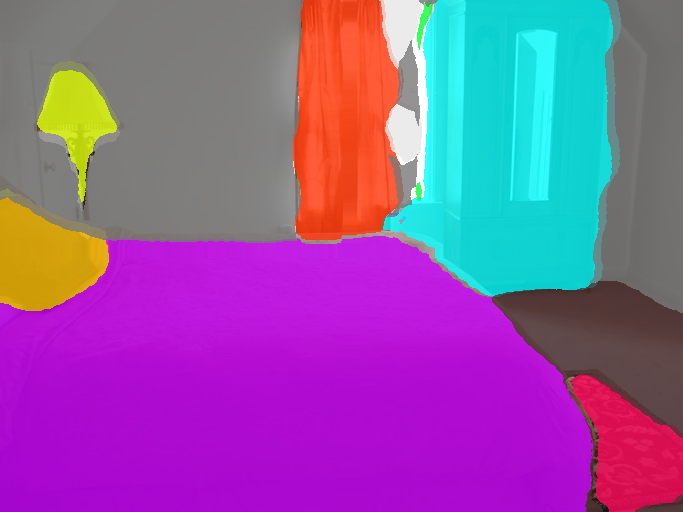}
  \includegraphics[width=\linewidth]{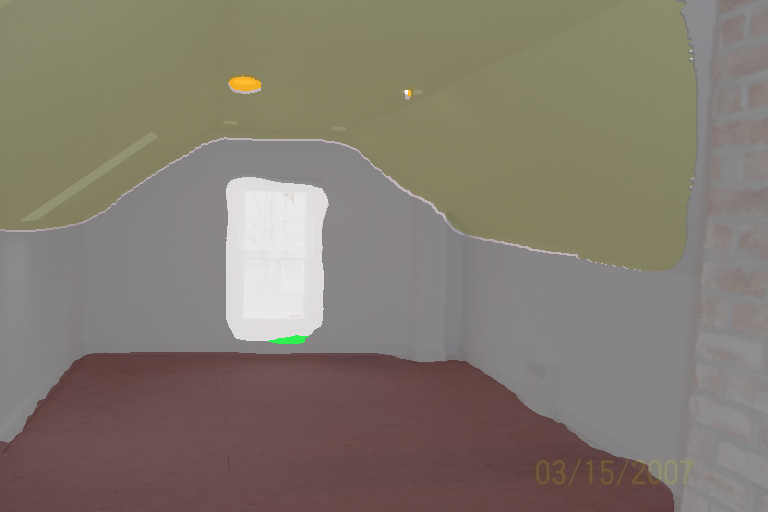}
  \includegraphics[width=\linewidth]{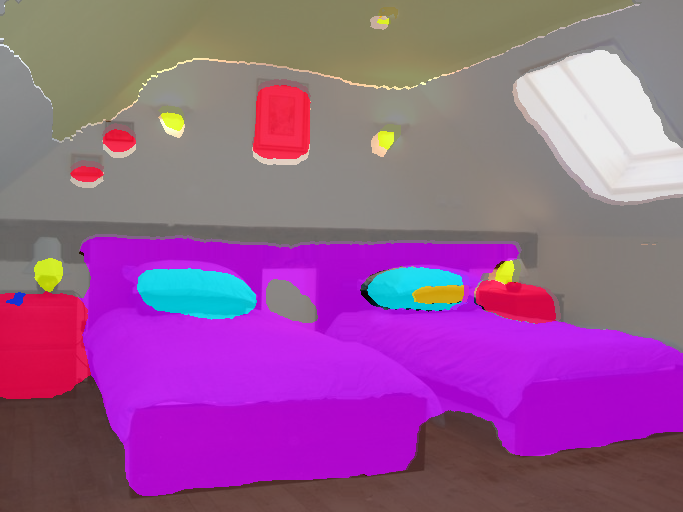}
    \includegraphics[width=\linewidth]{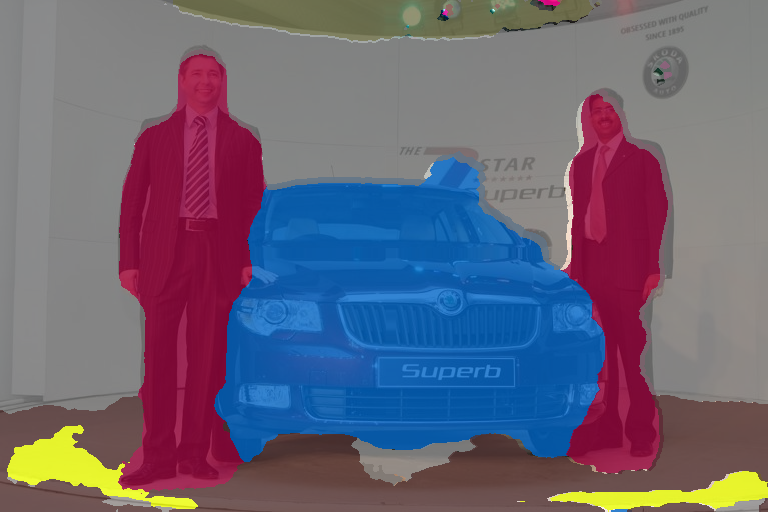}

  \vspace{0.2em}
  {\small b) \textsf{HS} ($0.02\,\mathrm{bpp}$)}
\end{minipage}\hfill
\begin{minipage}[t]{0.2455\textwidth}
  \centering
    \includegraphics[width=\linewidth]{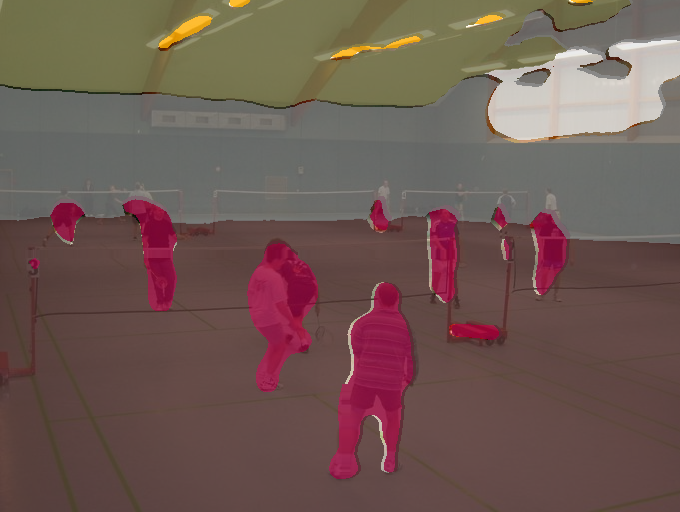}
  \includegraphics[width=\linewidth]{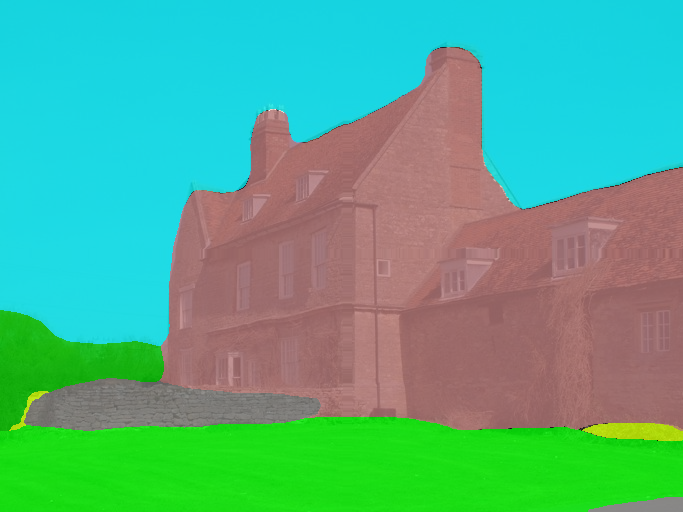}
  \includegraphics[width=\linewidth]{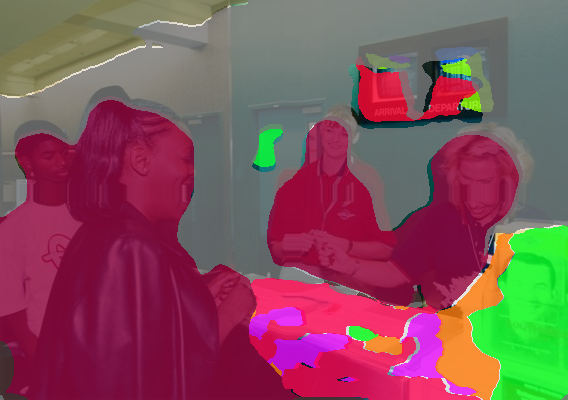}
  \includegraphics[width=\linewidth]{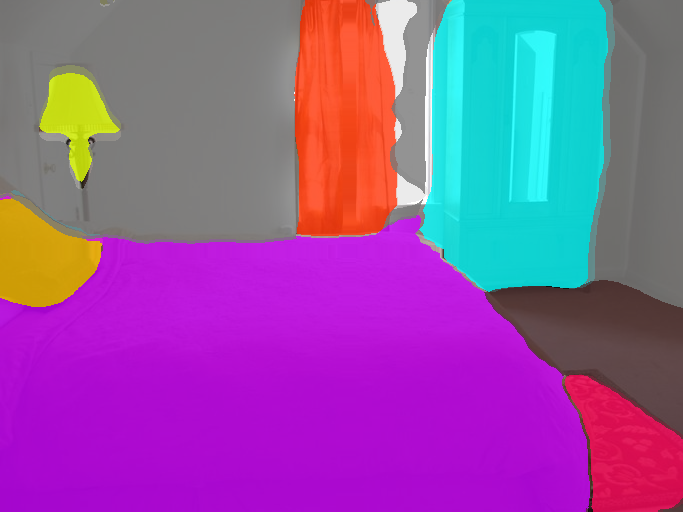}
  \includegraphics[width=\linewidth]{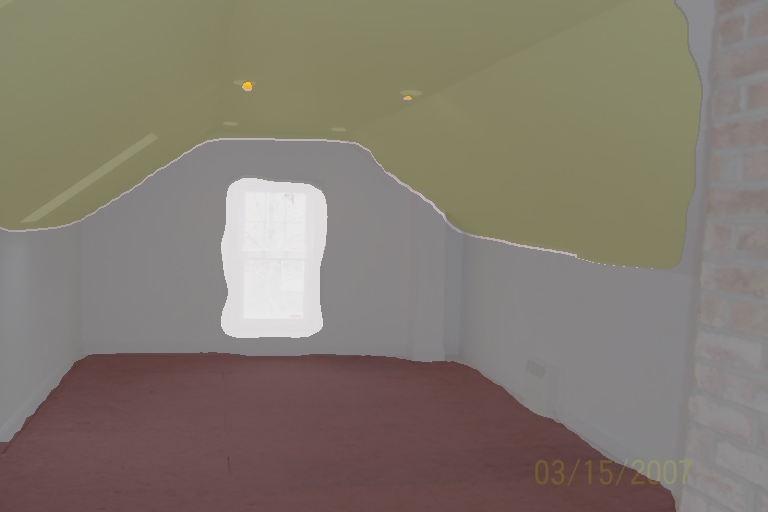}
  \includegraphics[width=\linewidth]{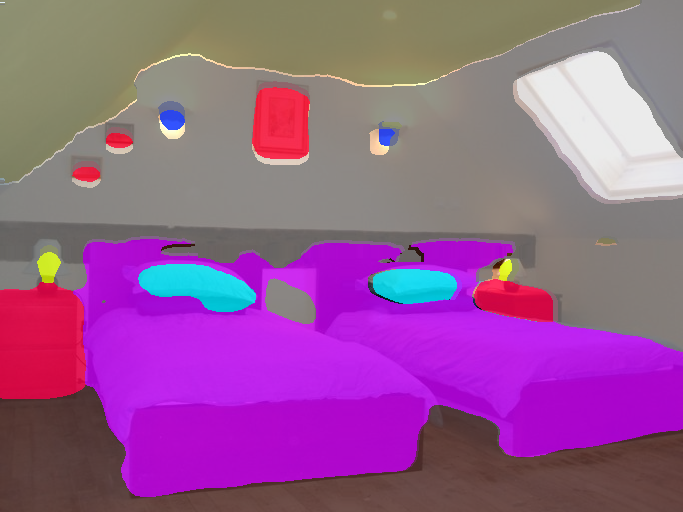}
    \includegraphics[width=\linewidth]{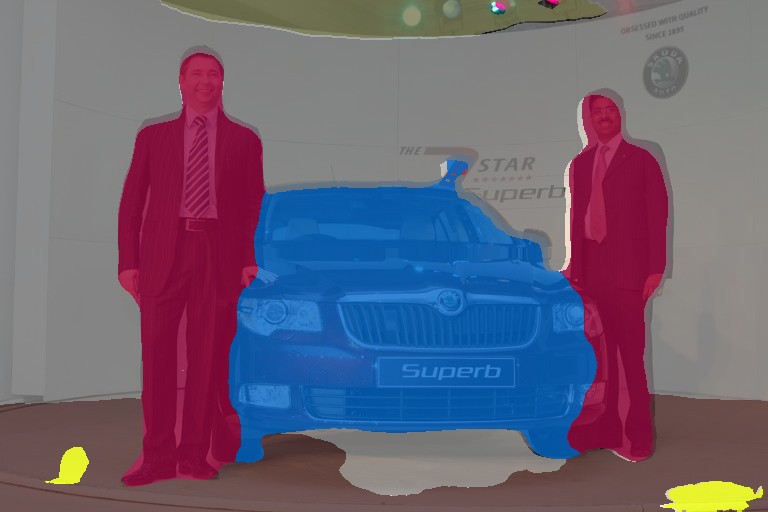}

  \vspace{0.2em}
  {\small c) \textsf{HSM} ($0.012\,\mathrm{bpp}$)}
\end{minipage}\hfill
%
\begin{minipage}[t]{0.2455\textwidth}
  \centering
    \includegraphics[width=\linewidth]{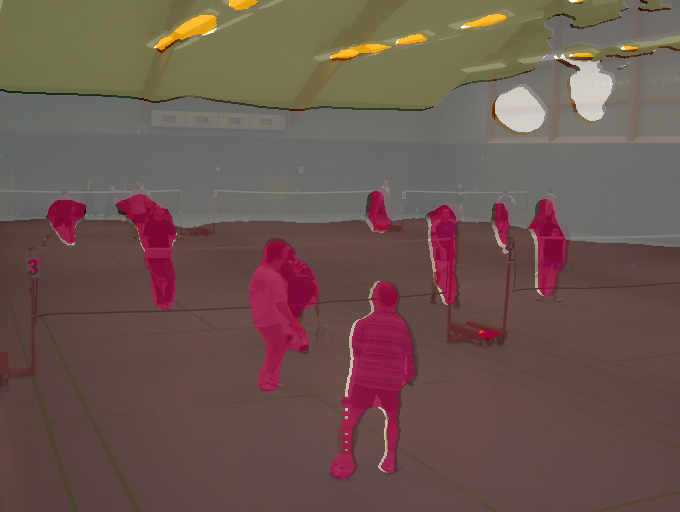}
  \includegraphics[width=\linewidth]{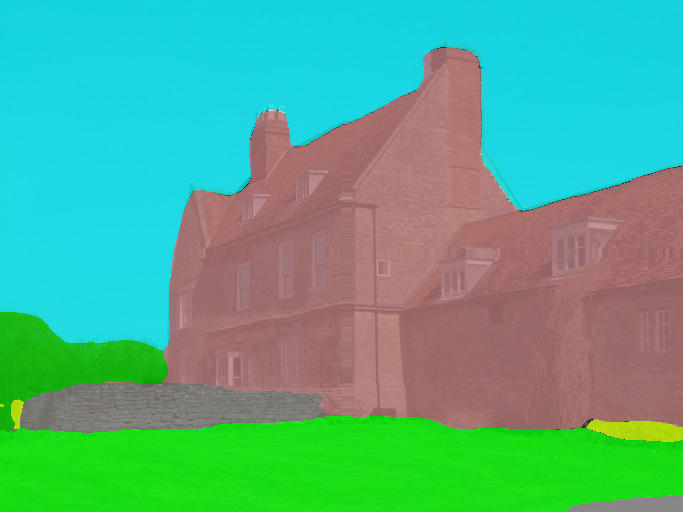}
  \includegraphics[width=\linewidth]{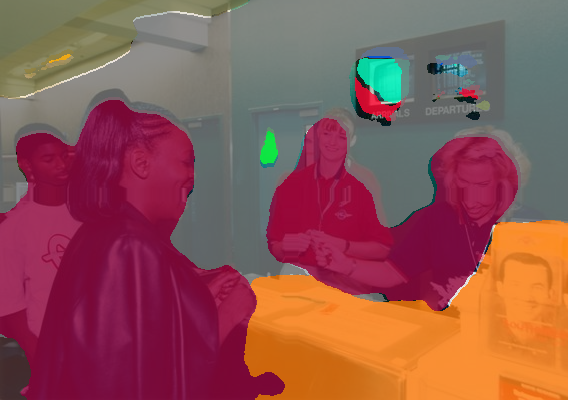}
  \includegraphics[width=\linewidth]{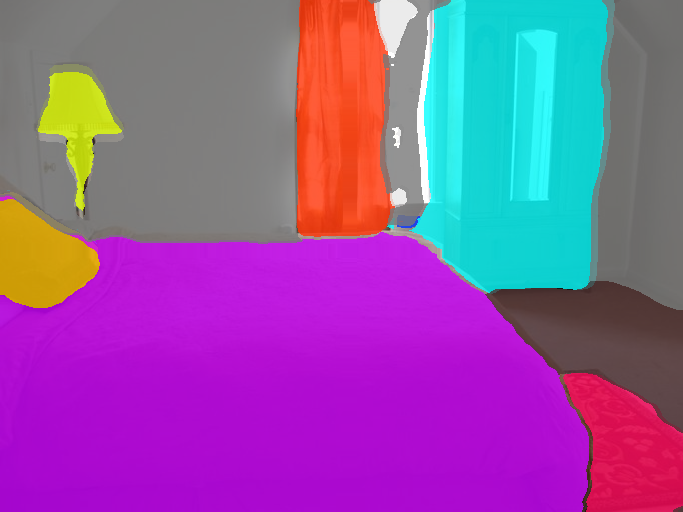}
  \includegraphics[width=\linewidth]{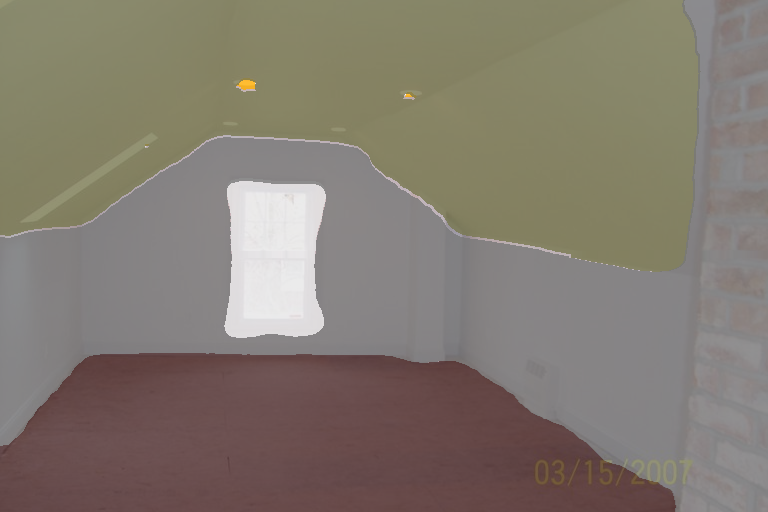}
  \includegraphics[width=\linewidth]{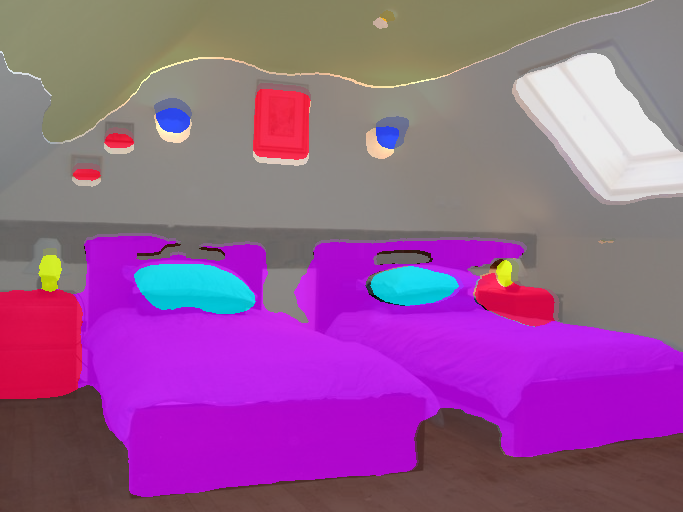}
    \includegraphics[width=\linewidth]{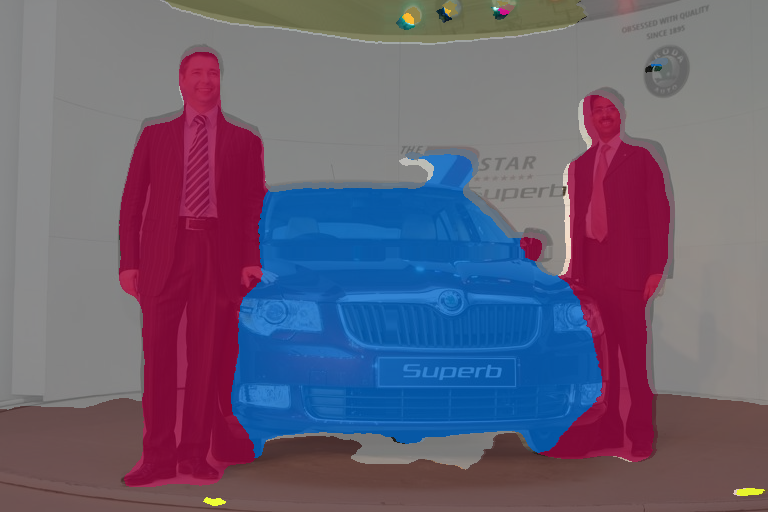}

  \vspace{0.2em}
  {\small d) \textsf{AR-HSM} ($0.007\,\mathrm{bpp}$)}
\end{minipage}

\caption{Qualitative comparison of the \textbf{proposed {\normalfont \textsf{HSM}} and {\normalfont \textsf{AR-HSM}} source codecs} against the so-far state-of-the-art {\normalfont \textsf{HS}} source codec on $\mathcal{D}^{\mathrm{ADE20K}}_{\mathrm{val}}$ samples \textbf{at low bitrates.}}
\label{fig:ade20k_low_qualitative}
\end{figure}

\bibliography{egbib}